\documentclass[10pt]{article}

\usepackage[margin=1.25in,letterpaper]{geometry}
\usepackage{microtype}
\usepackage{graphicx}
\usepackage{wrapfig}
\usepackage{booktabs}
\usepackage{amsmath,amssymb,amsfonts}
\usepackage{mathtools}
\usepackage{bm}
\usepackage{amsthm}
\usepackage{subcaption}
\usepackage{enumitem}
\usepackage{natbib}
\usepackage[usenames,dvipsnames]{xcolor}
\usepackage{hyperref}
\hypersetup{pdftitle={Amortized Vine Copulas for High-Dimensional Density and Information Estimation}, pdfauthor={Houman Safaai}}
\usepackage{url}
\IfFileExists{multirow.sty}{\usepackage{multirow}}{}
\IfFileExists{float.sty}{\usepackage{float}}{}

\definecolor{linkblue}{RGB}{20, 50, 140}
\hypersetup{colorlinks=true, linkcolor=linkblue, citecolor=linkblue, urlcolor=linkblue}

\usepackage{algorithm}
\IfFileExists{algpseudocode.sty}{%
  \usepackage{algpseudocode}%
}{%
  \usepackage{algorithmic}%
  \providecommand{\State}{\STATE}%
  \providecommand{\For}{\FOR}%
  \providecommand{\EndFor}{\ENDFOR}%
  \providecommand{\Comment}[1]{\hfill$\triangleright$~{\footnotesize ##1}}%
}

\newtheorem{proposition}{Proposition}

\newtheorem{definition}{Definition}

\title{Amortized Vine Copulas for High-Dimensional Density and Information Estimation}

\author{%
  Houman Safaai \\
  \normalsize Kempner Institute for the Study of Natural and Artificial Intelligence at Harvard University \\
  \normalsize \texttt{houman\_safaai@harvard.edu}%
}
\date{}

\usepackage[font=small,labelfont=bf]{caption}

\def\arxivmode{1}

\newcommand{\availabilitynote}{\footnote{Code: \url{https://github.com/KempnerInstitute/vine-denoising-copula}. Model: \url{https://huggingface.co/hsafaai/vdc-denoiser-m64-v1}.}}

\begin{document}

\newcommand{\copula}{\mathcal{C}}
\newcommand{\RR}{\mathbb{R}}
\newcommand{\EE}{\mathbb{E}}
\newcommand{\PP}{\mathbb{P}}
\newcommand{\defeq}{\vcentcolon=}
\newcommand{\hfunc}{h}
\newcommand{\norm}[1]{\lVert #1 \rVert}
\providecommand{\ours}{\textsc{VDC}}
\newcommand{\cmark}{\ensuremath{\checkmark}}
\newcommand{\xmark}{\ensuremath{\times}}
\newcommand{\paneltag}[1]{\colorbox{white}{\strut\small\textbf{(#1)}}}
\providecommand{\argmax}{\operatorname*{arg\,max}}
\providecommand{\argmin}{\operatorname*{arg\,min}}
\providecommand{\KL}{\operatorname{KL}}

\maketitle

\begin{abstract}
Modeling high-dimensional dependencies while keeping likelihoods tractable remains challenging. Classical vine copulas provide an explicit dependence structure but require repeated, costly bivariate fitting. We propose Vine Denoising Copula (VDC)\availabilitynote{}, an amortized estimator for simplified vines on continuous data. VDC trains a single bivariate denoising network on simulated copulas and reuses it across all vine edges. For each edge, the network maps an empirical histogram to a density grid, and iterative proportional fitting approximately enforces uniform grid marginals. The predicted densities and cached conditional transforms support standard vine recursion, likelihood evaluation, sampling, and information estimation. This retains the classical vine factorization while replacing repeated per-edge fitting with reusable neural inference. Across synthetic and real-data benchmarks, VDC combines accurate bivariate density estimation, competitive information estimates, and faster high-dimensional vine fitting after pretraining. Performance varies across datasets; finite grid resolution and conditional prediction remain limitations. The explicit fitted structure supports edge- and tree-level information decomposition, making repeated dependence analysis more practical.

\end{abstract}

\section{Introduction}
\label{sec:intro}

High-dimensional dependence modeling must balance flexible density estimation with tractable likelihoods and interpretable structure. Neural density estimators are flexible but often obscure dependence structure, whereas classical statistical models expose structure at the cost of repeated, expensive fitting.

\begin{figure*}[!t]
    \centering
    \includegraphics[width=\textwidth]{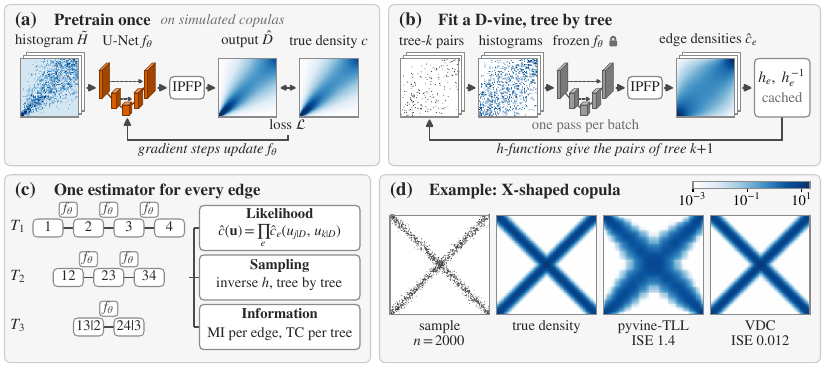}
    \caption{\textbf{VDC overview.} (a) A U-Net $f_\theta$ is trained once on simulated copulas (Appendix~\ref{app:training_data}): it takes a histogram corrupted at a random noise level $t$, conditioned on $t$, and its IPFP-projected output is trained against the true density (shown: one input with mixing weight $0.3$ and the network's output for it). (b) A D-vine is fitted tree by tree with the frozen $f_\theta$: edges are evaluated in batches, one network pass per batch, and cached h-functions give the pseudo-observations of the next tree. (c) One estimator serves every edge; the fitted vine provides likelihood, sampling and information. (d) On a fresh sample from the X-shaped zoo pattern, \ours{} preserves the crossing structure smoothed by pyvine-TLL (ISE: integrated squared error; aggregate errors in Appendix Table~\ref{tab:bivariate_complex_appendix}). Predicted densities in (a) and (b) are outputs of the deployed checkpoint on illustrative samples; (d) is a measured comparison.}
    \label{fig:overview}
\end{figure*}

Vine copulas \citep{aas2009pair, bedford2002vines} provide a useful middle ground. Via Sklar's theorem \citep{sklar1959fonctions}, they separate marginals from dependence and factorize a copula of dimension $d$ into $\binom{d}{2}$ pair-copulas over a tree hierarchy. This allows tractable likelihood evaluation under the fitted vine, efficient conditional sampling, and interpretable dependence attribution. A key bottleneck is that each edge requires fitting a separate bivariate copula. Parametric families are fast to fit but risk misspecification; nonparametric estimators are flexible \citep{nagler2018kdecopula} but expensive. The repeated fitting cost becomes prohibitive as dimension grows.

We propose \ours{} (Vine Denoising Copula), a method for \emph{amortized} bivariate estimation inside a classical vine. Instead of fitting a new model for every edge, we train a single neural network once on a diverse synthetic copula collection and reuse it throughout the vine. For each edge, the shared estimator reads an empirical bivariate histogram and predicts a density grid in one forward pass. We then project that prediction toward the set of copula densities using Iterative Proportional Fitting (IPFP), which enforces the uniform marginals required of a copula up to a controlled tolerance \citep{sinkhorn1964relationship,sinkhorn1967knopp}. \ours{} learns a bivariate estimator while retaining the classical vine factorization. The same estimator supplies the pair-copula densities needed throughout the vine, without fitting a separate neural model to each edge or replacing the factorization with a global density model. This reduces the cost of vine fitting at moderate and high dimension and supports repeated information analysis with an explicit dependence structure.

Concretely, our contributions are:
\begin{itemize}[leftmargin=*, itemsep=1pt, topsep=2pt]
    \item \textbf{An amortized, vine-compatible edge estimator.} A single bivariate denoising network, trained once on synthetic copulas, maps empirical copula histograms to explicit density grids; an IPFP/Sinkhorn projection restores copula marginal constraints to controlled numerical tolerance (Sections~\ref{subsec:vdc_denoiser}--\ref{subsec:ipfp}).
    \item \textbf{Integration into vine recursion.} A batched, cached integration into standard vine recursion---histogram, forward pass, projection, cached h-functions---that removes per-edge retraining and expensive nonparametric fitting while retaining explicit pair-copula recursion (Section~\ref{subsec:hfunc}).
    \item \textbf{Structured information estimation.} Copula-entropy estimates of mutual information (MI) and total correlation (TC) with edge- and tree-level decompositions aligned with the fitted vine; in our experiments these density-based estimates show significantly fewer data-processing-inequality (DPI) violations, at the level of independent base datasets, and lower additivity error than Kraskov--St{\"o}gbauer--Grassberger (KSG) estimates under the protocol in Appendix~\ref{app:self_consistency}.
\end{itemize}
Figure~\ref{fig:overview} previews the workflow.

Across our benchmarks, the same frozen estimator combines accurate bivariate fits with faster high-dimensional vine fitting. Its strongest real-data results are on S\&P100, HEPMASS and MINIBOONE; performance varies with the dataset and sample size. Sections~\ref{sec:experiments} and~\ref{sec:conclusion} report the comparisons and discuss grid resolution, imputation accuracy and tail calibration.

\section{Background}
\label{sec:background}

\subsection{Copulas and Vines}

Let $X=(X_1,\ldots,X_d)$ be a continuous random vector, let $F$ be its joint cumulative distribution function (CDF), and let $F_1,\ldots,F_d$ be its marginal CDFs. A copula $\copula: [0,1]^d \to [0,1]$ is a multivariate CDF with uniform marginals. Sklar's theorem \citep{sklar1959fonctions} gives, for $x=(x_1,\ldots,x_d)$,
\begin{equation}
F(x_1, \ldots, x_d) = \copula(F_1(x_1), \ldots, F_d(x_d)).
\end{equation}
When the densities exist, let $f$ denote the joint density, $f_i$ the marginal density of $X_i$, and $c$ the copula density. They satisfy $f(x)=c(F_1(x_1),\ldots,F_d(x_d))\prod_i f_i(x_i)$. Define the probability-integral-transform variables $U_i=F_i(X_i)$ and, for an index set $D$, the subvector $U_D=(U_i)_{i\in D}$. For the realization $x$, write $u_i=F_i(x_i)$.
Regular vines \citep{bedford2002vines} extend this decomposition to high dimension by writing a multivariate copula density as a hierarchy of bivariate building blocks. Let $T_\ell$ denote the edge set in tree $\ell$, and let $c_{j,k|D}$ denote the conditional pair-copula density for variables $j$ and $k$ given the index set $D$:
\begin{equation}
\label{eq:vine_factorization}
c(u_1, \ldots, u_d) = \prod_{\ell=1}^{d-1} \prod_{(j,k|D) \in T_\ell} c_{j,k|D}(u_{j|D}, u_{k|D}),
\end{equation}
where $(j,k|D)\in T_\ell$ is an edge with conditioning set $D \subset \{1,\ldots,d\}$, and $u_{j|D}$ denotes the realized conditional probability-integral-transform value of $U_j$ given $U_D$. Throughout the main text we adopt the simplifying assumption: each conditional copula $c_{j,k|D}$ depends on the conditioning variables only through the conditional arguments $(u_{j|D}, u_{k|D})$, not through the realized value of $U_D$. For a bivariate pair $(U,V)$ with copula density $c_{UV}$, these conditional values are computed recursively through \emph{h-functions},
\begin{equation}
h_{U|V}(u \mid v) = \int_0^u c_{UV}(s, v) \, ds = \PP(U \leq u \mid V = v),
\end{equation}
and standard structure learning uses Di{\ss}mann's greedy maximum-spanning-tree procedure \citep{dissmann2013selecting}. Figure~\ref{fig:overview}(c) illustrates a four-dimensional vine factorization.

\subsection{Related Work}

Vine copulas have become a standard tool for multivariate dependence modeling \citep{aas2009pair, joe2014dependence, czado2019analyzing}, with efficient implementations in libraries like \texttt{vinecopulib} \citep{nagler2025rvinecopulib}. However, these still require fitting $O(d^2)$ bivariate copulas, either through parametric family selection or kernel density estimation \citep{nagler2018kdecopula}. The simplifying assumption underlying most vine pipelines, that conditional pair-copulas depend on the conditioning set only through the conditional CDFs and not through realized values, has been studied as both a useful approximation \citep{haff2010simplified} and a potential source of misspecification \citep{spanhel2019simplified}; we probe this directly in a non-simplified stress test (Appendix~\ref{app:nonsimplified_stress}).

To sidestep the repeated-fitting cost, recent work has explored neural alternatives: implicit generative copulas \citep{janke2021igc}, neural copula functions \citep{zeng2022neuralcopula}, family-specific neural pair-copulas such as Deep Archimedean Copulas \citep{ling2020deep} and generative Archimedean constructions \citep{ng2021generative}, vine-structured latent generative models \citep[a vine fitted in an autoencoder latent space;][]{tagasovska2019vinecopula}, copula and marginal flows \citep{wiese2019copulaflow}, copula density estimators \citep{letizia2025codine}, classifier-based density-ratio copulas \citep{huk2025classifier}, and diffusion- and flow-based copulas \citep{huk2026diffusionflow}. We instead amortize only the bivariate edge estimator, enforce copula marginals by IPFP, and keep the explicit pair-copula densities and h-functions needed for vine recursion, conditional sampling, and edge-wise MI/TC decomposition. This approach builds on amortized inference methods, including neural processes, prior-fitted networks, and simulation-based inference \citep{garnelo2018cnp, mueller2022pfn, hollmann2023tabpfn, cranmer2020sbi}; copula projection and h-function recursion adapt these methods to vines, and an ablation demonstrates the role of projection in enforcing the copula constraints (Appendix~\ref{app:noipfp_ablation}). Normalizing flows \citep{dinh2017realnvp, papamakarios2017maf, durkan2019nsf} and diffusion models \citep{kotelnikov2023tabddpm} are flexible global density estimators but do not expose pair-copulas, h-functions, or edge-wise information decomposition.

\section{Method}
\label{sec:method}

\ours{} replaces the repeated bivariate fitting step of a classical vine with one pretrained operator: each edge's samples become a copula-space histogram, one forward pass predicts a density grid, IPFP projects it onto the copula constraints, and the resulting h-functions generate pseudo-observations for the next tree.

\subsection{Problem Formulation}

Given $n$ raw edge observations $\{(x_i,y_i)\}_{i=1}^n$, we first map them into copula space with the empirical probability integral transform (PIT), $u_i = r_i^{(x)}/(n+1)$ and $v_i = r_i^{(y)}/(n+1)$, where $r_i^{(x)}$ and $r_i^{(y)}$ are the ranks of $x_i$ and $y_i$ within their respective samples \citep{nelsen2006introduction, genest2009goodness}. Next, we discretize $[0,1]^2$ into an $m \times m$ grid, where $m$ is the resolution along each axis and $a,b\in\{1,\ldots,m\}$ index cells. Our goal is to estimate a valid density grid $\hat{D} \in \RR_{\geq 0}^{m \times m}$. Let $\Delta = 1/m$ be the bin width, let $B_{ab}$ denote cell $(a,b)$, and let $\mathbf{1}[\,\cdot\,]$ denote the indicator function. The model input is the histogram density estimate
\begin{equation}
H_{ab} = \frac{1}{n \Delta^2} \sum_{i=1}^n \mathbf{1}\!\left[(u_i,v_i) \in B_{ab}\right],
\end{equation}
so that $\sum_{a,b} H_{ab}\Delta^2 = 1$. A valid estimate must satisfy: (i) non-negativity, (ii) integration to one, and (iii) uniform marginals.

For real datasets, we estimate the marginals separately: empirical one-dimensional CDFs fitted on the training split map $X \mapsto U$, and their inverses are used only when a downstream task needs samples or imputations in data space (VaR backtesting, Appendix~\ref{app:var}, and imputation). \ours{} itself models the copula $c$; the copula-space benchmarks share the training-fitted marginal transform, while data-space sampling additionally uses its inverse.

\subsection{Amortized Denoising Edge Estimator}
\label{subsec:vdc_denoiser}

Our edge estimator is a two-dimensional (2D) U-Net \citep{ronneberger2015unet}, a convolutional encoder--decoder with skip connections, that takes a possibly corrupted histogram $\tilde{H}$, coordinate channels, and a constant $\log n$ sample-size channel as input. For one-indexed cells $(a,b)$, the two coordinate feature maps are constant over the batch and equal to
\[
\operatorname{coord}_1(a,b)=\Phi^{-1}\!\left(\frac{a-\tfrac12}{m}\right),
\qquad
\operatorname{coord}_2(a,b)=\Phi^{-1}\!\left(\frac{b-\tfrac12}{m}\right),
\]
where $\Phi^{-1}$ is the standard-normal quantile function. Let $\mathrm{coords}=(\operatorname{coord}_1,\operatorname{coord}_2)$, let $f_\theta$ denote the U-Net with parameters $\theta$, and let $Z\in\RR^{m\times m}$ denote its output log-density logits. Define the elementwise clipping operation by $\operatorname{clip}(Z,-20,20)=\max(-20,\min(Z,20))$, and let $\Pi_{\mathrm{IPFP}}$ denote the copula projection specified in Section~\ref{subsec:ipfp}. The deployed one-shot map is
\begin{equation}
\label{eq:edge_map}
\begin{aligned}
Z &= f_\theta(\tilde{H}, \mathrm{coords}, \log n;t), \\
D_{\text{raw}} &= \exp\{\operatorname{clip}(Z,-20,20)\}, \qquad
\hat{D} = \Pi_{\text{IPFP}}(D_{\text{raw}}).
\end{aligned}
\end{equation}
In Equation~\eqref{eq:edge_map}, $D_{\text{raw}}\in\RR_{>0}^{m\times m}$ is the positive raw density grid and $\hat D$ is its projected density; \emph{one-shot} means one network evaluation per edge, with multiple edges sharing a batched forward pass. The sample size enters as a constant input channel equal to $\log n$, and every U-Net block is conditioned on the noise level $t$ introduced below through FiLM (feature-wise linear modulation) \citep{perez2018film}, a per-channel affine modulation computed from an embedding of that level. The network is trained on synthetic bivariate copulas spanning Gaussian, Student-$t$, Archimedean families (Clayton, Gumbel, Frank, Joe), two-parameter Archimedean BB1/BB7 families, rotations, mixtures, and complex synthetic patterns (exact parameter distributions in Appendix Table~\ref{tab:benchmark_spec}). The original Joe, BB1, BB7 and conditional training generators contained errors that made sampled inputs disagree with their target densities. Appendix~\ref{app:training_data} documents their reconstructed prevalence. Main results evaluate the original frozen checkpoint on validated data; Appendix~\ref{app:retrained} reports exploratory retraining on corrected data. During training, a noise level $t\sim\mathrm{Unif}(0,1)$ sets a weight $\alpha=0.65\,t$ that interpolates the uncorrupted empirical histogram toward the uniform density grid $H_{\mathrm{unif}}$, whose entries all equal one: $\tilde{H}=(1-\alpha)H+\alpha H_{\mathrm{unif}}$. At test time, $\tilde H=H$ and $t=0$. This design lets one checkpoint handle histograms across a broad range of finite-sample noise levels while keeping deployment simple.

The corruption scheme is a design choice. On common synthetic and real-data inputs, the three retained alternatives to uniform mixing (direct histograms, Gaussian grid noise and multinomial count resampling) differ from the deployed model by at most $0.014$ bits/dim in end-to-end held-out vine likelihood and no scheme is best everywhere: uniform mixing is best on POWER and MINIBOONE, multinomial resampling on GAS and HEPMASS, and the direct variant trained without corruption is weakest on the real data among the primary checkpoints (Appendix~\ref{app:corruption_ablation}).

Let $\mathcal{L}_{\mathrm{CE}}$, $\mathcal{L}_{\mathrm{ISE}}$, $\mathcal{L}_{\mathrm{tail}}$, and $\mathcal{L}_{\mathrm{marg}}$ denote the cross-entropy, integrated-squared-error, tail, and marginal loss terms defined in Appendix~\ref{app:implementation}, and let the corresponding $\lambda$ coefficients denote their weights. We train the edge estimator with
\begin{equation}
\mathcal{L} = \mathcal{L}_{\text{CE}} + \lambda_{\text{ISE}} \mathcal{L}_{\text{ISE}} + \lambda_{\text{tail}} \mathcal{L}_{\text{tail}} + \lambda_{\text{marg}} \mathcal{L}_{\text{marg}}.
\end{equation}
The deployed weights are $\lambda_{\text{ISE}}=0$, $\lambda_{\text{tail}}=0.2$, and $\lambda_{\text{marg}}=0.02$, fixed for the deployed checkpoint.

\subsection{Copula Projection via IPFP}
\label{subsec:ipfp}

The raw network output $D_{\text{raw}}$ is positive but generally has the wrong total mass and non-uniform grid marginals. We first normalize its total mass, writing $Q=D_{\text{raw}}/(\Delta^2\sum_{a,b}D^{\text{raw}}_{ab})$. Iterative Proportional Fitting (IPFP) \citep{sinkhorn1964relationship,sinkhorn1967knopp,franklin1989scaling} then alternates row and column normalization. For a current density grid $D=(D_{ab})$, define its row mass $\mu_a^{\text{row}}=\Delta^2\sum_bD_{ab}$; one row update sets $D_{ab}\leftarrow D_{ab}(1/m)/\mu_a^{\text{row}}$ for every $a$, followed by the analogous column update. For normalized density grids $P,Q$, we use the discrete-grid Kullback--Leibler (KL) divergence $\KL(P\|Q)=\Delta^2\sum_{a,b}P_{ab}\log(P_{ab}/Q_{ab})$.

\begin{proposition}[Sinkhorn/IPFP projection]
\label{prop:ipfp}
Given the normalized, strictly positive grid $Q$, IPFP converges to the unique limit $\hat{D}_{\infty}$ minimizing $\KL(\hat{D}_{\infty}\|Q)$ subject to uniform marginal constraints.
\end{proposition}

This is the classical Sinkhorn/IPFP projection result in the strictly positive case \citep{sinkhorn1964relationship,sinkhorn1967knopp,franklin1989scaling,csiszar1975divergence}. We write $\hat{D}^{(K)}$ for the output after $K$ row/column updates and $\hat{D}_{\infty}$ for the exact limit. The deployed setting is $K{=}30$, with maximum marginal-density error around $2\times10^{-4}$ on four synthetic test edges and $8\times10^{-3}$ on strongly dependent GAS edges. UCI held-out likelihood and probability-integral-transform calibration change little when $K$ rises from 15 to 100 (Appendix~\ref{app:ipfp_iterations}). The exact projection can only reduce KL error to the true grid copula (Proposition~\ref{prop:pythagorean}, Appendix~\ref{app:guarantees}); this statement does not apply directly to interpolated likelihoods.

\subsection{H-Function Computation and Vine Assembly}
\label{subsec:hfunc}

With the projected density $\hat{D}$, we compute h-functions as cumulative sums over grid rows and columns and cache them for $O(1)$ lookup. The projected grids supply the edge densities and cached conditional transformations used in iterative pair-copula construction under the fitted numerical model. Edge densities use bilinear interpolation between cell centers, while h-functions interpolate cumulative sums. Their derivatives differ from the edge densities by $1$--$3\%$ on average on five tested synthetic copulas. Each edge density integrates to one, but the assembled vine is normalized only approximately: the measured discrepancy is at most $2.3\times10^{-4}$ on tested three-variable UCI sub-vines, not a bound for higher dimensions (Appendix~\ref{app:guarantees}). This assembly is still a simplified-vine procedure, so the learned edge represents a pooled conditional pair-copula rather than a pair-copula that changes with each realized conditioning value. The implementation supports regular vines selected with Di{\ss}mann's algorithm and D-vines (path-structured vines; Appendix~\ref{app:related_work}); unless stated otherwise our experiments use D-vines with a greedy $|\tau|$ order. In the main D-vine scaling runs, histograms, network inference, and IPFP projection are batched on the GPU; h-functions then update the pseudo-observations for the next tree. Each forward pass handles up to $256$ edges from one tree; the regular-vine path evaluates edges one at a time. Frozen weights replace per-edge statistical fitting with shared neural inference.

Algorithms~\ref{alg:vine_fit}, \ref{alg:vine_ll}, and~\ref{alg:vine_sampling} give pseudocode for fitting, likelihood evaluation, and sampling. Per edge, building the histogram costs $O(n)$ and the network and projection cost is constant in $n$, whereas nonparametric fits such as TLL grow more expensive with $n$ (Tables~\ref{tab:vine_nll} and~\ref{tab:vine_nll_50k}).

\section{Information Estimation}
\label{sec:information}

The fitted copula densities also provide estimates of mutual information and total correlation. Both quantities can be computed from the same edge densities used for likelihood evaluation, without training a separate variational estimator. For the bivariate copula density $c$, define its differential entropy by $H(c)=-\int_{[0,1]^2}c(u,v)\log c(u,v)\,du\,dv$; this $H(c)$ is distinct from the histogram $H$ of Section~\ref{sec:method}. Mutual information between $X$ and $Y$ equals the negative copula entropy \citep{davy2003copulas,calsaverini2009copula,ma2011mutual,safaai2018information}:
\begin{equation}
I(X;Y) = \EE[\log c(U,V)] = -H(c).
\end{equation}
The implementation estimates bivariate MI by integrating the normalized density grid, with $\bar D_{ab}=\hat D_{ab}/(\Delta^2\sum_{j,k}\hat D_{jk})$:
\begin{equation}
\hat{I}(X;Y) = \Delta^2 \sum_{a,b} \bar D_{ab}\log \bar D_{ab}.
\end{equation}
For an $m \times m$ piecewise-constant density with exactly uniform grid marginals, the largest representable MI is $\log m$ ($\approx 4.16$ nats at $m{=}64$); the same ceiling holds for the deployed output up to its residual marginal error. Tail dependence at scales finer than $1/m$ is not resolved; the dependencies in our benchmarks sit well below this ceiling.

Define total correlation as $\mathrm{TC}(X_1,\ldots,X_d)\defeq\EE[\log c(U_1,\ldots,U_d)]$, the KL divergence from the joint distribution to the product of its marginals (Appendix~\ref{app:theory}). Let $\mathcal E$ be the set of all vine edges and, for $e=(j,k|D)\in\mathcal E$, set $(U_e,V_e)=(U_{j|D},U_{k|D})$ and let $c_e$ be that edge's pair-copula density. Equation~\eqref{eq:vine_factorization} then gives
\begin{equation}
\label{eq:tc_decomposition}
\mathrm{TC}(X_1, \ldots, X_d) = \sum_{e \in \mathcal E} \EE[\log c_e(U_e, V_e)].
\end{equation}
For a fitted vine $\hat{c}(u) = \prod_e \hat{c}_e(u_{j \mid D}, u_{k \mid D})$, we estimate held-out TC by summing per-edge empirical mean log-density terms. True population edge terms are non-negative conditional MIs; fitted held-out contributions can be negative because of sampling error or model misspecification. Figure~\ref{fig:tc_decomposition} validates this decomposition against analytic edge MI and by tree knockout on real data.

\section{Experiments}
\label{sec:experiments}

We evaluate bivariate edge fitting, full-vine density estimation on UCI data and at high dimension, information estimation, and downstream conditional tasks (Sections~\ref{subsec:bivariate}--\ref{subsec:downstream}). Unless noted otherwise, \ours{} fits a D-vine whose variable order is chosen greedily to maximize adjacent pairwise $|\tau|$, whereas the pyvine baselines select regular vines with Di{\ss}mann's algorithm (the \texttt{pyvinecopulib} default), so full-vine comparisons are between complete pipelines rather than matched structures; matching the structures leaves every ranking and speedup we checked unchanged (UCI, S\&P100, and $d{=}100$ scaling; Appendices~\ref{app:uci_matched}, \ref{app:highdim_scaling}, and~\ref{app:sp100_density}). Except for the separate retraining study in Appendix~\ref{app:retrained} and the ablations, all \ours{} fits use grid size $m=64$ and $K{=}30$ IPFP iterations ($K{=}15$ in the scaling, fit-timing, S\&P100 density, sampling, and deep-tree runs).\footnote{Fit times include structure selection and exclude one-time pretraining, checkpoint loading and evaluation. The main scaling comparison uses one H100 GPU for \ours{} and $8$- or $16$-thread CPU TLL on the same node. Warm-up conventions, hardware and CPU-only controls are in Appendix~\ref{app:reproducibility}.}

\subsection{Bivariate Copula Estimation}
\label{subsec:bivariate}

Table~\ref{tab:bivariate} evaluates the pretrained estimator on nine fixed cases from families included in pretraining against raw histograms, kernel density estimation (KDE), and pyvine with parametric and transformation local likelihood (TLL) estimation.

\begin{table}[!t]
\centering
\caption{\textbf{Bivariate estimation on families included in pretraining.} Nine fixed cases listed in Appendix Table~\ref{tab:benchmark_spec}, from family classes represented in pretraining, each fitted on five fresh pyvinecopulib samples of size $n{=}2000$ whose seeds were not used for checkpoint selection ($45$ fits per method). ISE = integrated squared error of cell-averaged densities against the exact cell averages of the true copula; KL = held-out log-likelihood loss $\EE_c[\log c-\log\hat c]$ in nats on $20{,}000$ fresh samples per fit; $|\Delta\tau|$ = Kendall's $\tau$ error; $|\Delta\lambda_U|$ = upper-tail dependence error at $q{\approx}0.95$ (top-right $3\times3$ grid mass, computed identically for truth and fits). Time: per-fit time, including grid evaluation and projection, from the original timing run (\ours{}: one H100 GPU; pyvine: one CPU thread; histogram and KDE: up to $16$ OpenMP/MKL threads). Means over the $45$ fits; Appendix Table~\ref{tab:bivariate_dispersion_appendix} gives the spread across cases, and Table~\ref{tab:bivariate_replicates} a separate $800$-replication paired study. Evaluation provenance is summarized in Appendix~\ref{app:evaluation_provenance}.}
\label{tab:bivariate}
\footnotesize
\IfFileExists{tables/tab_bivariate.tex}{
\makebox[\columnwidth][c]{\scalebox{0.92}{\begin{tabular}{lccccc}
\toprule
Method & ISE $\downarrow$ & KL $\downarrow$ & $|\Delta\tau|$ $\downarrow$ & $|\Delta\lambda_U|$ $\downarrow$ & Time (ms) $\downarrow$ \\
\midrule
Histogram & 2.07e+00 & 1.3208 & \textbf{0.005} & 0.025 & \textbf{2.2} \\
KDE & 3.60e-02 & 0.0162 & 0.014 & 0.026 & 85.5 \\
pyvine-param & \textbf{3.41e-03} & \textbf{0.0004} & 0.006 & \textbf{0.008} & 591.0 \\
pyvine-TLL & 4.31e-02 & 0.0135 & 0.016 & 0.025 & 16.2 \\
VDC (one-shot) & 1.72e-02 & 0.0061 & 0.007 & 0.010 & 6.0 \\
\bottomrule
\end{tabular}
}}
}{
\fbox{\parbox{0.95\columnwidth}{\centering\textbf{[Bivariate table missing]}\\Run the artifact generation script to produce the bivariate table.}}
}
\end{table}

On this suite \ours{} is the most accurate flexible estimator in density and tail error at a few milliseconds per copula: relative to pyvine-TLL it lowers ISE by $2.5\times$ and held-out KL by $2.2\times$ (lower KL on $41$ of $45$ fits), Kendall-$\tau$ error by $2.3\times$, and tail error by $2.6\times$. Parametric family selection is more accurate on every quality metric here, as expected when every test family is in its library, but it costs about $100\times$ more per edge and is on average the least accurate method on the complex X, ring, and double-banana shapes (Appendix Table~\ref{tab:bivariate_complex_appendix}). To separate amortization quality from prior match, we also evaluate the identical frozen checkpoint on copula families entirely absent from the training zoo (BB6, BB8, Tawn, and Plackett): without any refitting, it matches per-case pyvine-TLL refits in held-out likelihood and roughly halves MI error against Monte Carlo reference values ($0.017$ vs.\ $0.040$ nats on average; Appendix Table~\ref{tab:ood_families}). Appendix Table~\ref{tab:neural_copula_appendix} adds a focused comparison against ACNet, a neural pair-copula estimator retrained per edge ($\sim\!10^3$\,s per pair versus milliseconds).

\subsection{Full-Vine Density Estimation on UCI}
\label{subsec:uci}

We next fit full vines on the standard UCI density benchmarks, with the preprocessing and train/test splits from Masked Autoregressive Flow (MAF) \citep{papamakarios2017maf}: POWER ($d{=}6$), GAS ($d{=}8$), HEPMASS ($d{=}21$), and MINIBOONE ($d{=}43$). All methods receive the same empirical marginal transform and the same $n_{\mathrm{train}}{=}5{,}000$ training rows (a $50{,}000$-row variant is in Appendix~\ref{app:uci_density}), and we report held-out copula-space negative log-likelihood (NLL) with bootstrap confidence intervals (CIs) (Appendix Table~\ref{tab:vine_nll}). All UCI results reported here were recomputed on verified MAF splits; Appendix~\ref{app:evaluation_provenance} documents the corrected evaluation inputs. On the real data, \ours{} attains the best NLL on the higher-dimensional HEPMASS and MINIBOONE ($-0.237$ and $-0.969$ bits/dim versus $-0.201$ and $-0.744$ for pyvine-TLL) while fitting about $8\times$ faster than pyvine-TLL and more than $100\times$ faster than parametric selection. On the low-dimensional POWER and the strongly dependent GAS, pyvine-TLL is better ($-0.182$ vs.\ $-0.141$; $-1.705$ vs.\ $-1.477$). With $n_{\mathrm{train}}{=}50{,}000$ (Appendix Table~\ref{tab:vine_nll_50k}) these gaps widen, pyvine-TLL also overtakes \ours{} on HEPMASS (by $0.018$ bits/dim), and the RealNVP flow~\citep{dinh2017realnvp} is best on all four datasets; \ours{} stays ahead of pyvine-TLL on MINIBOONE, and of parametric selection and the Gaussian copula on every dataset at both sizes. GAS contains near-deterministic sensor pairs (maximum pairwise $|\tau|=0.89$; Appendix Table~\ref{tab:uci_dependence_diag}), the regime in which the $64\times64$ grid is least accurate (Section~\ref{sec:conclusion}).

In two stress tests that violate the simplifying assumption (Appendix~\ref{app:nonsimplified_stress}), \ours{} is the best pooled estimator in a single draw each (excess NLL $0.279$ and $0.076$; other pooled estimators at least $0.291$ and $0.088$).

\subsection{High-Dimensional Vine Scaling}
\label{subsec:highdim}

\begin{figure*}[!t]
    \centering
    \includegraphics[width=\textwidth]{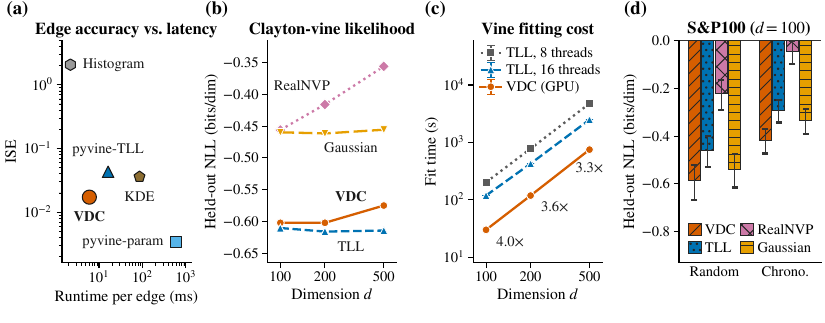}
    \caption{\textbf{Density and vine-fitting cost.} (a) Accuracy--latency on the bivariate suite of Table~\ref{tab:bivariate}: \ours{} has lower ISE than histogram, KDE and pyvine-TLL at about $6$\,ms per edge on GPU; parametric selection is more accurate at about $100\times$ the cost on one CPU thread (separate timing run, Table~\ref{tab:bivariate}). (b) Held-out NLL on the Clayton D-vine benchmark ($n_{\mathrm{train}}{=}20{,}000$, $n_{\mathrm{test}}{=}5{,}000$; Appendix~\ref{app:highdim_scaling}); means over seeds $\{7,17,42\}$; RealNVP and the Gaussian copula are quality references, and at $d{=}500$ the TLL value averages the two finite runs. (c) Fit time on one node (\ours{}: one H100 GPU and $16$ CPU threads; TLL: $8$ or $16$ threads; means and sample standard deviations over seeds $\{7,17,42\}$, error bars smaller than markers); numbers give the speedup over $16$-thread TLL. (d) Held-out NLL on S\&P100 daily returns ($d{=}100$) with $95\%$ i.i.d. bootstrap CIs over test days under random and chronological splits; lower is better. TLL denotes pyvine-TLL.}
    \label{fig:density_runtime_summary}
\end{figure*}

Figure~\ref{fig:density_runtime_summary} compares density quality and fitting cost; panel (a) summarizes the bivariate suite of Section~\ref{subsec:bivariate}. Panel (b) evaluates full-vine likelihood on a first-order Markov Clayton benchmark, with dependence only in tree 1 (the all-trees stress test is separate). Every dependent edge uses the same family; \ours{} stays close to pyvine-TLL, and both vines outperform RealNVP and Gaussian copula. Panel (c) compares GPU \ours{} with CPU TLL: speedups against $16$-thread TLL are $3.96\times$, $3.60\times$, and $3.32\times$ at $d{=}100,200,500$ ($6.3$--$6.7\times$ against $8$ threads). At $d{=}500$, \ours{} takes $749.0\pm0.5$\,s versus $2488.9\pm4.7$\,s for $16$-thread TLL. Optimized execution preserves every fitted edge and evaluated likelihood bitwise across all nine fits (Appendix Table~\ref{tab:fast_fit_timing}). At $d{=}500$, \ours{} trails TLL by $0.039$ bits/dim in held-out NLL (Appendix Table~\ref{tab:highdim_scaling_appendix}): the speed advantage comes at a modest likelihood cost. Appendix~\ref{app:mixed_family_scaling} gives the heterogeneous Gaussian/Clayton variant, where parametric family selection is $35$--$37\times$ slower.

Repeated use amortizes the $9.6$-hour pretraining: the time saved per fit grows from $88.9$ to $308.4$ to $1739.9$\,s at $d{=}100,200,500$ while the speedup ratio falls slightly (break-even: Appendix~\ref{par:break_even}).

\paragraph{Real S\&P100 returns.} On real S\&P100 daily returns ($d{=}100$, $n_{\mathrm{train}}{=}1356$), \ours{} attains the best held-out likelihood of the compared methods under both random and chronological splits (Figure~\ref{fig:density_runtime_summary}(d)); parametric family selection was not run on this benchmark. Under the chronological split it reaches $-0.418$ bits/dim versus $-0.293$ for pyvine-TLL, $-0.337$ for the Gaussian copula, and $-0.047$ for a small-budget RealNVP; the gain over TLL ($0.126$ bits/dim) holds on $87\%$ of test days and its $95\%$ CI stays within $[0.101, 0.144]$ under moving-block bootstraps with blocks of up to $40$ days (Appendix Table~\ref{tab:sp100_chrono}). TLL's poorer test likelihood than the Gaussian copula may reflect overfitting. VDC's advantage is consistent with regularization from pretraining, although this comparison does not isolate that effect.

\subsection{Information Estimation}
\label{subsec:information_results}

\begin{figure*}[!t]
    \centering
    \includegraphics[width=\textwidth]{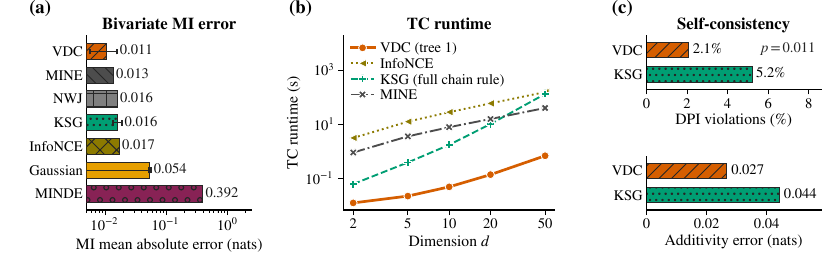}
    \caption{\textbf{Information-estimation results.} (a) Bivariate MI mean absolute error over the eight synthetic cases of Table~\ref{tab:mi_appendix} with numerically integrated ground truth (Joe, whose samples came from a defective sampler, is excluded); whiskers are $\pm1$ standard deviation across three seeds where available (MINE, InfoNCE, NWJ and MINDE have one seed). (b) Median total-correlation runtime versus $d$ for a Gaussian AR(1) family ($\rho{=}0.7$, $n{=}5000$, three trials for each of seeds $\{7,17,42\}$, one H200 GPU and $8$ CPU cores per run); \ours{} uses a D-vine truncated after its first tree and KSG is the full chain-rule estimator (matched adjacent-pair control in Appendix~\ref{par:tc_tree1_control}). (c) MI self-consistency: DPI violation rate over $480$ trials ($160$ base datasets at three noise levels; Fisher exact $p{=}0.011$ over base datasets) and mean additivity error over $40$ trials; lower is better (protocol in Appendix~\ref{app:self_consistency}).}
    \label{fig:info_results_composite}
\end{figure*}

\begin{figure*}[!t]
    \centering
    \includegraphics[width=\textwidth]{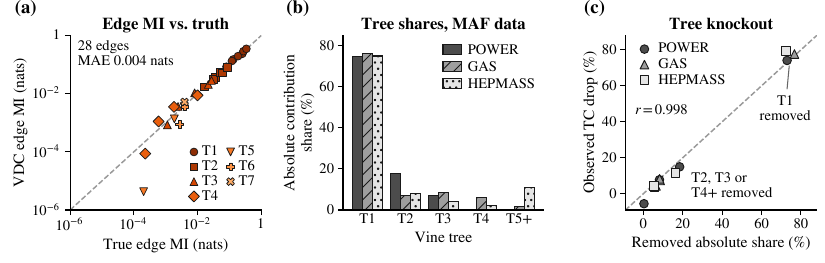}
    \caption{\textbf{TC decomposition.} (a) Estimated versus analytic MI for all $28$ edges of the generating-order D-vine of a $d{=}8$ Gaussian-factor copula ($n_{\mathrm{train}}{=}10{,}000$; markers give the tree). (b) Tree-level decomposition of held-out TC on the MAF POWER, GAS and HEPMASS data ($n_{\mathrm{train}}{=}10{,}000$): each tree's share of summed absolute edge contributions (trees $5$ and above pooled as T5+). (c) Tree knockout on separately fitted vines for the same datasets: absolute-contribution share of the removed tree group (T1, T2, T3 or T4+) versus the drop in \ours{} TC estimated from $2{,}000$ samples of the knocked-out vine (mean of three repeats; inverse-h interpolation as in Appendix~\ref{app:sampling_quality}), Pearson $r{=}0.998$. Absolute-contribution shares are defined in Appendix~\ref{par:absolute_tc_shares}.}
    \label{fig:tc_decomposition}
\end{figure*}

Figure~\ref{fig:info_results_composite} evaluates the density-based information estimates---bivariate MI accuracy (a), end-to-end TC runtime (b) and self-consistency (c)---and Figure~\ref{fig:tc_decomposition} the TC decomposition itself. We compare against the $k$-nearest-neighbor estimator KSG, the variational neural estimators MINE, InfoNCE and NWJ, the diffusion-based MINDE, and a closed-form Gaussian-copula baseline (Appendix~\ref{app:baselines}). \ours{} has low MI error on synthetic copulas with known ground truth and on adjacent pairs of Gaussian first-order autoregressive [AR(1)] processes (Appendix Figure~\ref{fig:dimension_appendix}(a)), and in a fixed-checkpoint non-Gaussian Clayton-chain study it is the most accurate tested estimator for sample sizes of $10{,}000$ or more, with relative MI error $2.8\%$ at $n{=}30{,}000$ and $2.7\%$ at $n{=}100{,}000$ but $104.5\%$ at $n{=}1000$ (Appendix~\ref{app:mi_sample_size}).

In self-consistency tests (Appendix~\ref{app:self_consistency}; $480$ DPI trials per estimator on $160$ independent base datasets at three noise levels), $10$ of $160$ base datasets yield a violation for \ours{} versus $25$ for KSG (Fisher exact and cluster permutation $p=0.011$), with additivity error $0.027$ versus $0.044$ nats (Figure~\ref{fig:info_results_composite}(c); Table~\ref{tab:self_consistency_scaleup}); the smaller $60$-trial protocol (Appendix Table~\ref{tab:self_consistency}) did not detect a significant DPI difference. We estimate TC by summing held-out mean edge log-densities (Equation~\eqref{eq:tc_decomposition}). On the first-order Markov AR(1) benchmark the D-vine is truncated after its first tree (exact here), and at $d=50$ \ours{} takes $0.71$\,s versus $41$--$161$\,s for neural and $k$-NN baselines that use no such structure (Figure~\ref{fig:info_results_composite}(b)). Given the same first-tree decomposition, no clear accuracy difference from KSG was detected: on $50$ fresh $d{=}50$ datasets the mean absolute TC errors are $0.187$ nats for \ours{} and $0.201$ for adjacent-pair KSG (paired difference $-0.013$, $95\%$ CI $[-0.070, 0.045]$), and \ours{} is $4.7\times$ faster on one node (H100 GPU and one CPU thread versus one CPU thread; Appendix~\ref{par:tc_tree1_control}).

\paragraph{TC decomposition.} In Figure~\ref{fig:tc_decomposition}, edge MI closely tracks the analytic ground truth on all $28$ edges of a $d{=}8$ Gaussian-factor vine (a; mean absolute error $0.004$ nats). On the MAF data the first tree accounts for roughly three quarters of the summed absolute edge contributions (b), and the TC changes in samples simulated from the knocked-out vine approximately track the removed tree share (c; Pearson $r=0.998$).

\subsection{Downstream Conditional Tasks}
\label{subsec:downstream}

Conditional sampling requires repeated h-function inversions, making its accuracy sensitive to errors accumulated across the vine (protocols in Appendix~\ref{app:downstream}). Simulated S\&P100 samples closely reproduce the held-out Kendall-$\tau$ matrix relative to the real-data bootstrap reference (root-mean-square error $0.052$ versus a reference of $0.039$), compared with $0.192$ for a per-dataset pyvine-TLL refit (Appendix~\ref{app:sampling_quality}, with sampling times). The conditional tasks reveal limitations in imputation accuracy and tail calibration: on the Combined Cycle Power Plant (CCPP) data \citep{tufekci2014ccpp}, inverse-Rosenblatt imputation trails $k$-NN imputation by about $20\%$ in root-mean-square error (RMSE) (Appendix Table~\ref{tab:imputation_exact}), and rolling VaR backtesting on S\&P100 overshoots the nominal $1\%$ level (observed $1.9\%$; Figure~\ref{fig:var_calibration}, Appendix~\ref{app:var}). These tasks test both the fitted conditional model and the procedure used to turn its samples into predictions.

\section{Discussion and Conclusion}
\label{sec:conclusion}

\ours{} fits classical simplified vines by reusing one pretrained bivariate estimator at every edge. IPFP enforces the grid copula constraints, and batched GPU inference reduces fitting time relative to multithreaded CPU baselines. The fitted vine retains explicit pair-copulas and h-functions for likelihood evaluation, sampling, and edge- and tree-level information analysis. This structure distinguishes \ours{} from monolithic flows and full-dimensional diffusion models.

The results suggest that amortized estimation is especially useful when many edges must be fitted from limited data. \ours{} leads the compared estimators on S\&P100 under both splits and improves on the strongest tested RealNVP variants on Clayton vines (Table~\ref{tab:flow_sensitivity_appendix}). Its advantage over the small-budget RealNVP also holds on raw S\&P100 returns with shared Student-$t$ marginals (Table~\ref{tab:raw_composition}). Pretraining may regularize estimation in these settings, although the comparisons do not isolate that mechanism. The UCI results show that the advantage is not universal: at moderate sample sizes \ours{} leads on HEPMASS and MINIBOONE, while TLL leads on POWER and GAS; with more data, RealNVP leads on all four. Choice of estimator therefore depends on sample size, dependence structure and fitting budget.

These results also suggest concrete ways to improve the estimator. Finer or adaptive grids may help resolve strongly dependent edges: even exact cell averages on the $64\times64$ grid, evaluated as a piecewise-constant density, lose about $0.007$ nats of held-out log-likelihood per edge for typical pretraining families and far more for near-deterministic pairs (Appendix~\ref{app:grid_floor}). This is consistent with TLL's advantage at large sample sizes and on GAS, but does not establish its cause. Parametric selection remains more accurate when the true family is in its library, and VaR calibration and imputation need improvement. The original checkpoint was trained with generator defects and evaluated on correct data (Section~\ref{subsec:vdc_denoiser}); Appendix~\ref{app:retrained} examines retraining on corrected data. Extending the approach to conditional copulas and mixed data, alongside improved calibration, would broaden its use beyond continuous simplified vines.

\ifdefined\arxivmode
\section*{Acknowledgements}
This work has been made possible in part by a gift from the Chan Zuckerberg Initiative Foundation to establish the Kempner Institute for the Study of Natural and Artificial Intelligence at Harvard University. The author declares no competing interests.
\else
\begin{ack}
This work has been made possible in part by a gift from the Chan Zuckerberg Initiative Foundation to establish the Kempner Institute for the Study of Natural and Artificial Intelligence at Harvard University. The author declares no competing interests.
\end{ack}
\fi

\ifdefined\arxivmode
\clearpage
\fi
\setlength{\bibsep}{1.5pt}

\bibliographystyle{plainnat}
\bibliography{references}

\ifdefined\NOAPPENDIX
\else
\appendix
\setcounter{figure}{0}
\setcounter{table}{0}
\setcounter{algorithm}{0}
\renewcommand{\thefigure}{A\arabic{figure}}
\renewcommand{\thetable}{A\arabic{table}}
\renewcommand{\thealgorithm}{A\arabic{algorithm}}

\section*{Broader Impact and Ethics}
This work studies density-estimation methodology rather than a deployment system. Potential positive uses include faster and more transparent dependence analysis in scientific, financial, and monitoring workflows, where explicit MI/TC decompositions can help audit which variables contribute to dependence. The main risk is over-trusting fitted dependence or conditional forecasts in high-stakes settings. In application areas such as finance or risk monitoring, model misspecification and poor calibration can create real harm, so downstream validation remains essential.

\section*{Appendix Roadmap and Claims--Evidence Map}
\label{app:roadmap}

The appendix is organized as follows: Appendix~\ref{app:related_work} expands the related work; Appendix~\ref{app:theory} collects theoretical background; Appendix~\ref{app:implementation} gives architecture, training, reproducibility, and baseline details; Appendix~\ref{app:downstream} contains experimental details and additional results; Appendix~\ref{app:algos} lists algorithms; Appendix~\ref{app:ablations} reports ablations; and Appendix~\ref{app:retrained} evaluates exploratory estimators retrained on corrected data. Table~\ref{tab:claims_evidence} maps the main claims of the paper to their supporting evidence.

\begin{table}[h]
    \centering
    \caption{\textbf{Claims--evidence map.} The main empirical claims with the artifacts that support them; negative results are listed on the same footing.}
    \label{tab:claims_evidence}
    \footnotesize
    \begin{tabular}{p{0.46\columnwidth}p{0.46\columnwidth}}
    \toprule
    Claim & Evidence \\
    \midrule
    Amortized edge estimation roughly halves density error relative to TLL on families included in pretraining; parametric selection is more accurate on in-library families & Table~\ref{tab:bivariate}; Figure~\ref{fig:density_runtime_summary}(a); dispersion in Table~\ref{tab:bivariate_dispersion_appendix}; harder families in Table~\ref{tab:bivariate_complex_appendix} \\
    The frozen estimator generalizes to families outside the training zoo & Table~\ref{tab:ood_families} \\
    IPFP converges to a valid pair-copula; the deployed iterate has controlled error and the projection is essential & Proposition~\ref{prop:ipfp} (Section~\ref{subsec:ipfp}) and Proposition~\ref{prop:pythagorean} (Appendix~\ref{app:guarantees}); Tables~\ref{tab:ipfp_iter_ablation}, \ref{tab:ipfp_uci_sensitivity}, \ref{tab:noipfp_ablation} \\
    Full-vine likelihood on the MAF UCI benchmarks: best on HEPMASS and MINIBOONE at $n_{\mathrm{train}}{=}5{,}000$, behind TLL on POWER and GAS; at $50{,}000$ rows RealNVP leads on all four and \ours{} leads TLL only on MINIBOONE & Tables~\ref{tab:vine_nll}, \ref{tab:vine_nll_50k}; matched structures in Table~\ref{tab:uci_matched}; dependence diagnostic in Table~\ref{tab:uci_dependence_diag} \\
    Faster high-dimensional fitting with amortized, batched GPU inference; at $d{=}500$ \ours{} trails TLL by $0.039$ bits/dim in held-out NLL & Figure~\ref{fig:density_runtime_summary}(c); Tables~\ref{tab:fast_fit_timing}, \ref{tab:highdim_scaling_appendix}, \ref{tab:mixed_family_scaling_appendix} \\
    Best held-out NLL among the compared methods on real S\&P100 returns (random and chronological splits; parametric selection not run) & Tables~\ref{tab:sp100_density_appendix}, \ref{tab:sp100_chrono} \\
    Composed with separate marginals, \ours{} beats a small-budget monolithic raw-space flow on S\&P100 & Table~\ref{tab:raw_composition} \\
    Repeated-fitting advantage over per-edge neural copulas & Table~\ref{tab:neural_copula_appendix} \\
    Accurate MI/TC with edge/tree decomposition; given the same first-tree decomposition, no clear TC accuracy difference from KSG was detected & Figures~\ref{fig:info_results_composite}, \ref{fig:tc_decomposition}, \ref{fig:dimension_appendix}, \ref{fig:mi_sample_size_clayton_appendix}; Tables~\ref{tab:mi_appendix}, \ref{tab:theory_mi_methods_appendix}, \ref{tab:tc_scaling_appendix}; structural control in Appendix~\ref{par:tc_tree1_control} \\
    Fewer DPI violations than KSG (significant over $160$ independent base datasets) and lower additivity error & Tables~\ref{tab:self_consistency}, \ref{tab:self_consistency_scaleup}; protocol in Appendix~\ref{app:self_consistency} \\
    Fitted vines produce realistic unconditional samples & Table~\ref{tab:sampling_quality} (Appendix~\ref{app:sampling_quality}) \\
    Sensitivity to the simplifying assumption & Table~\ref{tab:nonsimplified_stress}; Figure~\ref{fig:nonsimplified_stress} \\
    Tail calibration and imputation accuracy remain limited (VaR backtesting; CCPP imputation) & Tables~\ref{tab:imputation}, \ref{tab:imputation_exact}; Figure~\ref{fig:var_calibration} \\
    No corruption scheme is best everywhere; the direct variant is weakest among the primary checkpoints on the real data & Tables~\ref{tab:corruption_variants}, \ref{tab:corruption_edges} \\
    Grid-resolution references indicate larger representation error for near-deterministic pairs & Table~\ref{tab:grid_floor} \\
    The structured vine beats the strongest RealNVP variants on non-Gaussian Clayton vines & Table~\ref{tab:flow_sensitivity_appendix} \\
    Training-generator defects; retraining on corrected data is exploratory & Appendix~\ref{app:training_data}; Tables~\ref{tab:retrained_bivariate}, \ref{tab:retrained_uci} \\
    \bottomrule
    \end{tabular}
\end{table}

\section{Additional Related Work}
\label{app:related_work}

\subsection{Vine Copulas and Structure Learning}

Vines provide a flexible factorization of multivariate copulas into pair-copulas \citep{bedford2002vines, aas2009pair}; comprehensive treatments are given in \citet{joe2014dependence} and \citet{czado2019analyzing}. The key idea is that any $d$-dimensional copula density can be decomposed into $\binom{d}{2}$ bivariate building blocks arranged in a sequence of $d-1$ trees. Different orderings lead to canonical vines (C-vines), path-structured drawable vines (D-vines), or general regular vines (R-vines).

Practical pipelines typically learn vine structure via greedy maximum spanning tree heuristics \citep{dissmann2013selecting}. At each tree level, they choose edges with strongest dependence (often $|\tau|$) subject to regular-vine constraints.

\subsection{Pair-Copula Estimation Methods}

Once the vine structure is fixed, each edge requires fitting a bivariate copula. \textbf{Parametric methods} select a copula family and estimate parameters via maximum likelihood, which is fast but potentially misspecified. \textbf{Nonparametric methods} estimate the copula density directly using kernel density estimation with boundary corrections \citep{nagler2018kdecopula}, offering flexibility at higher computational cost.

Efficient libraries like \texttt{vinecopulib} \citep{nagler2025rvinecopulib} reduce overhead but do not change the repeated-fitting paradigm. Our method addresses this by amortizing fitting cost across all edges.

\subsection{Neural Copulas and Global Models}

Recent work explores neural copula estimation: implicit generative copulas \citep{janke2021igc}, neural copula functions \citep{zeng2022neuralcopula}, Deep Archimedean Copulas \citep{ling2020deep}, copula density estimators \citep{letizia2025codine}, and classifier-based density-ratio copulas \citep{huk2025classifier}. These methods cover global copula learning, family-specific neural pair-copulas, and standalone density estimators. The classification view of \citet{huk2025classifier} is especially relevant because it identifies copula density estimation with learning dependence against an independence reference distribution. Diffusion- and flow-based copulas \citep{huk2026diffusionflow} are also closely related, but target global dependence models using marginal-preserving processes for density estimation and sampling. Our use case is narrower and complementary: one pretrained bivariate edge operator that plugs directly into vine recursion through explicit densities, valid marginals after IPFP, and cached h-functions. Normalizing flows \citep{papamakarios2017maf, durkan2019nsf, papamakarios2021flows} and diffusion models \citep{ho2020ddpm, kotelnikov2023tabddpm} are powerful global estimators but learn monolithic models without vine modularity.

\subsection{Mutual Information Estimation}

Classical MI estimators include the KSG $k$-NN estimator \citep{kraskov2004ksg}. Neural estimators based on variational bounds (MINE, InfoNCE, NWJ) can be high variance or loose \citep{belghazi2018mine, oord2018representation, nguyen2010estimating, poole2019variational, mcallester2020formal}. The identity between mutual information and negative copula entropy was explicitly noted by \citet{davy2003copulas} before the later copula-entropy treatment of \citet{ma2011mutual}. Copula-based approaches have been used for information estimation in neuroscience, including nonparametric copula estimators \citep{safaai2018information} and copula and vine-copula models for joint neural responses \citep{onken2016mixed, onken2009analyzing}. Recent work explores diffusion-based estimation \citep{franzeseminde} and meta-learned prediction \citep{gritsai2026mist}. Vines give MI/TC estimates as by-products of fitting an explicit density, with decomposable edge-wise attribution.

\section{Theoretical Background}
\label{app:theory}

\subsection{Copula Definition}

\begin{definition}
For $u=(u_1,\ldots,u_d)\in[0,1]^d$, a $d$-dimensional copula $\copula: [0,1]^d \to [0,1]$ is a CDF with uniform marginals satisfying:
\begin{enumerate}[label=(\roman*), leftmargin=*, itemsep=1pt, topsep=2pt]
    \item $\copula(u) = 0$ if any $u_i = 0$.
    \item $\copula(u) = u_i$ if all $u_j = 1$ for $j \neq i$.
    \item $\copula$ is $d$-increasing.
\end{enumerate}
\end{definition}

When it exists, the copula density is $c(u) = \frac{\partial^d \copula(u)}{\partial u_1 \cdots \partial u_d}$.

\subsection{H-Functions}

For a pair of uniform random variables $(U,V)$ with bivariate copula $\copula(u,v)$ and density $c(u,v)$, h-functions are the corresponding conditional CDFs:
\begin{align}
h_{U|V}(u \mid v) &= \frac{\partial \copula(u,v)}{\partial v} = \int_0^u c(s,v) \, ds \\
h_{V|U}(v \mid u) &= \frac{\partial \copula(u,v)}{\partial u} = \int_0^v c(u,t) \, dt
\end{align}
These satisfy $h_{U|V}(u \mid v) \in [0,1]$ and are monotonically increasing in $u$. For a target conditional probability $w\in[0,1]$, the inverse h-function $h^{-1}_{U|V}(w \mid v)$ gives the conditional quantile.

\subsection{Information-Theoretic Quantities}

\textbf{Mutual Information.} For continuous random variables $(X,Y)$ with copula density $c$, let $H(c)$ denote the differential entropy of $c$:
\begin{equation}
I(X;Y) = -H(c) = \int_{[0,1]^2} c(u,v) \log c(u,v) \, du \, dv.
\end{equation}

\textbf{Total Correlation.} For $U_i=F_i(X_i)$, where $F_i$ is the marginal CDF of $X_i$:
\begin{equation}
\mathrm{TC}(X_1, \ldots, X_d) = \EE[\log c(U_1, \ldots, U_d)]
\end{equation}

Under vine factorization, let $\mathcal E$ be the vine edge set and let $c_e$ denote the pair-copula density on edge $e\in\mathcal E$. Then
\[
\mathrm{TC} = \sum_{e \in \mathcal E} \EE[\log c_e].
\]

\subsection{What Is and Is Not Guaranteed}
\label{app:guarantees}

\paragraph{Exact grid properties.} The clipped exponential makes the network output strictly positive. Proposition~\ref{prop:ipfp} therefore applies: at the IPFP limit, the unique KL projection has exactly uniform grid marginals and defines a valid pair-copula density. An exactly constrained $m\times m$ grid caps representable MI at $\log m$. The limiting projection also satisfies the KL-improvement statement below.

\paragraph{Numerical implementation.} For fixed edge grids and evaluation inputs, density evaluation, h-function recursion and the sum of edge log densities are deterministic. Sampling introduces Monte Carlo variability and inherits an interpolation discrepancy. Edge densities use bilinear interpolation between cell centers and integrate exactly to one; h-functions interpolate cumulative sums. Their derivatives differ from the edge densities by $1$--$3\%$ on average (mean log-ratio about $0.001$--$0.002$ nats) on fits to five synthetic copulas. Thus the assembled multivariate density is only approximately normalized, and inverse-h sampling targets a slightly different distribution from the reported density. Tested three-variable UCI sub-vines ($n{=}5{,}000$) integrate to one within $2.3\times10^{-4}$ for the strongest chains (GAS and MINIBOONE, pairwise $|\tau|$ of $0.80$--$0.88$; quadrature up to $512^3$ points, error still decreasing with resolution) and within $2.2\times10^{-5}$ for the other triples. Two-variable fits integrate to one within $10^{-7}$. These checks do not establish a bound for higher-dimensional vines or other grids. At $K{=}30$, the maximum marginal-density residual is about $2\times10^{-4}$ on four synthetic test edges and $8\times10^{-3}$ on GAS; UCI NLL and PIT-KS change by at most $0.001$ from $K{=}30$ to $K{=}100$ (Table~\ref{tab:ipfp_uci_sensitivity}).

\begin{proposition}[KL improvement for the cell-constant grid]
\label{prop:pythagorean}
Let $c^{\star}$ be the piecewise-constant density obtained by averaging the true copula density within each cell of the $m \times m$ grid, let $Q$ be the normalized, strictly positive raw density grid from Section~\ref{subsec:ipfp}, and let $\Pi$ be the set of normalized density grids with row and column masses $1/m$. Define the exact projection $\hat{D}_{\infty} = \argmin_{P \in \Pi} \KL(P \| Q)$. Then
\[
\KL(c^{\star} \,\|\, Q) \;=\; \KL(c^{\star} \,\|\, \hat{D}_{\infty}) \;+\; \KL(\hat{D}_{\infty} \,\|\, Q),
\]
so $\KL(c^{\star} \| \hat{D}_{\infty}) \leq \KL(c^{\star} \| Q)$, with strict improvement whenever $Q \notin \Pi$.
\end{proposition}

\begin{proof}
The uniform-marginal constraints are linear in $P$, so $\Pi$ is a linear family containing $c^{\star}$; the identity is the Pythagorean property of I-projections onto linear families \citep{csiszar1975divergence}. IPFP converges to this I-projection for strictly positive $Q$ (Proposition~\ref{prop:ipfp}).
\end{proof}

The projection inequality concerns cell-constant grid densities and their discrete-grid KL divergence. It does not imply an improvement in the continuous likelihood of the bilinearly interpolated density used by the implementation.

These projection results do not establish statistical consistency, convergence rates, or sample complexity for the amortized estimator. Nor do they guarantee that finite-iteration projection preserves sharp features or tail structure while enforcing the marginals. We examine these questions empirically through the out-of-distribution family benchmark, fixed-checkpoint sample-size sweep, projection-sensitivity study, depth-recovery stress test, and self-consistency tests.

\section{Implementation Details}
\label{app:implementation}

\subsection{Network Architecture}

The \ours{} edge denoiser is a convolutional encoder--decoder U-Net with $m \times m$ input (default $m=64$) and four input channels: the histogram, a constant channel equal to $\log n$, and two probit-transformed coordinate channels. It has depth $4$ with channel widths $128, 256, 512, 1024$ (doubling at each stride-2 downsampling), two residual blocks per level, additive skip connections, dropout $0.1$, and $1.33\times10^{8}$ parameters. Each residual block is modulated by FiLM \citep{perez2018film} from a 256-dimensional sinusoidal embedding of the corruption level of Appendix~\ref{app:training} (zero at inference); the sample size enters only through the $\log n$ channel. The output head returns log-density logits $Z$, with $D_{\text{raw}} = \exp(\operatorname{clip}(Z,-20,20))$ at inference (Step 2 of the inference procedure).

\subsection{Training Data}
\label{app:training_data}

The synthetic copula zoo samples from a generator covering Gaussian, Student-$t$, Archimedean (Clayton, Gumbel, Frank, Joe), BB1/BB7 families, conditional Gaussian/Clayton pair-copula examples, complex synthetic patterns, independence, and 2--4 component mixtures; sample sizes follow $n \sim \mathrm{LogUniform}(200, 8000)$, where $\mathrm{LogUniform}(a,b)$ means that $\log n$ is uniform on $[\log a,\log b]$, and the training data are streamed rather than cached. Table~\ref{tab:benchmark_spec} records the frozen checkpoint's actual generator configuration and the composition of the Table~\ref{tab:bivariate}, replication, and out-of-distribution evaluation suites. In the table, $\rho$ is a correlation coefficient, $\nu$ is the Student-$t$ degrees of freedom, $\theta$ and $\delta$ are copula-family parameters, $\sigma$ is a pattern-width parameter, $w_2$ is the relative weight of the second diagonal in the X pattern, $r_0$ is the ring radius, $k$ is the mixture component count, $\psi_1,\psi_2$ are Tawn asymmetry parameters, and amplitude and offset control the double-banana shape.

\begin{table}[t]
    \centering
    \caption{\textbf{Benchmark specification.} Frozen-checkpoint training-zoo configuration and the composition of the main bivariate evaluation suites. Both $\mathcal{U}$ and $\mathrm{Unif}$ denote uniform distributions, and $\mathbf{1}_k$ is the length-$k$ vector of ones.}
    \label{tab:benchmark_spec}
    \scriptsize
    \begin{tabular}{p{0.22\columnwidth}p{0.70\columnwidth}}
    \toprule
    \multicolumn{2}{l}{\emph{Training zoo (streamed; $\approx 6.1\times10^6$ examples at the deployed checkpoint)}} \\
    \midrule
    Gaussian & $\rho \sim \mathcal{U}(-0.95, 0.95)$ \\
    Student-$t$ & $\rho \sim \mathcal{U}(-0.9, 0.9)$, $\nu \sim \mathcal{U}(2, 30)$ truncated to an integer \\
    Clayton / Gumbel / Frank / Joe & $\theta \sim \mathcal{U}(0.1, 14)$ / $\mathcal{U}(1.1, 12)$ / $\mathcal{U}(-15, 15)$ / $\mathcal{U}(1.1, 12)$ \\
    BB1 / BB7 & $(\theta, \delta) \sim \mathcal{U}(0.1,5) \times \mathcal{U}(1,5)$ / $\mathcal{U}(1,5) \times \mathcal{U}(0.1,5)$ \\
    Rotations & with probability $0.3$, one of $\{0^\circ,90^\circ,180^\circ,270^\circ\}$ for Clayton/Gumbel/Joe (BB1/BB7 always unrotated) \\
    Structured patterns & X: $\sigma\!\sim\!\mathcal{U}(.02,.06)$, $w_2\!\sim\!\mathcal{U}(.5,2)$; ring: $r_0\!\sim\!\mathcal{U}(.20,.36)$, $\sigma\!\sim\!\mathcal{U}(.02,.06)$; double banana: amplitude $\sim\mathcal{U}(.10,.22)$, offset $\sim\mathcal{U}(.08,.25)$, $\sigma\sim\mathcal{U}(.02,.06)$; all projected by IPFP \\
    Conditional examples & Gaussian/Clayton parameters from ranges above; conditioning value $\sim\mathcal{U}(0.1,0.9)$ (defective, see text) \\
    Mixtures & probability $0.20$; $k\sim\mathrm{Unif}\{2,3,4\}$ parametric components; weights $\sim\mathrm{Dirichlet}(\mathbf{1}_k)$ \\
    Family sampling weights & Gaussian $.12$, Student $.08$, Clayton $.10$, Gumbel $.12$, Frank $.05$, Joe $.12$, BB1/BB7 $.08$ each; conditional Gaussian/Clayton $.07$ each; independence $.03$; X/ring/double-banana $.03/.03/.02$ \\
    Sample size & $n \sim \mathrm{LogUniform}(200, 8000)$ per example \\
    \midrule
    \multicolumn{2}{l}{\emph{Evaluation suites (fixed cases; evaluation samples are separate from training)}} \\
    \midrule
    Pretraining-family suite & Tables~\ref{tab:bivariate} and~\ref{tab:bivariate_dispersion_appendix}: $9$ fixed cases, five fresh pyvinecopulib samples ($n{=}2000$) per case: independence; Gaussian($\rho{=}{\pm}0.7$); $t(\rho{=}0.7,\nu{=}5)$; Clayton($\theta{=}3$), unrotated and $90^\circ$; Gumbel($2.5$); Joe($3$); Frank($5$) \\
    Replication suite & Table~\ref{tab:bivariate_replicates}: $8$ cases, $100$ replications each, $n{=}2000$: Gaussian($0.6$), $t$($0.6, 5$), Clayton($3$), Gumbel($2.5$), Frank($5$), Joe($3$), BB1($1.5,1.5$), BB7($2.5,1.5$) \\
    Out-of-distribution (OOD) suite & Table~\ref{tab:ood_families}: $9$ cases, seeds $\{7,17,42\}$, $n{=}2000$: BB6($2.0,1.5$), BB6($3.0,2.0$), BB8($3.0,0.8$), BB8($5.0,0.6$), Tawn($\psi_1{=}0.7,\psi_2{=}1.0,\theta{=}2.5$), Plackett($5$), Plackett($20$), BB6$^{90^\circ}$($2.0,1.5$), BB8$^{90^\circ}$($3.0,0.8$) \\
    \bottomrule
    \end{tabular}
\end{table} With batch size $32$, the selected step-$190{,}000$ checkpoint sees roughly $6.1\times10^6$ synthetic histogram-target pairs; the full $240{,}000$-step schedule sees roughly $7.7\times10^6$. Targets are the analytic density evaluated at cell centers and normalized (complex patterns: their projected grid), and inputs are histograms of raw copula samples.

\paragraph{Training-data defects.} Checks against pyvinecopulib and the target formulas identify four mismatched training generators: the Joe sampler (an incorrect conditional-CDF formula makes the bisection collapse to a two-valued second coordinate), the BB1 sampler (an incorrect frailty construction with non-uniform margins), the BB7 sampler (it inverts the wrong conditional), and the ``conditional'' examples (deterministic curves $(h^{-1}(w\mid v_0), w)$ supervised with the unconditional base density). Computing the expected composition of the training stream from the recorded commit's generator and the run's configuration (the run's uncommitted local changes were not recorded, so this is a reconstruction rather than an exact replay), about $43\%$ of training pairs are affected (including mixtures with a Joe component, assuming that mixture components were limited to the families the repository sampler supports, as in later code); a further $12\%$ ($18\%$ counting mixture components) have correct samples but cell-center targets that overstate the mass of the tail corner cell by more than a factor of $2.6$ (Clayton $\theta \geq 4.4$, Gumbel $\theta \geq 5$). The network learned these mappings (for example, it returns the intended Joe density from the degenerate Joe input). We report the deployed checkpoint as trained; every evaluation in this paper except the auxiliary probit ablation (Appendix~\ref{app:probit}) uses correctly simulated data (pyvinecopulib, the repository's verified samplers, a closed-form Plackett sampler, or exact vine samplers). Appendix~\ref{app:retrained} separately evaluates estimators retrained on corrected data; the main results retain this original checkpoint.

\subsection{Pseudo-Observations and Histogram Encoding}

For a raw bivariate sample $\{(x_i,y_i)\}_{i=1}^n$, let $r_i^{(x)}$ and $r_i^{(y)}$ be the within-sample ranks of $x_i$ and $y_i$. We convert to copula space using empirical PIT / rank pseudo-observations
\[
u_i = \frac{r_i^{(x)}}{n+1}, \qquad v_i = \frac{r_i^{(y)}}{n+1}.
\]
This is the transform of the method; the experiments on simulated copula data pass the simulated pseudo-observations directly, UCI training and test rows use $(r+0.5)/(n_{\mathrm{train}}+1)$ with ranks among the training rows (Appendix~\ref{app:uci_density}), S\&P100 training rows use $r/(n+1)$ and its test rows $(r+0.5)/(n+1)$ with ranks among the training rows, and the self-consistency protocol uses $(r-0.5)/n$.
With grid width $\Delta = 1/m$, cell indices $a,b\in\{1,\ldots,m\}$, cells $B_{ab}$, and indicator function $\mathbf 1[\,\cdot\,]$, the network input is the histogram density estimate
\[
H_{ab} = \frac{1}{n \Delta^2} \sum_{i=1}^n \mathbf{1}\!\left[(u_i,v_i) \in B_{ab}\right],
\qquad
\sum_{a,b} H_{ab}\Delta^2 = 1.
\]
During training, the histogram of the simulated copula sample is optionally corrupted to produce $\tilde H$. Independently for each example, draw a noise level $t\sim\mathrm{Unif}(0,1)$, which is also passed to the network (the direct variant is conditioned on $t=0$); let $H_{\mathrm{unif}}$ be the uniform density grid with entries $(H_{\mathrm{unif}})_{ab}=1$, let $\varepsilon=(\varepsilon_{ab})$ be a grid of independent Gaussian perturbations with standard deviation $\sigma_g t$, $\sigma_g=0.2$, and let $N=(N_{ab})$ be a count grid resampled from the cell probabilities $\{H_{ab}\Delta^2\}_{ab}$ with $n'=\max(n,32)$ draws. The four variants compared in Appendix~\ref{app:corruption_ablation} are:
\begin{align*}
\tilde H &= H
&& \text{(Direct)}, \\
\tilde H &= (1-\alpha)H + \alpha H_{\mathrm{unif}},\quad \alpha = 0.65\,t
&& \text{(Uniform-mix, deployed)}, \\
\tilde H &= \mathrm{clip}(H+\varepsilon, 0, \infty)\ \text{renormalized},\quad \varepsilon_{ab} \sim \mathcal{N}\!\left(0, (\sigma_g t)^2\right)
&& \text{(Gaussian)}, \\
N &\sim \mathrm{Mult}\!\left(n',\{H_{ab}\Delta^2\}_{ab}\right),
&& \\
\tilde H &= (1-t) H + t\, N / (n'\Delta^2)
&& \text{(Multinomial)}.
\end{align*}
At test time in the paper's one-shot setting, we always use $\tilde H = H$ and $t=0$.

\subsection{Training Objective}

Let $\mathcal{L}_{\mathrm{CE}}$, $\mathcal{L}_{\mathrm{ISE}}$, $\mathcal{L}_{\mathrm{tail}}$, and $\mathcal{L}_{\mathrm{marg}}$ denote the cross-entropy, integrated-squared-error, tail, and marginal loss terms below, and let $\lambda_{\mathrm{ISE}}$, $\lambda_{\mathrm{tail}}$, and $\lambda_{\mathrm{marg}}$ denote their respective weights. We minimize
\begin{equation}
\mathcal{L} = \mathcal{L}_{\text{CE}} + \lambda_{\text{ISE}} \mathcal{L}_{\text{ISE}} + \lambda_{\text{tail}} \mathcal{L}_{\text{tail}} + \lambda_{\text{marg}} \mathcal{L}_{\text{marg}}.
\end{equation}

\textbf{Cross-entropy loss} $\mathcal{L}_{\text{CE}}$ measures how well the predicted density matches the ground truth when used for likelihood evaluation. For cell indices $i,j\in\{1,\ldots,m\}$, let $T=(T_{ij})$ be the ground-truth density grid, let $\hat D=(\hat D_{ij})$ be the predicted density after the IPFP projection, and let $\Delta=1/m$ be the bin width. On cell masses (each clamped below at $10^{-12}$),
\begin{equation}
\mathcal{L}_{\text{CE}} = -\sum_{i,j} T_{ij} \Delta^2 \log(\hat{D}_{ij}\Delta^2),
\end{equation}
which differs from the density form by the constant $\log\Delta^2$.

\textbf{Integrated Squared Error} $\mathcal{L}_{\text{ISE}}$ directly penalizes pointwise density errors:
\begin{equation}
\mathcal{L}_{\text{ISE}} = \sum_{i,j} (\hat{D}_{ij} - T_{ij})^2 \Delta^2.
\end{equation}

\textbf{Tail loss} $\mathcal{L}_{\text{tail}}$ emphasizes the boundary band where tail dependence is captured. With cell centers $(u_i,v_j)$ and band width $\tau=0.15$, let $\mathcal R=\{(i,j): \min(u_i,v_j)<\tau \text{ or } \max(u_i,v_j)>1-\tau\}$:
\begin{equation}
\mathcal{L}_{\text{tail}} = \frac{1}{m^2}\sum_{(i,j) \in \mathcal{R}} (\log \hat{D}_{ij} - \log T_{ij})^2.
\end{equation}

\textbf{Scaled marginal-KL regularizer} $\mathcal{L}_{\text{marg}}$ penalizes residual non-uniformity of the row and column marginals. In the deployed implementation it is applied to the projected density $\hat D$, so it is small whenever the finite projection has converged. With marginal densities $g_a^{\text{row}} = \Delta \sum_b \hat D_{ab}$ and $g_b^{\text{col}} = \Delta \sum_a \hat D_{ab}$:
\begin{equation}
\mathcal{L}_{\text{marg}} = \sum_a g_a^{\text{row}} \log g_a^{\text{row}} + \sum_b g_b^{\text{col}} \log g_b^{\text{col}}.
\end{equation}
This is $m$ times the sum of the row and column marginal KL divergences: the integrated form includes $\Delta=1/m$. We retain this implementation's scaling with the reported loss weight.

\textbf{Canonical loss weights.} The deployed checkpoint uses
$\lambda_{\text{ISE}} = 0$, $\lambda_{\text{tail}} = 0.2$, and $\lambda_{\text{marg}} = 0.02$, kept fixed across all main experiments. An earlier ablation checkpoint used $\lambda_{\text{tail}}=0.1$ and $\lambda_{\text{marg}}=0$, but it is not the canonical checkpoint used for the paper results.

\subsection{Inference Procedure}

At inference time, given pseudo-observations $(u_1, \ldots, u_n)$ and $(v_1, \ldots, v_n)$ for a bivariate copula:

\textbf{Step 1: Histogram construction.} We bin the data into an $m \times m$ grid and normalize to obtain $H$.

\textbf{Step 2: Network forward pass.} The histogram $H$ and sample-size encoding $\log n$ are passed through the trained U-Net to obtain log-density logits $Z$, which are transformed to a positive raw density via $D_{\text{raw}} = \exp\!\bigl(\operatorname{clip}(Z, -20, 20)\bigr)$. Clipping bounds the dynamic range and avoids unbounded tails before projection.

\textbf{Step 3: IPFP projection.} We apply $K{=}30$ iterations of row/column normalization to obtain a projected copula density $\hat{D}$, with maximum row/column marginal error around $2\!\times\!10^{-4}$ on representative test densities (Table~\ref{tab:ipfp_iter_ablation}) and up to $8\times10^{-3}$ on the most strongly dependent real edges (Table~\ref{tab:ipfp_uci_sensitivity}).

\textbf{Step 4: H-function computation.} We compute cumulative sums to obtain $h_{U|V}$ and $h_{V|U}$ tables, which are cached for vine recursion and evaluated by interpolation; edge densities are evaluated by bilinear interpolation of $\hat D$ between cell centers.

\paragraph{Failure diagnostics at inference.}
The explicit edge object makes reliability checks available without knowing the true density. We recommend three per-edge checks: (i) evaluate held-out edge log-likelihood relative to the independence density using a sample split; (ii) record the pre-projection marginal deviation, $\max_a |\mu_a^{\text{row}}-1/m|$ and its column analogue, together with the magnitude of the IPFP correction; and (iii) test stability across bootstrap resamples or a stricter projection setting. A negative held-out gain over independence, an unusually large repair, or unstable estimates is an actionable reason to fall back to a per-edge TLL fit. These are practical alarms rather than certified out-of-distribution detectors: Table~\ref{tab:ood_families} reports performance on the tested excluded families, but it does not by itself calibrate universal thresholds for the diagnostics.

\subsection{Training Protocol}
\label{app:training}

We train the canonical \ours{} edge estimator with AdamW (learning rate $8\times 10^{-5}$, held constant; weight decay $0.01$) for $2.4\times 10^{5}$ optimizer steps with batches of $32$ synthetic edges streamed on the fly from the generator of Appendix~\ref{app:training_data} by four data-loader workers. Mixed precision uses PyTorch CUDA autocast at its default float16 with dynamic loss scaling; gradient norms are clipped at $1.0$. Inside the loss, the network output is projected with three log-domain IPFP iterations, extended to $30$ when the mean-squared marginal deviation of the three-iteration result exceeds $0.01$. During training we corrupt the input histogram by uniform mixing with strength $\alpha = 0.65\,t$, $t \sim \mathrm{Uniform}(0,1)$, and condition the network on $t$; the loss is evaluated against the clean target density. The loss weights are $\lambda_{\text{ISE}} = 0$, $\lambda_{\text{tail}} = 0.2$, and $\lambda_{\text{marg}} = 0.02$. The deployed checkpoint is the one saved at step $190{,}000$; it was chosen among $24$ saved checkpoints by mean ISE on the standard and complex bivariate suites. Table~\ref{tab:bivariate} uses fresh samples independent of checkpoint selection. Family and parameter draws are seeded ($42$ plus the worker offset), while sample generation also uses random states not fixed by that seed; network initialization was not separately seeded, and the checkpoint stores model weights and configuration but not optimizer state.

\subsection{Reproducibility and Compute}
\label{app:reproducibility}

\paragraph{Timing conventions.} Fit times include structure selection and exclude checkpoint loading, evaluation and the reusable estimator's one-time pretraining. Warm-up is excluded except in the first fit of each UCI run. The full $240{,}000$-step training run took $9.6$\,h on one GPU; the deployed step-$190{,}000$ checkpoint was reached after $7.6$\,h. Main scaling fits run sequentially on the same AMD EPYC 9454 node with $16$ CPU cores, with \ours{} adding one H100 80\,GB GPU. They measure the combined benefit of amortization and GPU batching. Tables~\ref{tab:pyvine_threads} and~\ref{tab:cpu_timing} give a thread-count sweep and CPU-only measurements with the original execution.

Figure~\ref{fig:training_diagnostics} shows the training curve and the projection of one raw output. Training ran on one NVIDIA H100 GPU (the job allocated four; the trainer used one) and took $9.6$ hours of wall-clock time for $240{,}000$ steps; the deployed step-$190{,}000$ checkpoint was written after $7.6$ hours. The training run records the library commit and configuration; uncommitted local changes at the time are not captured, so the exact training source is only partially pinned. The main scaling timings use one NVIDIA H100 80\,GB GPU with $16$ CPU threads for \ours{} and separate $8$- and $16$-thread TLL fits on the same AMD EPYC 9454 node (Table~\ref{tab:fast_fit_timing}). Unless a caption states otherwise, other GPU benchmarks report their original timings on one NVIDIA H200 (the original scaling sweep, the mixed-family vines, the random-split S\&P100 run and the TC runtime sweep) or H100 GPU, with $8$ CPU threads ($16$ at $d{=}500$ and for the mixed-family vines), and their pyvine baselines use the allocated CPU threads. The UCI benchmark rerun on the MAF splits (Appendix~\ref{app:uci_density}) used one A100 slice (MIG 3g.20gb) and $4$ CPU threads. The deployed checkpoint (sha256 \texttt{8a1b71ad\ldots f5c0f}) is public. All downstream evaluation runs fix their own RNG seeds. UCI evaluations use the MAF train/test splits of \citet{papamakarios2017maf}, subsampled as described in Appendix~\ref{app:uci_density}, and apply PIT via empirical CDFs fit on the training rows ($(r+0.5)/(n+1)$ with clipping, as specified in Appendix~\ref{app:uci_density}). The public code release (v0.2.6) includes portable experiment runners in \texttt{projects/neurips2026} and a downloader that rebuilds the MAF splits with hash checks. Its accompanying figure archive contains the scripts and saved outputs that regenerate every figure and $25$ tables, together with an inventory of the remaining tables. The runners and archived inputs record settings and input, code, and checkpoint hashes.
\ifdefined\arxivmode
For the public arXiv release, the code repository and released checkpoint are listed in the first-page availability footnote.
\fi

\begin{figure}[!htbp]
    \centering
    \includegraphics[width=\textwidth]{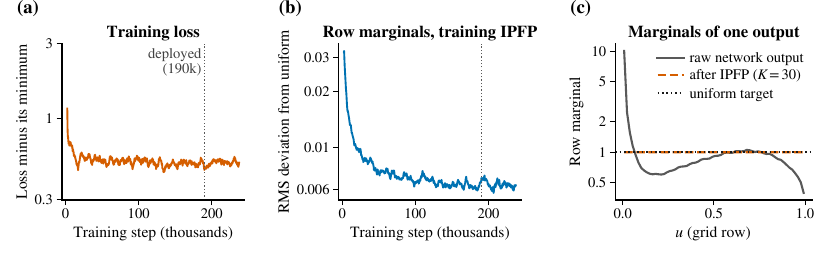}
    \caption{\textbf{Training and projection diagnostics of the deployed checkpoint.} (a) Training loss minus its minimum and (b) per-batch root-mean-square deviation from uniform of the grid row marginals after the training-time projection (three IPFP iterations, thirty when needed), both as centered moving averages over $25$ logged points ($5{,}000$ steps) shown where the full window is available, on log scales; dotted lines mark the deployed step-$190{,}000$ checkpoint. (c) Row-marginal profile of a raw network output for a Clayton ($\theta{=}3$, $n{=}1500$) sample and of the same grid after the deployed IPFP projection ($K{=}30$; Section~\ref{subsec:ipfp}), with the uniform target for reference.}
    \label{fig:training_diagnostics}
\end{figure}

\subsection{Baseline Configurations}
\label{app:baselines}

\textbf{Histogram and KDE (Table~\ref{tab:bivariate}).} The histogram baseline is the $64\times64$ histogram of the sample, and the KDE is a Gaussian kernel density estimate in probit coordinates with Scott's bandwidth, mapped back to copula space on the grid; both grids are IPFP-projected ($K{=}50$) before scoring.

\textbf{pyvine (parametric / TLL).} We use \texttt{pyvinecopulib} 0.7.5, the Python interface to the C++ library \texttt{vinecopulib}, which also underlies \texttt{rvinecopulib} \citep{nagler2025rvinecopulib}, with default settings for parametric family selection and transformation local likelihood (TLL) estimation, except that the UCI benchmarks and the non-simplified stress test evaluate TLL on a $64$-point grid (default $30$); learned regular-vine structure uses Di{\ss}mann's algorithm with the default $|\tau|$ criterion.

\textbf{RealNVP flow.} We use an $8$-block RealNVP-style affine-coupling flow \citep{dinh2017realnvp} with hidden size $128$, two hidden layers per coupling network, and probit-transformed copula inputs; the UCI runner trains it without weight decay or gradient clipping, the other experiments with weight decay $10^{-6}$ and gradient-norm clipping at $5$. It is trained with Adam at learning rate $10^{-3}$ for up to $25$ epochs, using a $10\%$ validation split and early stopping with patience $5$ on held-out NLL.

\textbf{ACNet.} We use the authors' reference implementation \citep{ling2020deep} (two hidden layers of width $10$), trained from scratch for each synthetic Archimedean pair with Adam (learning rate $10^{-3}$, batch size $200$) for a fixed $300$ epochs ($3{,}000$ steps, no early stopping) on $n{=}2000$ samples and evaluated on the $64\times 64$ grid; its metrics are averaged over three training seeds.

\textbf{MINE, InfoNCE and NWJ.} We use standard implementations with a $3$-layer multilayer-perceptron (MLP) critic (rectified-linear-unit activations) trained with Adam at batch size $\min(512, n)$; MINE uses no moving-average gradient correction. The settings differ by benchmark: the bivariate MI suite (Table~\ref{tab:mi_appendix}) uses $2{,}000$ steps, learning rate $10^{-3}$ and hidden width $128$; the MI-versus-dimension study (Appendix Figure~\ref{fig:dimension_appendix}(a)) uses $300$ steps with the same learning rate and width; the TC benchmark, which sums chain-rule terms, and the block-MI sample-size study, which trains one critic on the two blocks, use $800$ steps, learning rate $10^{-4}$ and hidden width $256$. Runtime is reported per fixed $(n, d)$ evaluation.

\textbf{MINDE.} We use the authors' reference implementation in conditional mode (\texttt{type=c}) on the unstandardized pseudo-observations (its rescaling option is not applied in this path) and with its MLP architecture. Each case uses an $80/20$ train/test split, $50$ epochs, learning rate $10^{-2}$, batch size $256$, $10$ Monte Carlo iterations, and neither exponential moving averaging nor importance sampling.

\textbf{KSG.} We implement the KSG estimator with $k=5$ neighbors and use \texttt{scipy.spatial.cKDTree} for nearest-neighbor queries; reported runtime uses a single CPU thread.

\textbf{Gaussian copula baseline.} Fits a single covariance matrix on probit-transformed PIT scores; MI/TC are computed in closed form from the fitted correlation structure.

\subsection{External Assets and Licenses}
\label{app:assets}

We use existing public benchmark data and external software only for evaluation. Raw third-party datasets are not redistributed with the paper; the UCI datasets are obtained from the public archive of \citet{papamakarios2017maf} (Zenodo record 1161203) and preprocessed with their reference code. The Combined Cycle Power Plant (CCPP) data used for sampling and imputation are from the UCI Machine Learning Repository \citep{tufekci2014ccpp, uci2014ccpp} under a CC BY 4.0 license. The S\&P100 experiment uses derived daily log-returns from Stooq daily close prices, with component symbols taken from the public Wikipedia S\&P100 component list at collection time; users must comply with the original Stooq and Wikipedia terms if they recreate the raw data. The public project lists the $100$ symbols, the source and the date range ($2019$-$05$-$13$ to $2026$-$02$-$06$, $1{,}695$ trading days); the price collection was not recorded in enough detail to rebuild the paper's matrix exactly, so a recreated matrix can reproduce the protocol but not the reported values digit for digit. External packages and baselines are cited where used, including \texttt{rvinecopulib}/\texttt{pyvinecopulib}, ACNet, RealNVP-style flows, KSG, MINE, InfoNCE, and MINDE. The public library project and figure archive include license and reproduction instructions; users remain responsible for complying with the original terms of any external datasets or packages they install.

\section{Experimental Details and Additional Results}
\label{app:downstream}

This appendix groups the experimental material into four themes. \emph{Edge-level diagnostics}: the bivariate dispersion and paired replication study (Tables~\ref{tab:bivariate_replicates} and~\ref{tab:bivariate_dispersion_appendix}) and the additional edge diagnostics, including hard synthetic families, out-of-distribution families, and the grid-resolution floor (\ref{app:grid_floor}); a comparison of the retained training-corruption variants on common inputs (\ref{app:corruption_ablation}); the neural pair-copula comparison (\ref{app:real_edge_scaling}) and the controlled depth stress test (Table~\ref{tab:deep_tree_recovery}) provide complementary diagnostics. \emph{Full-vine density benchmarks}: UCI details and dataset provenance (\ref{app:uci_density}), the dependence-signal diagnostic (\ref{app:uci_dependence_diag}), RealNVP capacity sensitivity (\ref{app:flow_capacity_sensitivity}), high-dimensional and mixed-family scaling (\ref{app:highdim_scaling}, \ref{app:mixed_family_scaling}), and the S\&P100 benchmark with random and chronological splits (\ref{app:sp100_density}). \emph{Information estimation}: MI benchmark numbers (Table~\ref{tab:mi_appendix}), the self-consistency protocol (\ref{app:self_consistency}), and fixed-checkpoint sample-size sensitivity (\ref{app:mi_sample_size}). \emph{Downstream stress tests}: the non-simplified conditional stress test (\ref{app:nonsimplified_stress}), unconditional sampling quality (\ref{app:sampling_quality}), VaR backtesting (\ref{app:var}), the rolling dependence dashboard (\ref{app:rolling_dashboard}), and missing-data imputation (\ref{app:imputation}).

\subsection{Comparison of Retained Corruption Variants}
\label{app:corruption_ablation}

Four edge estimators were trained with the same saved generator configuration, loss, optimizer, architecture and recorded seed but different histogram corruptions (the alternatives ran a later version of the trainer; see Scope) (formulas in Appendix~\ref{app:implementation}): direct histograms, uniform mixing (the deployed model), Gaussian grid noise and multinomial count resampling.

\paragraph{Protocol.} A protocol fixed before the run scores every retained checkpoint on identical inputs: the nine cases of Table~\ref{tab:bivariate} (five seeds; held-out KL against the true density, and ISE against exact cell averages), the nine out-of-distribution cases of Table~\ref{tab:ood_families} (three seeds), a common panel of real-data vine edges, and each checkpoint's own D-vine on the MAF data with the rows of Table~\ref{tab:vine_nll}. For the panel, one D-vine order and reference TLL pair-copulas fitted tree by tree give every checkpoint the same training and test pseudo-observations for all $1{,}156$ edges of the four datasets; each checkpoint estimates every edge from its training pseudo-observations and is scored by held-out edge log-density. Before any alternative was scored, the deployed checkpoint reproduced the saved likelihoods of Table~\ref{tab:bivariate} within $2\times10^{-6}$ nats, the out-of-distribution NLLs within $3\times10^{-6}$ nats and the UCI NLLs within $2\times10^{-4}$ bits/dim; one amendment, also made before any alternative was scored, widened the ISE check to $0.5\%$ because ISE magnifies last-digit grid differences between GPU configurations.

\paragraph{Results.} No corruption is best everywhere (Tables~\ref{tab:corruption_variants} and~\ref{tab:corruption_edges}). On the Table~\ref{tab:bivariate} cases the alternatives differ from the deployed model by at most $2\times10^{-4}$ nats in mean held-out KL ($3\times10^{-4}$ between any two), and on the out-of-distribution cases only Gaussian noise differs clearly (worse by $0.0015$ nats); ISE ranks them differently from KL, and every alternative has the lower out-of-distribution ISE. On the real data the edge panel and the end-to-end vines agree in direction for every dataset and variant: the deployed uniform-mix model is best on POWER and MINIBOONE, multinomial resampling is best on GAS (by $0.013$ bits/dim over the deployed model, end to end) and HEPMASS (by $0.001$), and the direct model trained without corruption is the weakest of the four schemes on all four datasets on the panel and, against each scheme's primary checkpoint, end to end (the multinomial run's step-$190{,}000$ checkpoint is marginally weaker than it end to end on MINIBOONE). Each alternative has a higher held-out edge log-density than the deployed model on $274$ to $427$ of the $1{,}156$ panel edges.

\paragraph{Scope.} This is a comparison of retained trained variants, not a causal ablation of the corruption: the deployed model was trained with an earlier version of the training code (the later one also skips non-finite batches; its explicit restriction of mixture components to sampler-supported families was presumably already in the deployed run's unrecorded code, Appendix~\ref{app:training_data}), the direct run kept only its step-$240{,}000$ checkpoint, each checkpoint was selected by its own run, and each variant has one training seed, so differences of the size seen here may lie within training-seed variability. Confidence intervals over test rows, in the released results, do not capture that variability.

\begin{table}[t]
    \centering
    \caption{\textbf{Retained corruption variants.} Mean held-out KL (nats) and ISE over the $45$ fits of the Table~\ref{tab:bivariate} suite and the $27$ fits of the out-of-distribution suite (Table~\ref{tab:ood_families}), and held-out NLL (bits/dim) of each checkpoint's own D-vine on the MAF data with the rows of Table~\ref{tab:vine_nll}; lower is better, best in bold. Step: retained checkpoint (the direct run kept only step $240{,}000$, the multinomial run steps $190{,}000$ and $220{,}000$). One training run per variant.}
    \label{tab:corruption_variants}
    \scriptsize
    \resizebox{\columnwidth}{!}{\begin{tabular}{llcccccccc}
\toprule
 & & \multicolumn{2}{c}{Table~1 suite} & \multicolumn{2}{c}{Out-of-distribution} & \multicolumn{4}{c}{Own D-vine, held-out NLL (bits/dim)} \\
\cmidrule(lr){3-4}\cmidrule(lr){5-6}\cmidrule(lr){7-10}
Variant & Step & KL & ISE & KL & ISE & POWER & GAS & HEPMASS & MINIBOONE \\
\midrule
Uniform-mix (deployed) & 190k & 0.0061 & 1.72e-02 & 0.0117 & 0.0386 & \textbf{-0.1413} & -1.4773 & -0.2372 & \textbf{-0.9688} \\
Direct & 240k & 0.0063 & 1.54e-02 & \textbf{0.0115} & 0.0309 & -0.1330 & -1.4716 & -0.2242 & -0.9620 \\
Gaussian & 190k & 0.0061 & 1.98e-02 & 0.0132 & 0.0315 & -0.1379 & -1.4815 & -0.2311 & -0.9644 \\
Multinomial & 220k & 0.0063 & 2.02e-02 & 0.0118 & \textbf{0.0297} & -0.1389 & \textbf{-1.4908} & \textbf{-0.2385} & -0.9653 \\
Multinomial & 190k & \textbf{0.0060} & \textbf{1.44e-02} & 0.0118 & 0.0322 & -0.1356 & -1.4899 & -0.2313 & -0.9614 \\
\bottomrule
\end{tabular}
}
\end{table}

\begin{table}[t]
    \centering
    \caption{\textbf{Common real-data edge panel.} Held-out NLL (bits/dim) summed over all vine edges of each MAF dataset when every edge estimator sees the same training and test pseudo-observations (one D-vine order, reference TLL h-functions); lower is better, best in bold. Last column: panel edges on which the variant's held-out edge log-density exceeds the deployed model's.}
    \label{tab:corruption_edges}
    \small
    \resizebox{\columnwidth}{!}{\begin{tabular}{llccccc}
\toprule
Edge estimator & Step & POWER & GAS & HEPMASS & MINIBOONE & Edges better (of 1156) \\
\midrule
Uniform-mix (deployed) & 190k & -0.1417 & -1.4781 & -0.2327 & \textbf{-1.0184} & -- \\
Direct & 240k & -0.1332 & -1.4690 & -0.2223 & -0.9956 & 274 \\
Gaussian & 190k & -0.1386 & -1.4925 & -0.2278 & -1.0060 & 378 \\
Multinomial & 220k & -0.1393 & -1.4933 & \textbf{-0.2345} & -1.0095 & 427 \\
Multinomial & 190k & -0.1356 & -1.4887 & -0.2273 & -1.0093 & 380 \\
pyvine-TLL (reference) & -- & \textbf{-0.1525} & \textbf{-1.5579} & -0.1982 & -0.6689 & -- \\
\bottomrule
\end{tabular}
}
\end{table}

\subsection{UCI Density Benchmark Details}
\label{app:uci_density}

\paragraph{Data.} We use the four standard UCI density-estimation datasets with the preprocessing and train/validation/test splits of \citet{papamakarios2017maf}, rebuilt from the authors' public archive (Zenodo record 1161203) with their reference code; the split sizes match the published ones exactly (POWER $1{,}659{,}917/184{,}435/204{,}928$ rows, $d{=}6$; GAS $852{,}174/94{,}685/105{,}206$, $d{=}8$; HEPMASS $315{,}123/35{,}013/174{,}987$, $d{=}21$; MINIBOONE $29{,}556/3{,}284/3{,}648$, $d{=}43$). Every method is fitted on the same random subsample of $n_{\mathrm{train}}$ training rows and evaluated on up to $50{,}000$ test rows after the same empirical marginal transform, so the comparison isolates dependence modeling. The UCI transform is $(r+0.5)/(n_{\mathrm{train}}+1)$, clipped to $[10^{-6},1-10^{-6}]$, where $r$ counts training observations less than or equal to the value; the same fitted marginal transform is applied to test observations. We report held-out copula-space NLL in bits/dim with bootstrap $95\%$ confidence intervals over test rows (paired bootstrap for the differences in Table~\ref{tab:uci_matched}), fit time, and PIT-KS (the mean over coordinates of the Kolmogorov--Smirnov statistic of the Rosenblatt-transformed test rows) for vine-based methods. Table~\ref{tab:vine_nll} uses $n_{\mathrm{train}}{=}5{,}000$, comparable to the sample sizes of the pretraining distribution ($200$--$8{,}000$) and of the S\&P100 benchmark; Table~\ref{tab:vine_nll_50k} repeats the benchmark with $n_{\mathrm{train}}{=}50{,}000$ (all $29{,}556$ rows for MINIBOONE).

\paragraph{Evaluation provenance.}
\label{app:evaluation_provenance}
The UCI evaluations in arXiv v1--v2 and the reviewed submission used five mislabeled files: four were seeded Gaussian arrays and the POWER file held the Combined Cycle Power Plant (CCPP) data. Those results, the real-edge ACNet comparison and the depth-stratified study are withdrawn. The UCI likelihood, corruption, projection and tree-decomposition results reported here use the authenticated MAF splits above. Bivariate evaluation uses reference samplers and the stated unit-square ISE convention; the defective Joe case is excluded from auxiliary suites except the explicitly identified probit ablation. Sampling-based results use the validated inverse-h implementation. The projection-iteration study uses training-order coordinate channels and synchronized timings; the unsupported pre-training validity diagnostic is omitted. These corrections and their source records are retained in the paper repository.

\begin{table}[t]
    \centering
    \caption{\textbf{UCI copula-space density benchmark ($n_{\mathrm{train}}{=}5{,}000$).} Held-out NLL (bits/dim) with bootstrap $95\%$ CIs over test rows, fitting time, and PIT-KS on the MAF splits of the four UCI datasets (up to $50{,}000$ test rows; all $3{,}648$ for MINIBOONE). PIT-KS denotes the Kolmogorov--Smirnov statistic of held-out probability-integral-transform values against $\mathrm{Unif}(0,1)$. Frozen \ours{} checkpoint; baselines refitted per dataset. Fit times are wall-clock seconds, with \ours{} and RealNVP on one A100 GPU slice (MIG 3g.20gb) and pyvine on $4$ CPU threads, one benchmark running at a time. RealNVP's fixed budget (at most $25$ epochs at batch size $2{,}048$) allows at most $75$ optimizer updates at this size versus $550$ at $50{,}000$ rows, so its poor scores here (a GAS interval about $1$ bit/dim wide; worse on HEPMASS than the independence copula, whose NLL is $0$) reflect that budget rather than an intrinsic small-sample limit of flows; with $50{,}000$ rows it is the best method on every dataset (Table~\ref{tab:vine_nll_50k}). The intervals reflect test-row variability for one training subsample and seed, and the two tables use independent random training and test subsets (MINIBOONE uses all its test rows in both, and all its training rows at $50{,}000$). \ours{} fits a greedy-order D-vine and the pyvine baselines select regular vines (Section~\ref{sec:experiments}).}
    \label{tab:vine_nll}
    \scriptsize
    \resizebox{\columnwidth}{!}{\begin{tabular}{llccc}
\toprule
Dataset & Method & NLL (bits/dim) [95\% CI] $\downarrow$ & Fit (s) $\downarrow$ & PIT-KS $\downarrow$ \\
\midrule
POWER ($d=6$) & VDC (Ours) & -0.1413 [-0.1431, -0.1395] & 0.56 & 0.017 \\
 & pyvine-param & -0.1078 [-0.1097, -0.1059] & 10.20 & 0.023 \\
 & pyvine-TLL & \textbf{-0.1822 [-0.1844, -0.1800]} & 0.97 & 0.017 \\
 & Flow (RealNVP) & -0.1221 [-0.1244, -0.1199] & 1.06 & -- \\
 & Gaussian copula & -0.0685 [-0.0699, -0.0669] & 0.00 & -- \\
\midrule
GAS ($d=8$) & VDC (Ours) & -1.4773 [-1.4816, -1.4732] & 0.21 & 0.049 \\
 & pyvine-param & -1.4040 [-1.4185, -1.3886] & 18.71 & 0.059 \\
 & pyvine-TLL & \textbf{-1.7050 [-1.7115, -1.6984]} & 1.97 & 0.043 \\
 & Flow (RealNVP) & -1.0991 [-1.5248, -0.5283] & 1.00 & -- \\
 & Gaussian copula & -1.2472 [-1.2694, -1.2206] & 0.00 & -- \\
\midrule
HEPMASS ($d=21$) & VDC (Ours) & \textbf{-0.2372 [-0.2385, -0.2359]} & 1.17 & 0.021 \\
 & pyvine-param & -0.1628 [-0.1641, -0.1616] & 125.84 & 0.020 \\
 & pyvine-TLL & -0.2009 [-0.2020, -0.1997] & 9.66 & 0.018 \\
 & Flow (RealNVP) & 0.2538 [0.2523, 0.2553] & 0.22 & -- \\
 & Gaussian copula & -0.1039 [-0.1050, -0.1029] & 0.01 & -- \\
\midrule
MINIBOONE ($d=43$) & VDC (Ours) & \textbf{-0.9689 [-0.9771, -0.9609]} & 4.49 & 0.027 \\
 & pyvine-param & -0.9334 [-0.9417, -0.9255] & 503.01 & 0.028 \\
 & pyvine-TLL & -0.7444 [-0.7520, -0.7366] & 35.72 & 0.024 \\
 & Flow (RealNVP) & -0.4145 [-0.4212, -0.4077] & 1.03 & -- \\
 & Gaussian copula & -0.8853 [-0.8984, -0.8706] & 0.01 & -- \\
\bottomrule
\end{tabular}
}
\end{table}

\IfFileExists{tables/tab_vine_nll_ci_50k.tex}{%
\begin{table}[t]
    \centering
    \caption{\textbf{UCI copula-space density benchmark ($n_{\mathrm{train}}{=}50{,}000$).} As Table~\ref{tab:vine_nll} with $50{,}000$ training rows (all $29{,}556$ for MINIBOONE).}
    \label{tab:vine_nll_50k}
    \scriptsize
    \resizebox{\columnwidth}{!}{\begin{tabular}{llccc}
\toprule
Dataset & Method & NLL (bits/dim) [95\% CI] $\downarrow$ & Fit (s) $\downarrow$ & PIT-KS $\downarrow$ \\
\midrule
POWER ($d=6$) & VDC (Ours) & -0.1495 [-0.1513, -0.1476] & 0.82 & 0.014 \\
 & pyvine-param & -0.1064 [-0.1083, -0.1047] & 109.67 & 0.022 \\
 & pyvine-TLL & -0.2169 [-0.2194, -0.2146] & 12.82 & 0.010 \\
 & Flow (RealNVP) & \textbf{-0.2280 [-0.2308, -0.2252]} & 7.63 & -- \\
 & Gaussian copula & -0.0679 [-0.0694, -0.0664] & 0.01 & -- \\
\midrule
GAS ($d=8$) & VDC (Ours) & -1.4926 [-1.4968, -1.4886] & 0.79 & 0.045 \\
 & pyvine-param & -1.4305 [-1.4347, -1.4264] & 191.34 & 0.058 \\
 & pyvine-TLL & -1.8113 [-1.8174, -1.8044] & 30.88 & 0.036 \\
 & Flow (RealNVP) & \textbf{-2.1343 [-2.1392, -2.1297]} & 6.45 & -- \\
 & Gaussian copula & -1.2498 [-1.2714, -1.2246] & 0.02 & -- \\
\midrule
HEPMASS ($d=21$) & VDC (Ours) & -0.2468 [-0.2481, -0.2454] & 5.89 & 0.020 \\
 & pyvine-param & -0.1657 [-0.1671, -0.1645] & 1201.63 & 0.018 \\
 & pyvine-TLL & -0.2653 [-0.2665, -0.2640] & 136.39 & 0.012 \\
 & Flow (RealNVP) & \textbf{-0.2898 [-0.2915, -0.2881]} & 7.73 & -- \\
 & Gaussian copula & -0.1041 [-0.1051, -0.1031] & 0.06 & -- \\
\midrule
MINIBOONE ($d=43$) & VDC (Ours) & -0.9784 [-0.9866, -0.9702] & 15.75 & 0.025 \\
 & pyvine-param & -0.9618 [-0.9699, -0.9538] & 2996.58 & 0.026 \\
 & pyvine-TLL & -0.8716 [-0.8797, -0.8635] & 278.59 & 0.019 \\
 & Flow (RealNVP) & \textbf{-1.0861 [-1.0965, -1.0761]} & 4.58 & -- \\
 & Gaussian copula & -0.8901 [-0.9012, -0.8777] & 0.07 & -- \\
\bottomrule
\end{tabular}
}
\end{table}
}{}

\paragraph{Structure-matched comparison.}
\label{app:uci_matched}
\ours{} fits a D-vine with a greedy $|\tau|$ order, while the pyvine baselines select regular vines with Di{\ss}mann's algorithm, so Table~\ref{tab:vine_nll} compares complete pipelines. Table~\ref{tab:uci_matched} separates structure from edge estimation on the same training and test rows ($n_{\mathrm{train}}{=}5{,}000$; the refitted \ours{} D-vine reproduces the per-row log-densities of Table~\ref{tab:vine_nll} exactly). It refits pyvine-TLL on \ours{}'s D-vine order, and \ours{} with the library's Di{\ss}mann regular-vine selection. Every ranking is unchanged. On POWER and GAS about a third of pyvine-TLL's advantage comes from its regular-vine structure, on MINIBOONE the advantage of \ours{} grows when the structures match, and \ours{} itself gains from a regular vine on three of the four datasets, so the D-vine default handicaps it.

\IfFileExists{tables/tab_uci_matched_structure.tex}{%
\begin{table}[t]
    \centering
    \caption{\textbf{Structure-matched UCI comparison} ($n_{\mathrm{train}}{=}5{,}000$, same rows as Table~\ref{tab:vine_nll}). Paired NLL differences in bits/dim with paired-bootstrap $95\%$ CIs over test rows; positive means \ours{} is worse. Columns 2--4: \ours{} minus pyvine-TLL as in Table~\ref{tab:vine_nll} (greedy D-vine vs.\ Di{\ss}mann regular vine), with both on \ours{}'s D-vine order, and with both on Di{\ss}mann regular vines. Last column: \ours{} with a regular vine minus \ours{} with its D-vine.}
    \label{tab:uci_matched}
    \scriptsize
    \resizebox{\columnwidth}{!}{\begin{tabular}{lcccc}
\toprule
 & \multicolumn{3}{c}{\ours{} $-$ pyvine-TLL} & \ours{} regular $-$ D-vine \\
\cmidrule(lr){2-4}
Dataset & as in Table~\ref{tab:vine_nll} & both D-vine & both regular vine & \\
\midrule
POWER ($d=6$) & +0.041 [+0.040, +0.042] & +0.029 [+0.027, +0.030] & +0.037 [+0.036, +0.039] & -0.003 [-0.004, -0.003] \\
GAS ($d=8$) & +0.228 [+0.223, +0.232] & +0.152 [+0.148, +0.156] & +0.151 [+0.146, +0.155] & -0.077 [-0.079, -0.076] \\
HEPMASS ($d=21$) & -0.036 [-0.037, -0.035] & -0.036 [-0.037, -0.035] & -0.036 [-0.037, -0.035] & +0.000 [-0.000, +0.000] \\
MINIBOONE ($d=43$) & -0.225 [-0.230, -0.219] & -0.295 [-0.300, -0.290] & -0.257 [-0.262, -0.252] & -0.032 [-0.035, -0.029] \\
\bottomrule
\end{tabular}
}
\end{table}
}{}

\IfFileExists{tables/tab_uci_dependence_diag.tex}{%
\subsection{How Much Copula Dependence Do the UCI Benchmarks Carry?}
\label{app:uci_dependence_diag}

To calibrate how much dependence each benchmark carries, let $R$ be the correlation matrix fitted on probit-transformed training PIT scores. Table~\ref{tab:uci_dependence_diag} reports pairwise Kendall's~$\tau$ summaries, the closed-form Gaussian-copula total correlation $\mathrm{TC}_{\mathrm{Gauss}} = -\tfrac{1}{2}\log\det R$, and the held-out Gaussian-copula log-likelihood in nats per row (the independence copula scores exactly zero), with $\tau$ computed on $4{,}000$ of the training rows. All four datasets carry clear dependence, from $\mathrm{TC}_{\mathrm{Gauss}}=0.28$ nats on POWER to $26.8$ nats on MINIBOONE, and GAS contains near-deterministic sensor pairs (maximum $|\tau|=0.89$), the regime where a $64\times64$ grid is least accurate (Appendix~\ref{app:grid_floor}).

\begin{table}[t]
    \centering
    \caption{\textbf{UCI copula dependence diagnostic.} Pairwise $|\tau|$ summaries (over all pairs, except $300$ randomly sampled of the $903$ MINIBOONE pairs), Gaussian-copula total correlation, and held-out Gaussian-copula log-likelihood (nats per row) with the corresponding NLL in bits/dim, under the same PIT protocol as Table~\ref{tab:vine_nll}.}
    \label{tab:uci_dependence_diag}
    \footnotesize
    \begin{tabular}{lrccccc}
\toprule
Dataset & $d$ & mean $|\tau|$ & max $|\tau|$ & $\mathrm{TC}_{\mathrm{Gauss}}$ (nats) & Gauss.\ LL (nats/row) & Gauss.\ NLL (bits/dim) \\
\midrule
POWER & 6 & 0.106 & 0.399 & 0.275 & 0.2849 & -0.0685 \\
GAS & 8 & 0.415 & 0.888 & 7.126 & 6.9159 & -1.2472 \\
HEPMASS & 21 & 0.040 & 0.488 & 1.543 & 1.5129 & -0.1039 \\
MINIBOONE & 43 & 0.170 & 0.843 & 26.822 & 26.3858 & -0.8853 \\
\midrule
\multicolumn{7}{l}{\scriptsize Independence copula scores exactly $0$ nats/row ($0.0000$ bits/dim) by definition.} \\
\bottomrule
\end{tabular}

\end{table}
}{}

\subsection{RealNVP Capacity Sensitivity}
\label{app:flow_capacity_sensitivity}

The main density table uses a fixed-budget RealNVP baseline so that fit times remain comparable across all datasets. To check whether the synthetic Clayton-vine result is simply due to a limited flow training budget, we ran a targeted sensitivity study ($n_{\mathrm{train}}{=}5{,}000$, $2{,}000$ test rows, $d\in\{20,50\}$, batch size $2048$) with two stronger variants: a longer 8-block flow trained up to 200 epochs, and a wider 12-block flow with 256 hidden units. Table~\ref{tab:flow_sensitivity_appendix} reports the result: on the non-Gaussian Clayton-vine benchmark, increasing flow capacity and training budget does not close the gap to \ours{} at either $d=20$ or $d=50$. On the real MAF data at $n_{\mathrm{train}}{=}50{,}000$, the fixed-budget flow outperforms all three vine methods on all four datasets (Table~\ref{tab:vine_nll_50k}). We therefore treat RealNVP as a strong global density reference, not as a weak baseline, while using the Clayton-vine stress test to isolate the value of structured vine factorization.

\begin{table}[t]
    \centering
    \caption{\textbf{RealNVP capacity sensitivity.} Lower NLL is better. Rows include VDC and Gaussian-copula references on the same Clayton-vine splits (five runs each); NLL is mean $\pm$ standard deviation across these runs. ``Paper budget'' is the RealNVP configuration used in the main tables.}
    \label{tab:flow_sensitivity_appendix}
    \footnotesize
    \IfFileExists{tables/tab_flow_sensitivity.tex}{
        \begin{tabular}{lccrcc}
\toprule
Method & Variant & $d$ & Runs & NLL (bits/dim) $\downarrow$ & Fit (s) $\downarrow$ \\
\midrule
VDC & canonical & 20 & 5 & \textbf{-0.5835$\pm$0.0117} & 1.4 \\
Gaussian copula & closed form & 20 & 5 & -0.4443$\pm$0.0095 & \textbf{0.0} \\
RealNVP & paper budget & 20 & 5 & -0.4647$\pm$0.0075 & 1.0 \\
RealNVP & long & 20 & 5 & -0.4694$\pm$0.0087 & 1.7 \\
RealNVP & wide & 20 & 5 & -0.4659$\pm$0.0100 & 1.7 \\
\midrule
VDC & canonical & 50 & 5 & \textbf{-0.5965$\pm$0.0077} & 4.1 \\
Gaussian copula & closed form & 50 & 5 & -0.4539$\pm$0.0092 & \textbf{0.0} \\
RealNVP & paper budget & 50 & 5 & -0.3809$\pm$0.0086 & 0.9 \\
RealNVP & long & 50 & 5 & -0.3839$\pm$0.0088 & 1.4 \\
RealNVP & wide & 50 & 5 & -0.3604$\pm$0.0083 & 1.4 \\
\bottomrule
\end{tabular}

    }{
        \fbox{\parbox{0.95\columnwidth}{\centering\textbf{[Flow-sensitivity table missing]}\\Run the supplemental evaluation script to produce the flow-sensitivity table.}}
    }
\end{table}

\subsection{Non-Simplified Conditional Pair-Copula Stress Test}
\label{app:nonsimplified_stress}

To make the simplifying-assumption boundary explicit without giving Gaussian baselines an overly favorable benchmark, we constructed two three-variable copulas in which $U_1 \sim \mathrm{Unif}(0,1)$ and the conditional copula of $(U_2,U_3)\mid U_1$ changes with the conditioning value.
The first is a Gaussian sign-flip diagnostic: the conditional correlation is $\rho=-0.75$ for $U_1<0.5$ and $\rho=0.75$ for $U_1\geq0.5$.
The second is a non-Gaussian tail-switch diagnostic: a Clayton copula with lower-tail dependence for $U_1<0.5$ and a Gumbel copula with upper-tail dependence for $U_1\geq0.5$, both with Kendall's $\tau \approx 0.6$.
At threshold $q$, we report the empirical tail concentrations $\lambda_L(q)=\PP(U_2\leq q,U_3\leq q)/q$ and $\lambda_U(q)=\PP(U_2>1-q,U_3>1-q)/q$.
Both cases are deliberately outside the simplified-vine setting used in the paper, in line with the approximation discussion of \citet{haff2010simplified} and the formal characterization of \citet{spanhel2019simplified}.
Table~\ref{tab:nonsimplified_stress} compares the true conditional oracle, a regime-wise Gaussian fit, and the simplified pooled edge estimators used elsewhere in the paper (one draw with $n_{\mathrm{train}}{=}2000$ and $n_{\mathrm{test}}{=}10{,}000$; here the pyvine fits exclude the independence copula and use a $64$-point grid).
In the Gaussian sign-flip case, conditioning solves the problem and the pooled edge necessarily pays an excess-NLL cost.
Among the simplified baselines, however, VDC has the lowest excess NLL in this single draw ($0.279$ versus $0.291$ for pyvine-parametric, $0.297$ for pyvine-TLL, $0.418$ for Gaussian, and $0.425$ for the histogram).
In the non-Gaussian tail-switch case, VDC is again the best simplified pooled estimator ($0.076$ excess NLL), beating pyvine-parametric ($0.088$), pyvine-TLL ($0.090$), pooled Gaussian ($0.118$), and the histogram ($0.322$); it also beats the regime-wise Gaussian fit ($0.117$), so a flexible pooled estimator beats even a Gaussian model that conditions on the regime.
The remaining gap to the conditional oracle is the intended boundary: extending the amortized edge operator to conditional pair-copulas is a natural route to non-simplified vines.
Figure~\ref{fig:nonsimplified_stress}(a) visualizes the excess-NLL comparison, while panels (b)--(c) show the regime-wise sign reversal and tail-asymmetry flip that make pooling hard; Table~\ref{tab:nonsimplified_stress} gives the exact values.

\begin{figure}[!htbp]
    \centering
    \includegraphics[width=\textwidth]{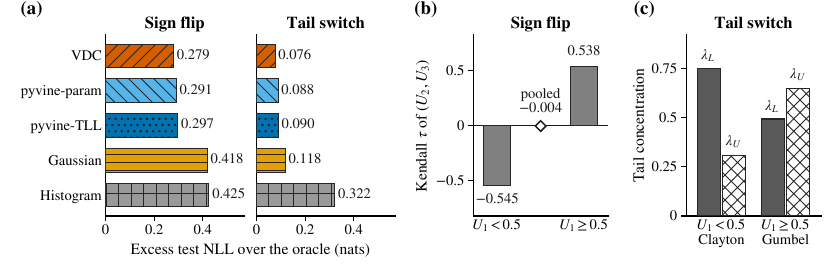}
    \caption{\textbf{Non-simplified stress tests.} (a) Excess test NLL (nats per test pair) over the true conditional oracle for the five simplified pooled edge estimators on the two regime-switching cases (single seed); each estimator keeps its color from the main figures, and VDC is the best pooled estimator in both. (b) Empirical Kendall's $\tau$ of $(U_2,U_3)$ within each conditioning regime ($U_1<0.5$ vs.\ $U_1\geq 0.5$) for the Gaussian sign-flip case: the conditional dependence reverses sign while the pooled $\tau\approx 0$ (diamond). (c) Empirical lower- and upper-tail concentrations ($\lambda_L$ solid, $\lambda_U$ cross-hatched; $q{=}0.1$) per regime for the Clayton/Gumbel tail switch: tail asymmetry flips between regimes at matched $\tau\approx 0.6$. The remaining gap to the oracle in (a) marks the simplifying-assumption boundary.}
    \label{fig:nonsimplified_stress}
\end{figure}

\begin{table}[t]
    \centering
    \caption{\textbf{Non-simplified conditional edge stress tests.} Test and excess NLL in nats per test pair; lower is better. The oracle and regime Gaussian can condition on $U_1$; the remaining estimators fit one simplified pooled edge density for $(U_2,U_3)$. The Clayton/Gumbel setting tests non-Gaussian tail geometry rather than only correlation sign.}
    \label{tab:nonsimplified_stress}
    \small
    \IfFileExists{tables/tab_nonsimplified_stress.tex}{
        \begin{tabular}{llccc}
\toprule
Estimator & Conditions on $U_1$? & Test NLL $\downarrow$ & Excess NLL & Fitted model \\
\midrule
\multicolumn{5}{l}{\emph{Gaussian sign flip}}\\
Oracle & yes & -0.418 & 0.000 & true conditional density \\
Regime Gaussian & yes & -0.416 & 0.002 & $\rho = -0.75 / 0.77$ \\
VDC & no & -0.139 & 0.279 & pooled grid \\
pyvine-param & no & -0.128 & 0.291 & Student-$t$ \\
pyvine-TLL & no & -0.121 & 0.297 & TLL \\
Gaussian & no & 0.000 & 0.418 & $\rho \approx 0$ \\
Histogram & no & 0.007 & 0.425 & pooled grid \\
\addlinespace[0.25em]
\multicolumn{5}{l}{\emph{Clayton/Gumbel tail switch}}\\
Oracle & yes & -0.596 & 0.000 & true conditional density \\
Regime Gaussian & yes & -0.478 & 0.117 & $\rho = 0.79 / 0.81$ \\
VDC & no & -0.520 & 0.076 & pooled grid \\
pyvine-param & no & -0.507 & 0.088 & BB1 \\
pyvine-TLL & no & -0.506 & 0.090 & TLL \\
Gaussian & no & -0.478 & 0.118 & $\rho = 0.80$ \\
Histogram & no & -0.274 & 0.322 & pooled grid \\
\bottomrule
\end{tabular}

    }{
        \fbox{\parbox{0.95\columnwidth}{\centering\textbf{[Non-simplified stress table missing]}}}
    }
\end{table}

\begin{figure}[!htbp]
    \centering
    \includegraphics[width=\textwidth]{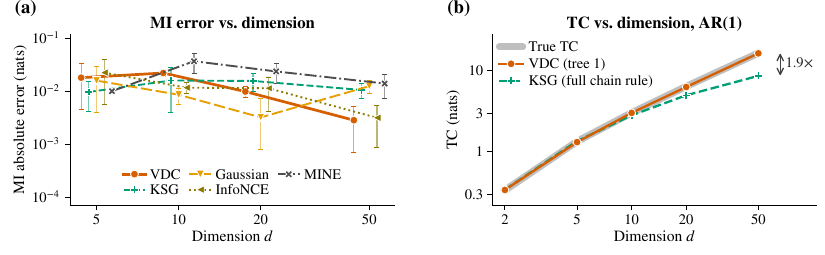}
    \caption{\textbf{MI and TC versus dimension.} (a) MI absolute error for the first adjacent pair of a Gaussian AR(1) process ($\rho{=}0.6$, $n{=}2000$) at each $d$; the pair's distribution does not depend on $d$, so the panel shows replicate estimates rather than a dimension effect; mean $\pm1$ standard deviation over $3$ trials (markers offset for legibility; lower whiskers truncated on the log axis); pooled values, including NWJ (not plotted), are in Table~\ref{tab:theory_mi_methods_appendix}. (b) TC versus dimension for a Gaussian AR(1) process with lag-one correlation $\rho{=}0.7$ ($n{=}5000$, mean over three trials for each of seeds $\{7,17,42\}$): \ours{}, with a D-vine truncated after its first tree (exact for this first-order Markov process), tracks the true TC, while the full chain-rule KSG estimator underestimates it by roughly a factor of two at $d{=}50$ ($1.9\times$); KSG summed over the same adjacent pairs recovers it (Appendix~\ref{par:tc_tree1_control}).}
    \label{fig:dimension_appendix}
\end{figure}

\subsection{MI Benchmark Numbers}

For completeness, Table~\ref{tab:mi_appendix} reports the exact mean absolute MI errors for the bivariate synthetic-copula comparison discussed in the main text. The appendix MI baselines include Gaussian, KSG, MINE, InfoNCE, the Nguyen--Wainwright--Jordan (NWJ) variational bound \citep{nguyen2010estimating}, and MINDE (Mutual Information Neural Diffusion Estimation).

\begin{table}[t]
    \centering
    \caption{\textbf{Bivariate MI estimation.} Mean absolute error (MAE, nats) and runtime per estimate on eight synthetic copulas with MI from numerical integration of the analytic density on a $256\times256$ grid (independence, Gaussian $\rho{=}\pm0.7$, Student-$t$, Clayton at two rotations, Gumbel, Frank; $n{=}5000$); mean $\pm$ standard deviation over three seeds where available.}
    \label{tab:mi_appendix}
    \small
    \IfFileExists{tables/tab_mi.tex}{
        \makebox[\columnwidth][c]{\scalebox{0.92}{\begin{tabular}{lccc}
\toprule
Method & MAE (nats) & Time (s) & Rank \\
\midrule
\textbf{VDC (Ours)} & \textbf{0.0106 $\pm$ 0.0050} & 0.36 $\pm$ 0.02 & \textbf{1} \\
MINE & 0.0135 (1 seed) & 2.65 (1 seed) & 2 \\
NWJ & 0.0157 (1 seed) & 3.09 (1 seed) & 3 \\
KSG & 0.0161 $\pm$ 0.0026 & 0.08 $\pm$ 0.00 & 4 \\
InfoNCE & 0.0170 (1 seed) & 4.88 (1 seed) & 5 \\
Gaussian & 0.0539 $\pm$ 0.0022 & \textbf{0.00 $\pm$ 0.00} & 6 \\
MINDE & 0.3915 (1 seed) & 5.97 (1 seed) & 7 \\
\bottomrule
\end{tabular}
}}
    }{
        \fbox{\parbox{0.95\columnwidth}{\centering\textbf{[MI table missing]}\\Run the table-generation script to produce the MI table.}}
    }
\end{table}

\subsection{Self-Consistency Protocol}
\label{app:self_consistency}

We now specify the protocol used to produce Table~\ref{tab:self_consistency}. All three tests use the same generators and the same fixed VDC checkpoint, so that DPI, additivity, and monotone invariance are diagnostics of one estimator rather than of separate models. The procedure is implemented in the released code and run with default arguments unless stated: $n=10{,}000$ samples per pair, $n_{\text{trials}}=5$ random repetitions per condition, random-number-generator seed $42$, and KSG with neighborhood size $k=5$.

\paragraph{DPI test.}
Let $(X,Y)$ be drawn from one of four bivariate copulas: Gaussian ($\rho{=}0.7$), Student-$t$ ($\rho{=}0.7,\nu{=}5$), Clayton ($\theta{=}3$), and Frank ($\theta{=}5$). For a Gaussian-noise standard deviation $\sigma$, define $Y_\sigma$ by mapping $Y$ through the standard-normal quantile, adding that noise, and pushing the result back through the standard-normal CDF. This gives the Markov chain $X \to Y \to Y_\sigma$, where $Y_\sigma$ is a strictly noisier copy of $Y$, so the data-processing inequality requires $I(X;Y_\sigma) \leq I(X;Y)$. We sweep $\sigma \in \{0.1, 0.3, 0.5\}$, so that the total number of trials is $4 \times n_{\text{trials}} \times 3 = 60$ at the default settings. Set the violation tolerance to $\tau = 10^{-3}$ nats; a trial is counted as a \emph{violation} if $\hat I(X;Y_\sigma) > \hat I(X;Y) + \tau$, and the reported violation rate is the fraction of violating trials. For the 0/60 outcome of the original protocol we report the Wilson 95\% upper confidence bound; with 60 trials this is approximately $6.0\%$.

\paragraph{Additivity test.}
For two \emph{independent} bivariate copulas $(X_1,Y_1)$ and $(X_2,Y_2)$, with families drawn uniformly from $\{$Gaussian, Clayton, Frank$\}$, the chain rule gives $I((X_1,X_2);(Y_1,Y_2)) = I(X_1;Y_1) + I(X_2;Y_2)$ at the population level. Define the probability-integral-transform variables $U_i=F_{X_i}(X_i)$ and $V_i=F_{Y_i}(Y_i)$. We compute the right-hand side (RHS) as the sum of two bivariate MI estimates and the left-hand side (LHS) as $\mathrm{TC}(U_1,U_2,V_1,V_2)$ on the four-dimensional copula via VDC's D-vine total correlation with the vine truncated after its first tree, exploiting independence across blocks. The reported additivity error is the mean absolute deviation $|\mathrm{LHS}-\mathrm{RHS}|$ over $n_{\text{trials}}$ trials with $n=4000$ per pair. We note a protocol asymmetry: the block D-vine used for the LHS encodes independence across the two blocks, which gives VDC structural side information that KSG does not receive; the comparison should be read with that caveat.

\paragraph{Monotone invariance test.}
For each of three base copulas (Gaussian, Student-$t$, Clayton) we apply each of two strictly increasing marginal transforms, $g_1(u)=u^3$ and $g_2(u)=(e^u-1)/(e-1)$, to both $X$ and $Y$. Mutual information is invariant under any homeomorphic marginal transform; for $g\in\{g_1,g_2\}$, the reported monotone error is the mean of $|\hat I(g(X);g(Y)) - \hat I(X;Y)|$ over the resulting $3 \times n_{\text{trials}} \times 2 = 30$ trials at $n=10{,}000$.

The protocol intentionally tests structural properties that any well-behaved MI estimator should satisfy. It is an empirical sanity check rather than a formal guarantee: a low violation rate does not establish bias-free estimation, but a high violation rate or a large additivity error indicates an estimator whose outputs are not internally coherent across operations that should preserve or decompose information. These are properties that the true MI satisfies exactly. Both estimators rank-transform their inputs, so monotone invariance holds by construction (exactly for \ours{}, and for KSG with $k{=}5$ to within about $10^{-4}$ on average and $4\times10^{-4}$ per trial); the DPI and additivity tests probe the estimates themselves.

\begin{table}[t]
    \centering
    \caption{\textbf{MI self-consistency checks.} Columns report the data-processing-inequality (DPI) violation rate, additivity error, and monotone-invariance error defined in Appendix~\ref{app:self_consistency}. VDC shows no observed DPI violations and lower additivity error than KSG under this protocol; at these trial counts the DPI contrast alone is not significant, and the scaled-up protocol in Table~\ref{tab:self_consistency_scaleup} ($480$ trials) is reported separately.}
    \label{tab:self_consistency}
    \small
    \begingroup
    \setlength{\tabcolsep}{4pt}
    \IfFileExists{tables/tab_self_consistency.tex}{
        \resizebox{0.70\columnwidth}{!}{\begin{tabular}{lccc}
\toprule
Method & DPI Viol. (\%) $\downarrow$ & Add. Err. (nats) $\downarrow$ & Mono. Err. (nats) $\downarrow$ \\
\midrule
KSG & 3.3 & 0.027 & 0.000 \\
VDC & 0.0 & 0.021 & 0.000 \\
\bottomrule
\end{tabular}
}
    }{
        \fbox{\parbox{0.9\columnwidth}{\centering\textbf{[Self-consistency table missing]}}}
    }
    \endgroup
\end{table}

\IfFileExists{tables/tab_self_consistency_scaleup.tex}{%
\paragraph{Scaled-up protocol.}
Table~\ref{tab:self_consistency_scaleup} repeats the protocol with $40$ repetitions per condition, giving $480$ noise-level evaluations across $160$ independent base datasets per estimator. The base-dataset comparison gives Fisher exact $p=0.011$. Trials that share a base dataset are dependent, so we assess significance over the $160$ independent base datasets; no base dataset violates at more than one noise level. The VDC and KSG trials use different random draws, so the comparison is two-sample rather than paired.

\begin{table}[t]
    \centering
    \caption{\textbf{Scaled-up MI self-consistency checks.} Same DPI-violation, additivity-error, and monotone-invariance-error metrics as Table~\ref{tab:self_consistency}, with $8\times$ the number of trials per condition. At this scale the DPI contrast is significant: the $480$ trials are $160$ independent base datasets at three noise levels, and $10$ versus $25$ base datasets contain a violation (Fisher exact $p=0.011$; cluster permutation $p=0.011$; pooled over trials, $10/480$ vs.\ $25/480$).}
    \label{tab:self_consistency_scaleup}
    \small
    \makebox[\columnwidth][c]{\scalebox{0.92}{\begin{tabular}{lccc}
\toprule
Method & DPI Viol. (\%) $\downarrow$ & Add. Err. (nats) $\downarrow$ & Mono. Err. (nats) $\downarrow$ \\
\midrule
KSG & 5.2 & 0.044 & 0.000 \\
VDC & 2.1 & 0.027 & 0.000 \\
\bottomrule
\end{tabular}
}}
\end{table}
}{}

\IfFileExists{tables/tab_dpi_tolerance.tex}{%
\paragraph{Tolerance sensitivity.}
The violation threshold $\tau$ is a free parameter of the DPI test, so Table~\ref{tab:dpi_tolerance} reports the violation counts over the full tolerance range rather than the single default. The ordering (\ours{} fewer violations) holds at every $\tau$ including the strictest $\tau=0$, and the contrast is significant for all $\tau \leq 3\times10^{-3}$ nats; at the largest tabulated tolerance $\tau=10^{-2}$, \ours{} records zero violations while KSG retains $4/480$.

\begin{table}[t]
    \centering
    \caption{\textbf{DPI tolerance sensitivity} on the $480$-trial protocol. Violation counts and Fisher exact $p$ as the tolerance $\tau$ varies; the paper's default is $\tau=10^{-3}$ nats. The $p$-values pool trials; because no base dataset violates at more than one noise level, the same counts out of $160$ base datasets give equal or smaller $p$-values.}
    \label{tab:dpi_tolerance}
    \small
    \makebox[\columnwidth][c]{\scalebox{0.92}{\begin{tabular}{lccc}
\toprule
Tolerance $\tau$ (nats) & VDC violations & KSG violations & Fisher $p$ \\
\midrule
0 & 13/480 (2.7\%) & 32/480 (6.7\%) & 0.005 \\
$10^{-4}$ & 13/480 (2.7\%) & 29/480 (6.0\%) & 0.017 \\
$3\times10^{-4}$ & 13/480 (2.7\%) & 27/480 (5.6\%) & 0.035 \\
$10^{-3}$ (paper) & 10/480 (2.1\%) & 25/480 (5.2\%) & 0.015 \\
$3\times10^{-3}$ & 5/480 (1.0\%) & 16/480 (3.3\%) & 0.025 \\
$10^{-2}$ & 0/480 (0.0\%) & 4/480 (0.8\%) & 0.124 \\
\bottomrule
\end{tabular}
}}
\end{table}
}{}

\IfFileExists{tables/tab_additivity_learned.tex}{%
\paragraph{Additivity with learned structure.}
The additivity test's left-hand side uses a block D-vine that encodes the true across-block independence---structural side information KSG does not receive. Table~\ref{tab:additivity_learned} removes the block information by re-estimating the four-dimensional TC with a Di{\ss}mann-learned R-vine (truncated after its first tree, as throughout this protocol). The block-D-vine VDC estimate, learned-structure VDC estimate, and KSG estimate use identical data in each trial. The learned-structure additivity error matches the block-D-vine one, so the block information does not drive the gap to KSG, although both VDC estimates keep the first-tree truncation that KSG does not use; KSG's known bias growth with dimension plausibly contributes to its residual, and we retain that caveat.

\begin{table}[t]
    \centering
    \caption{\textbf{Additivity error with block versus learned vine structure} ($120$ paired trials; same generator and $n{=}4000$ as the main protocol). Entries are mean $\pm$ standard deviation across the $120$ trials. Learned Di{\ss}mann structure reproduces the block-D-vine result.}
    \label{tab:additivity_learned}
    \small
    \makebox[\columnwidth][c]{\scalebox{0.92}{\begin{tabular}{lcc}
\toprule
Estimator & Additivity err.\ (nats) $\downarrow$ & Trials \\
\midrule
VDC (oracle structure) & $0.024 \pm 0.020$ & 120 \\
VDC (learned structure) & $0.024 \pm 0.020$ & 120 \\
KSG & $0.039 \pm 0.029$ & 120 \\
\bottomrule
\end{tabular}
}}
\end{table}
}{}

\subsection{Bivariate Benchmark Dispersion}

For completeness, Table~\ref{tab:bivariate_dispersion_appendix} reports mean $\pm$ standard deviation across the held-out copula cases underlying the main bivariate benchmark in Table~\ref{tab:bivariate}. This complements the suite-average main-text table with the same method ranking. The reported standard deviation is dispersion \emph{across heterogeneous cases} (spanning weak to strong dependence and light to heavy tails), not run-to-run noise; the replication study below separates the two sources.

\IfFileExists{tables/tab_bivariate_replicates.tex}{%
\paragraph{Large-replication paired study.}
Table~\ref{tab:bivariate_replicates} reports a $100$-replication-per-family paired benchmark: for each of eight families included in pretraining, $100$ fresh $n{=}2000$ samples are drawn and both \ours{} and pyvine-TLL are fit on the identical sample, giving per-replication win rates with Wilson $95\%$ CIs. Pooled over $800$ paired replications, \ours{} wins $87.9\%$ $[85.4, 90.0]$ on density ISE and $88\%$ $[85, 90]$ on MI error; the one exception (BB7 density ISE) is reported as-is. Within-case replication noise is small (e.g., Clayton ISE $4.4\times10^{-1} \pm 9.1\times10^{-3}$, $\approx 2\%$ relative), confirming that the large standard deviations in Table~\ref{tab:bivariate_dispersion_appendix} reflect across-case heterogeneity.

\begin{table}[t]
    \centering
    \caption{\textbf{Paired replication study} ($100$ fresh pyvinecopulib replications per family; both methods fit on identical samples; $n{=}2000$). Errors are mean $\pm$ standard deviation across the $100$ replications within each family; win rates have Wilson $95\%$ CIs. ISE is integrated over the unit square against the normalized cell-center density of the true copula. Under exact cell-average truth, with TLL also evaluated as cell averages, \ours{} wins $753/800$ ISE comparisons, and on held-out likelihood with TLL's own density it wins $605/800$ (BB7 $2/100$, BB1 $53/100$).}
    \label{tab:bivariate_replicates}
    \footnotesize
    \begin{tabular}{lccc}
\toprule
\multicolumn{4}{l}{ISE $\downarrow$} \\
Case & VDC & pyvine-TLL & VDC win\% [95\% CI] \\
\midrule
Gaussian ($\rho{=}0.6$) & \textbf{2.5e-03 $\pm$ 2.2e-03} & 1.4e-02 $\pm$ 4.9e-03 & 99\% [95, 100] \\
Student-$t$ ($\rho{=}0.6,\,\nu{=}5$) & \textbf{5.2e-03 $\pm$ 4.0e-03} & 2.3e-02 $\pm$ 9.4e-03 & 98\% [93, 99] \\
Clayton ($\theta{=}3$) & \textbf{4.4e-01 $\pm$ 9.1e-03} & 6.7e-01 $\pm$ 9.6e-02 & 100\% [96, 100] \\
Gumbel ($\theta{=}2.5$) & \textbf{6.1e-02 $\pm$ 5.7e-03} & 1.5e-01 $\pm$ 4.6e-02 & 97\% [92, 99] \\
Frank ($\theta{=}5$) & \textbf{1.4e-03 $\pm$ 1.2e-03} & 1.5e-02 $\pm$ 4.4e-03 & 100\% [96, 100] \\
Joe ($\theta{=}3$) & \textbf{1.7e-01 $\pm$ 8.8e-03} & 3.8e-01 $\pm$ 6.1e-02 & 100\% [96, 100] \\
BB1 ($\theta{=}1.5,\,\delta{=}1.5$) & \textbf{1.8e-01 $\pm$ 1.7e-02} & 2.5e-01 $\pm$ 7.0e-02 & 85\% [77, 91] \\
BB7 ($\theta{=}2.5,\,\delta{=}1.5$) & 2.3e-01 $\pm$ 3.0e-02 & \textbf{1.9e-01 $\pm$ 5.9e-02} & 24\% [17, 33] \\
\midrule
Suite (8 cases, 800 paired reps) & \textbf{1.4e-01} & 2.1e-01 & 88\% [85, 90] \\
\bottomrule
\end{tabular}
\par\medskip
\begin{tabular}{lccc}
\toprule
\multicolumn{4}{l}{MI error $\downarrow$} \\
Case & VDC & pyvine-TLL & VDC win\% [95\% CI] \\
\midrule
Gaussian ($\rho{=}0.6$) & \textbf{0.011 $\pm$ 0.008} & 0.016 $\pm$ 0.011 & 71\% [61, 79] \\
Student-$t$ ($\rho{=}0.6,\,\nu{=}5$) & \textbf{0.018 $\pm$ 0.010} & 0.021 $\pm$ 0.013 & 69\% [59, 77] \\
Clayton ($\theta{=}3$) & \textbf{0.039 $\pm$ 0.018} & 0.117 $\pm$ 0.020 & 100\% [96, 100] \\
Gumbel ($\theta{=}2.5$) & \textbf{0.023 $\pm$ 0.014} & 0.057 $\pm$ 0.020 & 95\% [89, 98] \\
Frank ($\theta{=}5$) & \textbf{0.014 $\pm$ 0.010} & 0.019 $\pm$ 0.011 & 76\% [67, 83] \\
Joe ($\theta{=}3$) & \textbf{0.023 $\pm$ 0.014} & 0.086 $\pm$ 0.018 & 100\% [96, 100] \\
BB1 ($\theta{=}1.5,\,\delta{=}1.5$) & \textbf{0.028 $\pm$ 0.016} & 0.073 $\pm$ 0.020 & 100\% [96, 100] \\
BB7 ($\theta{=}2.5,\,\delta{=}1.5$) & \textbf{0.048 $\pm$ 0.018} & 0.069 $\pm$ 0.021 & 92\% [85, 96] \\
\midrule
Suite (8 cases, 800 paired reps) & \textbf{0.026} & 0.057 & 88\% [85, 90] \\
\bottomrule
\end{tabular}

\end{table}
}{}

\begin{table}[t]
    \centering
    \caption{\textbf{Bivariate benchmark dispersion.} Mean $\pm$ standard deviation across nine case means (each averaged over five sample seeds) for the same methods and metrics reported in Table~\ref{tab:bivariate}.}
    \label{tab:bivariate_dispersion_appendix}
    \small
    \IfFileExists{tables/tab_bivariate_dispersion.tex}{
        \makebox[\columnwidth][c]{\scalebox{0.92}{\begin{tabular}{lcccc}
\toprule
Method & ISE & KL (nats) & $|\Delta\tau|$ & $|\Delta\lambda_U|$ \\
\midrule
Histogram & 2.07e+00 $\pm$ 4.1e-02 & 1.321 $\pm$ 0.222 & 0.005 $\pm$ 0.002 & 0.025 $\pm$ 0.015 \\
KDE & 3.60e-02 $\pm$ 1.9e-02 & 0.016 $\pm$ 0.008 & 0.014 $\pm$ 0.002 & 0.026 $\pm$ 0.019 \\
pyvine-param & 3.41e-03 $\pm$ 2.9e-03 & 0.000 $\pm$ 0.000 & 0.006 $\pm$ 0.002 & 0.008 $\pm$ 0.008 \\
pyvine-TLL & 4.31e-02 $\pm$ 3.9e-02 & 0.013 $\pm$ 0.007 & 0.016 $\pm$ 0.007 & 0.025 $\pm$ 0.022 \\
VDC (one-shot) & 1.72e-02 $\pm$ 1.6e-02 & 0.006 $\pm$ 0.004 & 0.007 $\pm$ 0.003 & 0.010 $\pm$ 0.007 \\
\bottomrule
\end{tabular}
}}
    }{
        \fbox{\parbox{0.95\columnwidth}{\centering\textbf{[Bivariate dispersion table missing]}\\Run the table-generation script to produce the bivariate dispersion table.}}
    }
\end{table}

\subsection{High-Dimensional Scaling Benchmark}
\label{app:highdim_scaling}

We extend the synthetic Clayton-vine sweep to $d \in \{100, 200, 500\}$, where pyvine-TLL fitting becomes the dominant cost. The data generator is the same Clayton D-vine used in Appendix~\ref{app:flow_capacity_sensitivity}, with $\theta=2$ and $n_{\mathrm{train}}=20{,}000$, $n_{\mathrm{test}}=5{,}000$. This generator is first-order Markov, so its D-vine decomposition carries Clayton dependence at tree $1$ only, with exact independence at deeper trees --- appropriate for a density/fitting-cost benchmark, but not a probe of deep-tree recovery, which is tested separately with an all-trees construction in Appendix Table~\ref{tab:deep_tree_recovery}. We compare \ours{}, pyvine-TLL, RealNVP under the main-table training budget, and a Gaussian copula reference. \ours{} is the same frozen checkpoint as in the main results; RealNVP uses the same configuration as Table~\ref{tab:vine_nll}. Table~\ref{tab:highdim_scaling_appendix} gives the all-method NLL and timing sweep with the original \ours{} execution; Table~\ref{tab:fast_fit_timing} gives the fixed-thread timing comparison used in the main text, with an optimized execution that produces bitwise-identical fits (below). These results do not show that \ours{} dominates pyvine on NLL; rather, they show that \ours{} continues to fit substantially faster as $d$ grows, while keeping NLL well ahead of RealNVP and Gaussian copula. At $d{=}500$, one pyvine-TLL seed produced a non-finite held-out NLL ($-\infty$: the backend returned an infinite test log-likelihood; the numerical cause was not diagnosed); we exclude that seed from the NLL average, retain its fit time, and note that \ours{} returned finite held-out likelihoods on all seeds.

\begin{table}[t]
    \centering
    \caption{\textbf{High-dimensional Clayton-vine sweep (original execution).} Held-out NLL and fit time (mean $\pm$ population standard deviation over seeds $\{7,17,42\}$); optimized-execution timings appear in Table~\ref{tab:fast_fit_timing}. \ours{} uses one H200 GPU; CPU allocations and TLL thread counts are $8$ at $d{=}100,200$ and $16$ at $d{=}500$. Lower NLL is better. At $d{=}500$, pyvine-TLL NLL is averaged over the two finite runs because one seed produced a numerical $-\infty$ held-out NLL; fit time is averaged over all three runs.}
    \label{tab:highdim_scaling_appendix}
    \small
    \IfFileExists{tables/tab_highdim_scaling.tex}{
        \makebox[\columnwidth][c]{\scalebox{0.92}{\begin{tabular}{llcc}
\toprule
$d$ & Method & NLL (bits/dim) $\downarrow$ & Fit time (s) $\downarrow$ \\
\midrule
100 & VDC (Ours) & -0.6020 {\scriptsize $\pm$ 0.0034} & 44.1 {\scriptsize $\pm$ 0.1} \\
 & pyvine-TLL & -0.6102 {\scriptsize $\pm$ 0.0038} & 179.6 {\scriptsize $\pm$ 0.8} \\
 & Flow (RealNVP) & -0.4560 {\scriptsize $\pm$ 0.0018} & 10.6 {\scriptsize $\pm$ 0.8} \\
 & Gaussian copula & -0.4597 {\scriptsize $\pm$ 0.0022} & 0.1 {\scriptsize $\pm$ 0.0} \\
\midrule
200 & VDC (Ours) & -0.6020 {\scriptsize $\pm$ 0.0026} & 172.8 {\scriptsize $\pm$ 0.3} \\
 & pyvine-TLL & -0.6158 {\scriptsize $\pm$ 0.0026} & 693.1 {\scriptsize $\pm$ 1.4} \\
 & Flow (RealNVP) & -0.4159 {\scriptsize $\pm$ 0.0025} & 2.8 {\scriptsize $\pm$ 0.0} \\
 & Gaussian copula & -0.4615 {\scriptsize $\pm$ 0.0019} & 0.1 {\scriptsize $\pm$ 0.0} \\
\midrule
500 & VDC (Ours) & -0.5749 {\scriptsize $\pm$ 0.0006} & 1081.6 {\scriptsize $\pm$ 0.2} \\
 & pyvine-TLL & -0.6142 {\scriptsize $\pm$ 0.0007} & 2177.6 {\scriptsize $\pm$ 4.4} \\
 & Flow (RealNVP) & -0.3561 {\scriptsize $\pm$ 0.0027} & 16.8 {\scriptsize $\pm$ 4.4} \\
 & Gaussian copula & -0.4556 {\scriptsize $\pm$ 0.0010} & 0.4 {\scriptsize $\pm$ 0.0} \\
\bottomrule
\end{tabular}
}}
    }{
        \fbox{\parbox{0.95\columnwidth}{\centering\textbf{[High-dim scaling table missing]}\\Run the supplemental evaluation script to produce the high-dimensional scaling table.}}
    }
\end{table}

\paragraph{Optimized execution and fixed-thread timings.}
The same frozen checkpoint is fitted with cached, orientation-preserving Kendall-$\tau$ calculations, vectorized monotone h-function table construction, and batched grid transfers. These optimizations change execution only: the $64\times64$ grid, $15$ IPFP iterations, precision, and network edge-batch size of $256$ are unchanged. Edges within a tree are processed together in batches of up to $256$; conditional inputs require successive trees to be processed sequentially. Across all nine dimension/seed combinations, the optimized and original executions produce bitwise-identical fitted grids, h-tables, structures, $5000$ held-out likelihoods, and $128$ evaluated PIT and generated samples per fit. Table~\ref{tab:fast_fit_timing} shows a $34$--$35\%$ reduction in VDC fit time. Against $8$-thread TLL, the $9.6$-hour pretraining cost corresponds to approximately $201$, $52$, and $9$ fits at $d{=}100,200,500$; the main text uses the more conservative $16$-thread comparison.

\begin{table}[t]
    \centering
    \caption{\textbf{Fixed-thread deployment timings with identical VDC outputs.} Complete fit time in seconds (mean $\pm$ sample standard deviation over data seeds $\{7,17,42\}$) on the Clayton benchmark, $n_{\mathrm{train}}{=}20{,}000$. All $36$ fits run sequentially in separate workers on one AMD EPYC 9454 node, with method order randomized within each dimension/seed. VDC uses one H100 80\,GB GPU and $16$ CPU threads; TLL uses the stated thread count. Timers include order/structure selection and fitting, and exclude checkpoint loading, warmup, evaluation, and one-time pretraining. Both VDC paths fit the same greedy D-vine; TLL selects a regular vine. All three seeds contribute to every timing; TLL's seed-$7$ NLL is non-finite at $d{=}500$ for both thread counts. Ratios divide the mean times.}
    \label{tab:fast_fit_timing}
    \small
    \makebox[\columnwidth][c]{\scalebox{0.92}{\begin{tabular}{lccc}
\toprule
Method & $d=100$ & $d=200$ & $d=500$ \\
\midrule
VDC original execution & 45.83 $\pm$ 0.09 & 182.32 $\pm$ 0.16 & 1147.22 $\pm$ 0.50 \\
VDC optimized execution & 30.05 $\pm$ 0.01 & 118.68 $\pm$ 0.08 & 748.99 $\pm$ 0.47 \\
TLL (8 threads) & 202.19 $\pm$ 1.20 & 779.61 $\pm$ 2.59 & 4746.16 $\pm$ 8.92 \\
TLL (16 threads) & 118.96 $\pm$ 0.34 & 427.13 $\pm$ 0.46 & 2488.91 $\pm$ 4.71 \\
\midrule
TLL (8 threads) / optimized VDC & 6.73$\times$ & 6.57$\times$ & 6.34$\times$ \\
TLL (16 threads) / optimized VDC & 3.96$\times$ & 3.60$\times$ & 3.32$\times$ \\
\bottomrule
\end{tabular}
}}
\end{table}

\paragraph{Matched structures.} pyvine-TLL selects a Di{\ss}mann regular vine here, while \ours{} fits a greedy-order D-vine, which on this generator recovers the generating order. At $d{=}100$ (seed $42$; one node, $4$ CPU threads, the three fits run sequentially), fixing pyvine-TLL to that D-vine does not shorten its fit ($508$\,s versus $492$\,s with Di{\ss}mann selection, against $59$\,s for \ours{}) and leaves its held-out NLL unchanged ($-0.6052$ bits/dim in both cases, versus $-0.5975$ for \ours{}). The reported speedups and NLL gap are therefore not artifacts of structure selection.

\IfFileExists{tables/tab_pyvine_threads.tex}{%
\paragraph{Pretraining break-even.}
\label{par:break_even}
With $\Delta_t = t_{\text{pyvine-TLL}} - t_{\ours{}}$ the mean per-vine time saved in Table~\ref{tab:fast_fit_timing}, the $9.6$-hour single-GPU pretraining run is recovered after $N^{\star} = 9.6\,\text{h} / \Delta_t$ vines: approximately $389$, $112$, and $20$ at $d{=}100,200,500$ against $16$-thread TLL. These are elapsed-time break-even counts, not equal-hardware or monetary-cost estimates.

\paragraph{CPU thread-count sensitivity.}
The main timing curve fixes the TLL thread count across dimensions and shows $8$- and $16$-thread curves separately (Table~\ref{tab:fast_fit_timing}). Table~\ref{tab:pyvine_threads} gives a thread sweep on the same Clayton benchmark, seeds, and sample sizes with $1$, $8$, and $16$ TLL threads, alongside the original VDC execution. Its VDC allocations were $8$ CPU cores at $d{=}100,200$ and $16$ at $d{=}500$; read each fixed-thread TLL column separately. These ratios document single-core and multicore comparisons for the original execution, not the optimized one: the sweep gave $33$--$34\times$ speedups against one CPU thread and $2.5$--$4.6\times$ against $8$--$16$ threads at $d{=}100,200$, and against one thread the $9.6$-hour pretraining was recovered after roughly $24$ and $6$ fits.

\begin{table}[t]
    \centering
    \caption{\textbf{pyvine-TLL thread sweep.} VDC timings use the original execution. Same generator, seeds, and sample sizes as Table~\ref{tab:highdim_scaling_appendix}; $T$ denotes the number of CPU threads, and the last column lists the TLL-to-\ours{} time ratios for the three thread settings. Values are mean $\pm$ population standard deviation across seeds $\{7,17,42\}$; timed-out runs are identified in the table.}
    \label{tab:pyvine_threads}
    \small
    \makebox[\columnwidth][c]{\scalebox{0.92}{\begin{tabular}{lccccc}
\toprule
$d$ & VDC (s) & TLL $T{=}1$ (s) & TLL $T{=}8$ (s) & TLL $T{=}16$ (s) & VDC speedup ($T{=}1$ / $8$ / $16$) \\
\midrule
100 & 44.1 {\scriptsize $\pm$ 0.1} & 1465.5 {\scriptsize $\pm$ 0.2} & 203.3 {\scriptsize $\pm$ 0.6} & 119.6 {\scriptsize $\pm$ 0.3} & $33.2\times$ / $4.6\times$ / $2.7\times$ \\
200 & 172.8 {\scriptsize $\pm$ 0.3} & 5897.4 {\scriptsize $\pm$ 3.9} & 775.6 {\scriptsize $\pm$ 0.9} & 426.3 {\scriptsize $\pm$ 0.1} & $34.1\times$ / $4.5\times$ / $2.5\times$ \\
500 & 1081.6 {\scriptsize $\pm$ 0.2} & $>\!3.0\times10^{4}$\,$^{\dagger}$ & 4726.0 {\scriptsize $\pm$ 2.9}\,$^{\ddagger}$ & 3101.9 {\scriptsize $\pm$ 886.4} & $>\!28\times$ / $4.4\times$ / $2.9\times$ \\
\midrule
\multicolumn{6}{p{0.95\linewidth}}{\scriptsize $^{\dagger}$A single-threaded fit at $d{=}500$ had not completed after $8.5$ hours, when the sweep job reached its time limit (one run; lower bound). The $d{=}500$ $T{=}16$ dispersion reflects node contention across runs (range $2471$--$4355$\,s); pyvine times in this sweep were measured on different nodes from the main-table runs. $^{\ddagger}$Mean over two seeds; the third $d{=}500$, $T{=}8$ fit exceeded the $7200$\,s cap at $7634$\,s.} \\
\bottomrule
\end{tabular}
}}
\end{table}
}{}

\paragraph{Fixed-thread, same-node control.}
A separate control with the original VDC execution uses one H100 and a fixed $16$-core allocation on one AMD EPYC 9454 host, with \ours{} using the GPU and $16$ CPU threads and pyvine-TLL using either $8$ or $16$ threads throughout. Both methods fit the same greedy-order D-vine on the Clayton benchmark at $d\in\{20,50,100,200,500\}$. We time the common order selection separately and add it to each fixed-structure fit, excluding model loading and one-time pretraining. Over three timing repetitions of data seed $42$, the mean totals for \ours{} are $45.6$, $181.9$, and $1168.7$\,s at $d{=}100,200,500$, versus $151.0$, $552.9$, and $3316.4$\,s for $16$-thread TLL: $3.3\times$, $3.0\times$, and $2.8\times$ faster. Against $8$-thread TLL the corresponding speedups are $5.1\times$, $4.9\times$, and $4.7\times$. Absolute time saved therefore grows with dimension, while the speedup ratio declines modestly. These are timing repetitions of one dataset per dimension, not independent data seeds; the $d{=}500$ likelihood gap remains $0.0395$ bits/dim in TLL's favor. This control supplements the comparison with TLL's regular-vine fitting above.

\IfFileExists{tables/tab_cpu_timing.tex}{%
\paragraph{Matched-resource timing: \ours{} without a GPU.}
To separate amortization from GPU acceleration, Table~\ref{tab:cpu_timing} reports \ours{} fit time for the original execution using CPU inference only, with the same $8$-thread allocation as the pyvine-TLL baseline (held-out NLL equal to the GPU fits to reported precision). Across these cluster runs, CPU-only \ours{} takes $182$ vs.\ $203$\,s at $d{=}100$ and $730$ vs.\ $776$\,s at $d{=}200$; the GPU reduces the VDC times further. The VDC and TLL jobs used different cluster nodes, so this is a matched-thread-count comparison, not a same-node hardware benchmark. A single-job, single-node rerun of all three configurations found the CPU-only comparison node-dependent: on that node (seeds $7$, $17$ and $42$), CPU-only \ours{} took $304$/$1222$\,s versus $201$/$776$\,s for $8$-thread TLL, while GPU \ours{} remained $4.1\times$/$4.0\times$ faster ($49.7$/$195.2$\,s), and a $10$-seed $d{=}100$ repetition reproduced the $3$-seed NLL and timing values (VDC--TLL NLL difference $+0.0079 \pm 0.0007$ bits/dim; fit-time relative standard deviations below $1\%$). We therefore summarize CPU-only \ours{} as within about $1.6\times$ of $8$-thread TLL in either direction depending on the CPU, with the GPU supplying the consistent same-node speedup.

\begin{table}[t]
    \centering
    \caption{\textbf{Matched-resource timing.} \ours{} fit seconds on CPU only ($8$ threads, the same allocation as the TLL baseline) and on an RTX PRO 6000 (both over seeds $7$ and $17$), and on the H200 of the Table~\ref{tab:highdim_scaling_appendix} sweep (seeds $7$, $17$, $42$), against the pyvine-TLL thread sweep. These measurements use the original VDC execution. At $d{=}500$ only the H200 and TLL timings exist (TLL $T{=}8$ over two seeds; Table~\ref{tab:pyvine_threads}). In the TLL columns, $T$ denotes CPU thread count. CPU-only VDC and TLL were measured on different cluster nodes.}
    \label{tab:cpu_timing}
    \small
    \makebox[\columnwidth][c]{\resizebox{\columnwidth}{!}{\begin{tabular}{lccccc}
\toprule
$d$ & VDC CPU (8 thr.) & VDC RTX PRO 6000 Blackwell Server Edition & VDC H200 (ref) & TLL $T{=}8$ (ref) & TLL $T{=}16$ (ref) \\
\midrule
100 & 181.8 & 35.4 & 44.1 & 203.3 & 119.6 \\
200 & 729.8 & 145.6 & 172.8 & 775.6 & 426.3 \\
500 & -- & -- & 1081.6 & 4726.0 & 3101.9 \\
\bottomrule
\end{tabular}
}}
\end{table}
}{}

\subsection{Mixed-Family Vine Scaling (Gaussian + Clayton)}
\label{app:mixed_family_scaling}

The synthetic Clayton vine in Appendix~\ref{app:highdim_scaling} keeps every tree-1 edge in the same parametric family, so the candidate family set contains the data-generating pair-copulas. To test a heterogeneous family mix, we generate a D-vine in which odd tree-1 edges use a Gaussian copula with random per-edge $\rho \in [0.4, 0.8]$ (sign randomized) and even tree-1 edges use a Clayton copula with random per-edge $\theta \in [1.5, 4.0]$.\footnote{This scaling experiment uses $K{=}15$ IPFP iterations rather than the canonical $K{=}30$; Table~\ref{tab:ipfp_uci_sensitivity} shows that this difference changes held-out NLL and PIT-KS by at most $0.001$ on the UCI benchmarks ($d \le 43$); it was not checked separately at these dimensions.} Closed-form h-inverses guarantee uniform marginals at every depth. We then run the same method set as Appendix~\ref{app:highdim_scaling} at $d \in \{100, 200\}$ with $n=20{,}000$ training rows. Table~\ref{tab:mixed_family_scaling_appendix} reports held-out NLL and fit time with the original VDC execution (one seed, $42$; one H200 GPU and $16$ CPU threads, also used by TLL). The pattern is consistent with the uniform-Clayton setting: pyvine-param attains the best NLL on this mix ($-0.600$/$-0.604$ bits/dim), but its automatic family-selection step is $35$--$37\times$ slower than \ours{} ($1977$ vs.\ $53$\,s at $d{=}100$; $7349$ vs.\ $211$\,s at $d{=}200$). Among the fast methods, pyvine-TLL is slightly ahead on NLL, with \ours{} within $0.6$--$1.2 \times 10^{-2}$ bits/dim while running $1.8$--$2.0\times$ faster ($53$ vs.\ $106$\,s at $d{=}100$; $211$ vs.\ $376$\,s at $d{=}200$). Under the tested training budget, both vine methods achieve lower held-out NLL than RealNVP and the Gaussian copula by about $0.10$ bits/dim. Appendix~\ref{app:sp100_density} reports the S\&P100 comparison, where \ours{} also improves likelihood relative to TLL.

\begin{table}[t]
    \centering
    \caption{\textbf{Mixed-family (Gaussian + Clayton) D-vine scaling.} Tree-1 edges alternate between Gaussian (with random $\rho$) and Clayton (with random $\theta$). Held-out NLL and fit time at $n=20{,}000$.}
    \label{tab:mixed_family_scaling_appendix}
    \small
    \IfFileExists{tables/tab_heterogeneous_scaling.tex}{
        \makebox[\columnwidth][c]{\scalebox{0.92}{\begin{tabular}{llcc}
\toprule
$d$ & Method & NLL (bits/dim) $\downarrow$ & Fit (s) $\downarrow$ \\
\midrule
100 & VDC (Ours) & -0.5870 & 53.1 \\
 & pyvine-param & -0.5999 & 1976.5 \\
 & pyvine-TLL & -0.5931 & 105.9 \\
 & Flow (RealNVP) & -0.4803 & 11.0 \\
 & Gaussian copula & -0.4851 & 0.1 \\
\midrule
200 & VDC (Ours) & -0.5854 & 210.8 \\
 & pyvine-param & -0.6043 & 7349.4 \\
 & pyvine-TLL & -0.5973 & 375.8 \\
 & Flow (RealNVP) & -0.4679 & 3.0 \\
 & Gaussian copula & -0.4887 & 0.1 \\
\bottomrule
\end{tabular}
}}
    }{
        \fbox{\parbox{0.95\columnwidth}{\centering\textbf{[Mixed-family scaling table missing]}\\Run the supplemental evaluation script to produce the mixed-family scaling table.}}
    }
\end{table}

\subsection{S\&P100 Real-World Density Benchmark}
\label{app:sp100_density}

This appendix records the S\&P100 daily-returns density benchmark referenced in Section~\ref{subsec:highdim}. Setup: $d{=}100$ S\&P100 components, daily log-returns from 2019-05-13 through 2026-02-06 ($n{=}1695$ rows), random $80/20$ split with seed $42$ ($n_{\mathrm{train}}{=}1356$, $n_{\mathrm{test}}{=}339$). The raw close prices were collected from Stooq, using symbols from the public Wikipedia S\&P100 component list at collection time; raw prices are not redistributed. Marginals are fit via empirical PIT on the training split and applied to test rows; all methods receive the same PIT-transformed inputs. The pyvine-TLL fitting wall-clock budget is $3600$\,s; this is comfortably above the actual fit time at this $(d,n)$, so the comparison is not budget-limited. Table~\ref{tab:sp100_density_appendix} reports per-method NLL with $95\%$ confidence intervals from $n_{\mathrm{boot}}{=}2000$ bootstrap resamples of the test rows, and fit time. \ours{} is significantly the best by paired-bootstrap difference vs.\ each of pyvine-TLL, RealNVP, and Gaussian copula, but it is not the fastest method on this short-history real dataset.

\begin{table}[t]
    \centering
    \caption{\textbf{S\&P100 daily-returns density benchmark.} $d{=}100$, $n_{\mathrm{train}}{=}1356$, $n_{\mathrm{test}}{=}339$. NLL in bits/dim with $95\%$ bootstrap CI on test rows; lower is better.}
    \label{tab:sp100_density_appendix}
    \small
    \IfFileExists{tables/tab_sp100_density.tex}{
        \makebox[\columnwidth][c]{\scalebox{0.92}{\begin{tabular}{lcc}
\toprule
Method & NLL (bits/dim) [95\% CI] $\downarrow$ & Fit (s) $\downarrow$ \\
\midrule
VDC (Ours) & -0.5861 [-0.6655, -0.5225] & 12.4 \\
pyvine-TLL & -0.4594 [-0.5282, -0.4018] & 7.5 \\
Flow (RealNVP) & -0.2223 [-0.2901, -0.1642] & 17.8 \\
Gaussian copula & -0.5413 [-0.6153, -0.4758] & 0.0 \\
\bottomrule
\end{tabular}
}}
    }{
        \fbox{\parbox{0.95\columnwidth}{\centering\textbf{[S\&P100 density table missing]}\\Run the supplemental evaluation script to produce the S\&P100 density table.}}
    }
\end{table}

\IfFileExists{tables/tab_sp100_chrono.tex}{%
\paragraph{Chronological split.}
Daily returns exhibit volatility clustering, so a random split interleaves test days between training days. Table~\ref{tab:sp100_chrono} repeats the identical method set under a chronological split (train on the earliest $80\%$ of days, test on the latest $20\%$), the stricter forward-looking protocol, with the same paired-bootstrap difference tests. Under this protocol \ours{} remains the best method ($-0.418$ bits/dim versus $-0.293$ for pyvine-TLL and $-0.337$ for the Gaussian copula), with every paired-bootstrap difference favoring \ours{} at $p<0.001$. Because test days are serially dependent (the daily VDC--TLL gain has lag-one autocorrelation $0.20$), we also used moving-block bootstraps of the $339$ test days with block lengths of $5$, $10$, $20$, and $40$ days: the $95\%$ CI of the gain over TLL stays within $[0.101, 0.144]$ bits/dim, those over the Gaussian copula and RealNVP stay within $[0.055, 0.101]$ and $[0.295, 0.423]$, and \ours{} is better than TLL on $87\%$ of individual test days. These intervals condition on a single fitted model and do not include fitting uncertainty. TLL's poorer test likelihood than the Gaussian copula may reflect overfitting at $n_{\mathrm{train}}{=}1356$, $d{=}100$. VDC's advantage is consistent with regularization from pretraining, but this comparison does not isolate that effect.

\paragraph{Matched vine structures.} As elsewhere, \ours{} fits a greedy-order D-vine and pyvine-TLL a Di{\ss}mann regular vine. Matching the structures does not change the result: refitting pyvine-TLL on \ours{}'s D-vine order, the gain of \ours{} is $0.136$ bits/dim under the chronological split and $0.185$ under the random split (versus $0.126$ and $0.127$ as reported), and with both methods on Di{\ss}mann regular vines it is $0.135$ and $0.139$ (paired-bootstrap $95\%$ CIs within $[0.12, 0.21]$; \ours{} better on $88$--$91\%$ of test days). Refitting \ours{} with library commit b629041 on another device configuration reproduces its reported NLL to within $2\times10^{-4}$ bits/dim on both splits, although individual test days differ by up to $2.3$ nats; the released replay runner, on the original GPU configuration, reproduces the chronological run's NLL to all reported digits.

\begin{table}[t]
    \centering
    \caption{\textbf{S\&P100 chronological-split density benchmark.} Same data, methods, and bootstrap protocol as Table~\ref{tab:sp100_density_appendix}, with train/test split by date rather than at random.}
    \label{tab:sp100_chrono}
    \small
    \makebox[\columnwidth][c]{\scalebox{0.92}{\begin{tabular}{lcc}
\toprule
Method & NLL (bits/dim) [95\% CI] $\downarrow$ & Fit (s) $\downarrow$ \\
\midrule
VDC (Ours) & -0.4185 [-0.4732, -0.3709] & 10.6 \\
pyvine-TLL & -0.2928 [-0.3436, -0.2489] & 8.7 \\
Flow (RealNVP) & -0.0471 [-0.0984, -0.0018] & 6.8 \\
Gaussian copula & -0.3366 [-0.3909, -0.2884] & 0.0 \\
\bottomrule
\end{tabular}
}}
\end{table}
}{}

\IfFileExists{tables/tab_raw_composition.tex}{%
\paragraph{Raw-space composition versus a monolithic flow.}
The copula and flow families compose: fitting each marginal by maximum likelihood estimation (MLE) with a per-dimension Student-$t$ distribution on the training columns and the dependence with a copula module yields a raw-space density $f(x) = c\bigl(F_1(x_1),\ldots,F_d(x_d)\bigr)\prod_j f_j(x_j)$, evaluated on held-out rows. Table~\ref{tab:raw_composition} reports held-out raw-space NLL under the chronological split with the copula module varied and the marginals shared: the \ours{}-composed model attains $-4.285$ bits/dim versus $-4.154$ for the TLL composition, $-4.199$ for the Gaussian-copula composition, and $-3.676$ for a monolithic RealNVP trained on standardized raw returns with the identical architecture and training budget as the copula-space flow used throughout the paper (25 epochs at batch size 2048, at most $25$ optimizer steps on the $1{,}220$ training rows left after a validation split, so this is a small-budget flow rather than a tuned one); every paired-bootstrap difference favors the \ours{} composition at $p<0.001$, with a gain over the monolithic flow of $0.609$ bits/dim (paired bootstrap over test days, 95\% CI $[0.559, 0.666]$). At $n_{\mathrm{train}}{=}1356$ the monolithic flow trails even the dependence-free $t$-marginal model ($-3.864$), while the composition isolates dependence in copula space, where the frozen synthetic prior regularizes; the copula-space NLLs under $t$-CDF PIT ($-0.421$/$-0.290$/$-0.335$/$-0.043$) closely reproduce the empirical-PIT results in Table~\ref{tab:sp100_chrono}. We present this as evidence in the small-sample, high-dimensional regime---consistent with the capacity study in Appendix~\ref{app:flow_capacity_sensitivity}---rather than a universal claim about flow models.

\begin{table}[t]
    \centering
    \caption{\textbf{Raw-space S\&P100 density with composed marginals and dependence} (chronological split; $d{=}100$, $n_{\mathrm{train}}{=}1356$). All composed models share the same per-dimension Student-$t$ marginals; lower NLL is better. Brackets are $95\%$ bootstrap CIs over held-out test rows for one fitted model. The monolithic RealNVP is trained directly on standardized raw returns with the same architecture and budget as the copula-space flow.}
    \label{tab:raw_composition}
    \small
    \makebox[\columnwidth][c]{\scalebox{0.92}{\begin{tabular}{lcc}
\toprule
Model & Raw NLL (bits/dim) [95\% CI] $\downarrow$ & Copula NLL (bits/dim) \\
\midrule
\ours{} copula + $t$ marginals & \textbf{-4.285 [-4.328, -4.243]} & -0.421 \\
pyvine-TLL copula + $t$ marginals & -4.154 [-4.200, -4.104] & -0.290 \\
Gaussian copula + $t$ marginals & -4.199 [-4.253, -4.143] & -0.335 \\
RealNVP copula + $t$ marginals & -3.907 [-3.978, -3.829] & -0.043 \\
Independence + $t$ marginals & -3.864 [-3.933, -3.788] & --- \\
RealNVP on raw returns (monolithic) & -3.676 [-3.747, -3.596] & --- \\
Multivariate Gaussian (raw) & -3.989 [-4.066, -3.908] & --- \\
\bottomrule
\end{tabular}
}}
\end{table}
}{}

\subsection{Fixed-Checkpoint Sample-Size Sensitivity}
\label{app:mi_sample_size}

To isolate inference-time sample-size effects from retraining, we keep one frozen \ours{} checkpoint and vary $n$ at fixed dimension $d=50$. We use two block-MI benchmarks. The Gaussian AR(1) case is a useful sanity check in which the Gaussian baseline is correctly specified (and most accurate for $n\geq 30{,}000$); even there, \ours{} bias shrinks steadily with sample size, from $+0.661$ at $n{=}1000$ to $-0.019$ at $n{=}100{,}000$. The more informative non-Gaussian case is a Clayton-chain benchmark with true block MI $0.636$ nats (the exact MI of the chain's Clayton($\theta{=}3$) pair copula). There, \ours{} absolute MI error drops from $0.665$ at $n{=}1000$ to $0.083$ at $n{=}10{,}000$, $0.018$ at $n{=}30{,}000$, and $0.017$ at $n{=}100{,}000$, corresponding to relative errors of $13.1\%$, $2.8\%$, and $2.7\%$. The neural MI baselines added for comparison are faster but materially less accurate in this regime: InfoNCE is at $18.0\%$, $19.7\%$, and $19.4\%$ relative error at $n{=}10{,}000$, $30{,}000$, and $100{,}000$, while MINE is near $29\%$ at these sizes; KSG remains above $56\%$, and the Gaussian baseline stays around $18$--$23\%$. Runtime still crosses over against KSG: at $n{=}10{,}000$, $30{,}000$, and $100{,}000$, \ours{} takes $15.3$, $41.6$, and $139.1$ s, versus $20.9$, $130.3$, and $1359.2$ s for KSG, while MINE and InfoNCE take about $1.0$ s and $3.7$ s. Figure~\ref{fig:mi_sample_size_clayton_appendix} gives the full sensitivity study---relative MI error in panel (a) and runtime in panel (b); the dimension sweep is in Figure~\ref{fig:dimension_appendix}(a). We present this as a sensitivity study of a fixed pretrained estimator, not as a universal small-$n$ advantage claim.

\begin{figure}[!htbp]
    \centering
    \includegraphics[width=\textwidth]{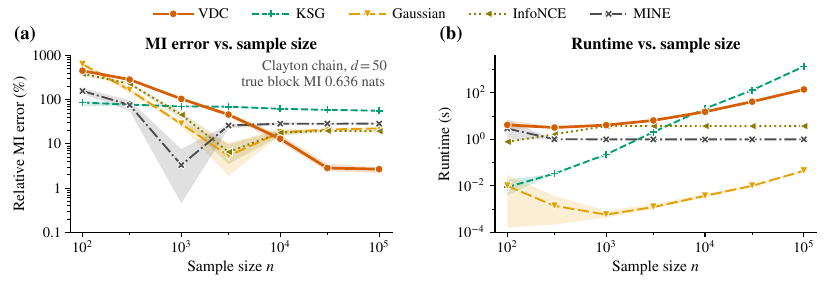}
    \caption{\textbf{Fixed-checkpoint MI sensitivity.} Non-Gaussian $d=50$ Clayton-chain benchmark (true block MI $0.636$ nats) with one frozen VDC checkpoint. (a) Relative MI error versus inference sample size. (b) Runtime versus inference sample size. VDC becomes the most accurate method by $n=10^4$; MINE and InfoNCE remain faster but less accurate, while KSG scales much worse at large $n$. ``Gaussian'' denotes the closed-form Gaussian-copula MI baseline. Lines show the mean and shaded bands the range (minimum to maximum) over $3$ independent trials per sample size; the wide runtime bands at $n=10^2$ come from a slower first trial of each method.}
    \label{fig:mi_sample_size_clayton_appendix}
\end{figure}

\subsection{Additional Edge and Information Diagnostics}

We include additional diagnostics that complement the main-text evaluation and sharpen the comparison to neural copula baselines. Table~\ref{tab:bivariate_complex_appendix} extends the bivariate benchmark to harder qualitative families. Table~\ref{tab:neural_copula_appendix} isolates the comparison to ACNet on matched pair-copula tasks. Tables~\ref{tab:theory_mi_methods_appendix} and~\ref{tab:tc_scaling_appendix} collect supplementary information-estimation diagnostics that support the main text.

\begin{table}[t]
    \centering
    \caption{\textbf{Hard synthetic bivariate benchmark.} Results on the complex synthetic patterns of the training zoo (X, ring, double-banana; samples drawn from each pattern's grid density), showing density and conditional-transform accuracy. These cases were used for checkpoint selection, so this table reports model-selection diagnostics rather than an independent test. ISE is integrated over the unit square. Edge-ISE is integrated squared error restricted to boundary cells (either coordinate below $0.1$ or above $0.9$); $h$-MAE is the mean absolute error averaged over both directional h-function grids.}
    \label{tab:bivariate_complex_appendix}
    \small
    \IfFileExists{tables/tab_bivariate_complex.tex}{
        \makebox[\columnwidth][c]{\scalebox{0.92}{\begin{tabular}{lcccc}
\toprule
Method & ISE $\downarrow$ & Edge-ISE $\downarrow$ & $h$-MAE $\downarrow$ & Time (ms) $\downarrow$ \\
\midrule
Histogram & 1.891e+00 & 4.908e-01 & 0.043 & \textbf{2.2} \\
KDE & 1.565e+00 & 2.936e-01 & 0.066 & 81.0 \\
pyvine-param & 2.715e+00 & 1.062e+00 & 0.149 & 434.4 \\
pyvine-TLL & 9.568e-01 & 1.676e-01 & 0.043 & 14.9 \\
VDC (one-shot) & \textbf{4.067e-03} & \textbf{1.836e-03} & \textbf{0.004} & 6.1 \\
\bottomrule
\end{tabular}
}}
    }{
        \fbox{\parbox{0.95\columnwidth}{\centering\textbf{[Hard bivariate table missing]}}}
    }
\end{table}

\IfFileExists{tables/tab_ood_families.tex}{%
\paragraph{Out-of-distribution copula families.}
Table~\ref{tab:bivariate} uses fresh samples at the nine fixed parameter settings in Table~\ref{tab:benchmark_spec}, from family classes represented in pretraining. Table~\ref{tab:ood_families} evaluates the identical frozen checkpoint on families entirely absent from the training zoo (BB6, BB8, Tawn, and Plackett), against per-case pyvine-TLL refits, to separate amortization quality from prior match. Averaged over the nine cases, the frozen estimator has lower ISE ($0.039$ vs.\ $0.054$; lower in $7/9$ cases), lower Kendall-$\tau$ error ($0.012$ vs.\ $0.015$), less than half the MI error ($0.017$ vs.\ $0.040$ nats; lower in $7/9$), and the same mean held-out NLL ($-0.385$; lower for the frozen estimator in $6/9$ cases, one of them by less than $10^{-4}$, and for TLL in $3/9$). TLL is better on every metric for BB8 ($\theta{=}3.0,\,\delta{=}0.8$), on Kendall-$\tau$ and MI error for BB8 ($\theta{=}5.0,\,\delta{=}0.6$), and on ISE and NLL for Tawn; its NLL is also lower by $0.002$ on BB6 ($\theta{=}2.0,\,\delta{=}1.5$).

\begin{table}[t]
    \centering
    \caption{\textbf{Out-of-distribution bivariate families.} Frozen \ours{} checkpoint versus per-case pyvine-TLL refits on copula families absent from the training zoo (three seeds per case, $n{=}2000$; pyvinecopulib samples, closed-form sampler for Plackett). ISE: cell-averaged densities against the exact cell averages of the true copula (TLL on its own cell averages); $|\Delta\tau|$: exact Kendall's $\tau$ of the fitted grid; MI error against a Monte Carlo estimate of the true MI; NLL: held-out, \ours{} through its bilinear edge density and TLL through its own pdf.}
    \label{tab:ood_families}
    \small
    \makebox[\columnwidth][c]{\scalebox{0.92}{\begin{tabular}{llcccc}
\toprule
Case & Method & ISE $\downarrow$ & $|\Delta\tau|$ $\downarrow$ & MI err $\downarrow$ & NLL $\downarrow$ \\
\midrule
BB6 ($\theta{=}2.0,\,\delta{=}1.5$) & VDC (Ours) & \textbf{0.0357} & 0.009 & \textbf{0.010} & -0.517 \\
 & pyvine-TLL & 0.0587 & 0.009 & 0.046 & \textbf{-0.519} \\
BB6 ($\theta{=}3.0,\,\delta{=}2.0$) & VDC (Ours) & \textbf{0.1253} & \textbf{0.007} & \textbf{0.020} & \textbf{-1.045} \\
 & pyvine-TLL & 0.2395 & 0.018 & 0.116 & -1.042 \\
BB8 ($\theta{=}3.0,\,\delta{=}0.8$) & VDC (Ours) & 0.0194 & 0.018 & 0.016 & -0.137 \\
 & pyvine-TLL & \textbf{0.0115} & \textbf{0.009} & \textbf{0.008} & \textbf{-0.138} \\
BB8 ($\theta{=}5.0,\,\delta{=}0.6$) & VDC (Ours) & \textbf{0.0099} & 0.015 & 0.016 & -0.160 \\
 & pyvine-TLL & 0.0108 & \textbf{0.011} & \textbf{0.009} & -0.160 \\
Tawn ($\theta{=}2.5,\,\psi_1{=}0.7,\,\psi_2{=}1.0$) & VDC (Ours) & 0.0790 & 0.014 & \textbf{0.020} & -0.312 \\
 & pyvine-TLL & \textbf{0.0507} & 0.014 & 0.047 & \textbf{-0.335} \\
Plackett ($\theta{=}5.0$) & VDC (Ours) & \textbf{0.0024} & \textbf{0.008} & \textbf{0.008} & \textbf{-0.136} \\
 & pyvine-TLL & 0.0106 & 0.019 & 0.014 & -0.131 \\
Plackett ($\theta{=}20.0$) & VDC (Ours) & \textbf{0.0303} & \textbf{0.021} & \textbf{0.038} & \textbf{-0.477} \\
 & pyvine-TLL & 0.0320 & 0.026 & 0.053 & -0.465 \\
BB6$^{90}$ ($\theta{=}2.0,\,\delta{=}1.5$) & VDC (Ours) & \textbf{0.0327} & \textbf{0.008} & \textbf{0.012} & \textbf{-0.528} \\
 & pyvine-TLL & 0.0617 & 0.015 & 0.051 & -0.522 \\
BB8$^{90}$ ($\theta{=}3.0,\,\delta{=}0.8$) & VDC (Ours) & \textbf{0.0130} & \textbf{0.012} & \textbf{0.011} & \textbf{-0.157} \\
 & pyvine-TLL & 0.0140 & 0.015 & 0.013 & -0.155 \\
\midrule
Mean & VDC (Ours) & \textbf{0.0386} & \textbf{0.012} & \textbf{0.017} & -0.385 \\
 & pyvine-TLL & 0.0544 & 0.015 & 0.040 & -0.385 \\
\bottomrule
\end{tabular}
}}
\end{table}

}{}

\paragraph{Grid-resolution floor.}
\label{app:grid_floor}
A grid estimator cannot be more accurate than the grid allows. Table~\ref{tab:grid_floor} gives, for $31$ bivariate cases ($17$ from pretraining families: independence, Gaussian, Student-$t$, Clayton, Gumbel, Frank and BB7; $6$ Joe and BB1 cases, reported separately; $5$ from the unseen families BB6, BB8 and Tawn; $3$ exact two- and three-component mixtures; some cases rotated, including rotations absent from pretraining), the held-out log-likelihood loss $\EE_c[\log c-\log c^{\star}]$ of the \emph{exact} cell-averaged truth $c^{\star}$ evaluated as a piecewise-constant density, estimated on $50{,}000$ fresh samples per case (truth from the pyvinecopulib CDF). Exact cell averages minimize this loss among piecewise-constant densities on a given grid, so these values are floors for piecewise-constant estimators: about $0.007$ nats per edge for pretraining families at the deployed $64\times64$ uniform grid, growing quickly for near-deterministic pairs. The deployed model instead evaluates its grids by bilinear interpolation. Interpolating the exact cell averages gives $0.0055$ nats, but cell averages are not the optimal bilinear coefficients, so this is a reference value rather than a lower bound. At $n{=}20{,}000$ the frozen checkpoint reaches $0.0063$ nats on the pretraining families, while pyvine-TLL, which is not tied to a grid, reaches $0.0040$; this suggests that the grid representation limits the large-sample regime but does not isolate the cause. Cells spaced uniformly in probit coordinates are narrowest near the edges of the unit square, where Archimedean and Student-$t$ copulas place their tail mass, and lower the piecewise-constant floor about fourfold at the same cell count; they help much less for mass concentrated along the diagonal. Whether a trained estimator realizes these gains is not tested here. In the ablation of Appendix~\ref{app:probit}, which used an auxiliary denoising diffusion implicit model (DDIM) recipe on uniform grids, the trained $128\times128$ models (evaluated after fewer training steps) were less accurate than the $64\times64$ ones in both density parameterizations, so a finer representation alone need not improve a trained estimator.

\begin{table}[t]
    \centering
    \caption{\textbf{Representation floor of the copula grid.} Held-out KL (nats per edge) of the exact cell-averaged true copula, as a piecewise-constant density, for uniform and probit-spaced grids (cell edges equally spaced in probit coordinates on $[-3.5,3.5]$ and mapped through $\Phi$, so cells are narrowest near the boundaries). Left: means over case groups ($17$ pretraining-family, $6$ Joe/BB1, $5$ unseen-family and $3$ mixture cases; $50{,}000$ held-out samples per case). Right: Gaussian copulas of increasing correlation ($2{,}000{,}000$ samples each; Monte Carlo standard errors below $0.0003$).}
    \label{tab:grid_floor}
    \small
    \begin{tabular}{lcccc@{\hspace{10pt}}ccc}
    \toprule
    & \multicolumn{4}{c}{case group} & \multicolumn{3}{c}{Gaussian $\rho$} \\
    Grid & pretraining & Joe/BB1 & unseen & mixtures & $0.5$ & $0.9$ & $0.99$ \\
    \midrule
    uniform $64$ (deployed) & 0.0072 & 0.0138 & 0.0066 & 0.0079 & 0.0013 & 0.0112 & 0.0556 \\
    uniform $128$ & 0.0035 & 0.0069 & 0.0033 & 0.0038 & 0.0006 & 0.0049 & 0.0245 \\
    probit-spaced $64$ & 0.0017 & 0.0034 & 0.0016 & 0.0017 & 0.0004 & 0.0043 & 0.0471 \\
    probit-spaced $128$ & 0.0005 & 0.0010 & 0.0004 & 0.0004 & 0.0001 & 0.0012 & 0.0126 \\
    \bottomrule
    \end{tabular}
\end{table}

\paragraph{Focused neural pair-copula comparison.}\label{app:real_edge_scaling}
To isolate the distinction between our amortized edge estimator and prior neural copula models in a controlled synthetic setting, we also ran a task-matched bivariate comparison against ACNet \citep{ling2020deep} on three Archimedean families that overlap cleanly with our evaluation protocol: Clayton, Frank, and Gumbel. ACNet is trained from scratch per pair on CPU on $n{=}2000$ samples and evaluated on the same $64 \times 64$ grid, with metrics averaged over three seeds (each also redraws the training sample); the \ours{} metrics are single deterministic fits of the deployed checkpoint on the model-selection evaluation suite, on samples drawn separately from ACNet's, so the comparison is not on matched samples. \ours{} attains lower ISE, lower $h$-function error, and lower MI error on all three cases, while reducing per-edge fit time by roughly $2.9\times 10^5\times$ on average (\ours{} inference on one GPU). Table~\ref{tab:neural_copula_appendix} gives the case-by-case breakdown.

Tables~\ref{tab:theory_mi_methods_appendix} and~\ref{tab:tc_scaling_appendix} summarize the synthetic information benchmarks from the appendix protocol. Table~\ref{tab:theory_mi_methods_appendix} focuses on pairwise MI on Gaussian AR(1), and Table~\ref{tab:tc_scaling_appendix} reports exact total-correlation scaling with dimension.

\begin{table}[t]
    \centering
    \caption{\textbf{Focused neural pair-copula comparison.} Matched Archimedean edge tasks comparing the deployed VDC checkpoint used throughout the paper against ACNet trained from scratch per pair. ISE is integrated squared error, $h$-MAE is mean absolute error over the h-function grid, and MI err is absolute mutual-information error. ACNet rows average three training seeds and \ours{} rows are single deterministic fits. ISE follows the model-selection protocol (projected cell-center truth), so its scale is not comparable with Tables~\ref{tab:bivariate} and~\ref{tab:bivariate_replicates}.}
    \label{tab:neural_copula_appendix}
    \small
    \IfFileExists{tables/tab_neural_copula.tex}{
        \makebox[\columnwidth][c]{\scalebox{0.92}{\begin{tabular}{llcccc}
\toprule
Case & Method & ISE $\downarrow$ & $h$-MAE $\downarrow$ & MI err $\downarrow$ & Speedup \\
\midrule
Clayton($\theta$=3.0) & VDC & \textbf{4.100e-04} & \textbf{0.001} & \textbf{0.004} & 331.6k$\times$ \\
 & ACNet & 2.450e+00 & 0.028 & 0.221 &  \\
\midrule
Frank($\theta$=5.0) & VDC & \textbf{4.683e-04} & \textbf{0.002} & \textbf{0.001} & 284.2k$\times$ \\
 & ACNet & 7.189e-03 & 0.007 & 0.007 &  \\
\midrule
Gumbel($\theta$=2.5) & VDC & \textbf{8.294e-04} & \textbf{0.001} & \textbf{0.004} & 258.2k$\times$ \\
 & ACNet & 1.120e+00 & 0.022 & 0.173 &  \\
\midrule
\textbf{Mean} & VDC & \textbf{5.692e-04} & \textbf{0.002} & \textbf{0.003} & \textbf{291.3k$\times$} \\
 & ACNet & 1.192e+00 & 0.019 & 0.134 &  \\
\bottomrule
\end{tabular}
}}
    }{
        \fbox{\parbox{0.95\columnwidth}{\centering\textbf{[Neural-copula table missing]}}}
    }
\end{table}

\begin{table}[t]
    \centering
    \caption{\textbf{MI method comparison on Gaussian AR(1).} Absolute MI error and runtime under the synthetic protocol. The table reports mean $\pm$ standard deviation across pair/trial/dimension records: MI errors use the $12$ Gaussian AR(1) records ($\rho=0.6$, $n=2000$, three trials at each of four dimensions), while times pool $48$ records from all four synthetic scenarios.}
    \label{tab:theory_mi_methods_appendix}
    \small
    \IfFileExists{tables/tab_theory_mi_methods.tex}{
        \makebox[\columnwidth][c]{\scalebox{0.92}{\begin{tabular}{lcc}
\toprule
Method & $|\Delta \mathrm{MI}|$ on Gaussian AR(1) $\downarrow$ & Time (s, mean$\pm$std) $\downarrow$ \\
\midrule
Gaussian & \textbf{0.010 $\pm$ 0.009} & \textbf{0.00 $\pm$ 0.00} \\
InfoNCE & 0.012 $\pm$ 0.012 & 0.73 $\pm$ 0.76 \\
KSG & 0.013 $\pm$ 0.008 & 0.02 $\pm$ 0.00 \\
VDC (Ours) & 0.013 $\pm$ 0.011 & 0.01 $\pm$ 0.00 \\
MINE & 0.021 $\pm$ 0.014 & 0.33 $\pm$ 0.01 \\
NWJ & 0.022 $\pm$ 0.015 & 0.36 $\pm$ 0.01 \\
\bottomrule
\end{tabular}}}
    }{
        \fbox{\parbox{0.95\columnwidth}{\centering\textbf{[Theory MI-method table missing]}}}
    }
\end{table}

\begin{table}[t]
    \centering
    \caption{\textbf{TC estimation by dimension.} Synthetic Gaussian-copula benchmark with analytic ground truth (the AR(1) process of Figure~\ref{fig:info_results_composite}(b) and Appendix Figure~\ref{fig:dimension_appendix}(b); \ours{} fits a D-vine truncated after its first tree, which is exact for this process); TC (KSG) is the full chain-rule KSG estimator in $d$ dimensions (a structurally matched KSG control is in Appendix~\ref{par:tc_tree1_control}); entries are mean $\pm$ standard deviation over seeds $\{7,17,42\}$, and $|\mathrm{Err}_{\mathrm{VDC}}|$ is the absolute difference between the VDC and true TC values.}
    \label{tab:tc_scaling_appendix}
    \small
    \IfFileExists{tables/tab_tc_scaling.tex}{
        \makebox[\columnwidth][c]{\scalebox{0.92}{\begin{tabular}{lcccc}
\toprule
$d$ & TC (True) & TC (VDC) & TC (KSG) & $|$Err$_\text{VDC}|$ \\
\midrule
5 & 1.347 & 1.320 {\scriptsize $\pm$ 0.021} & 1.360 {\scriptsize $\pm$ 0.007} & 0.027 \\
10 & 3.030 & 3.025 {\scriptsize $\pm$ 0.025} & 2.820 {\scriptsize $\pm$ 0.014} & 0.005 \\
20 & 6.397 & 6.302 {\scriptsize $\pm$ 0.054} & 4.985 {\scriptsize $\pm$ 0.018} & 0.095 \\
50 & 16.497 & 16.375 {\scriptsize $\pm$ 0.148} & 8.643 {\scriptsize $\pm$ 0.020} & 0.122 \\
\bottomrule
\end{tabular}
}}
    }{
        \fbox{\parbox{0.95\columnwidth}{\centering\textbf{[TC scaling table missing]}}}
    }
\end{table}

\paragraph{Absolute-contribution shares.}
\label{par:absolute_tc_shares}
For each edge, let $a_e$ be its mean log-density contribution on the held-out sample. Figure~\ref{fig:tc_decomposition}(b) reports $s_\ell=\sum_{e\in T_\ell}|a_e|/\sum_e|a_e|$ for tree $\ell$; panel (c) sums these shares over the removed tree group. The denominator is the sum of absolute edge contributions, not the absolute value of their sum, $|\widehat{\mathrm{TC}}|=|\sum_e a_e|$. These differ when fitted held-out contributions are negative.

\paragraph{Structural control for the TC benchmark.}
\label{par:tc_tree1_control}
On the AR(1) benchmark of Figure~\ref{fig:info_results_composite}(b) and Table~\ref{tab:tc_scaling_appendix}, \ours{} uses the first-order Markov structure: its D-vine is truncated after the first tree, so its TC is a sum of $d{-}1$ adjacent-pair estimates. \ours{} is fitted on $75\%$ of the rows and scored by its mean held-out log-density on the other $25\%$, whereas KSG and the neural estimators use all rows; every chain-rule term is clipped at zero. As a structural control we applied the same decomposition to KSG ($k{=}5$), summing bivariate KSG estimates over the adjacent pairs. On the three seed-$42$ trials at each dimension of the Figure~\ref{fig:info_results_composite}(b) sweep it gives a TC of $16.63$ nats at $d{=}50$, against $16.24$ for \ours{} and the true $16.50$, and both estimators are within $0.09$ nats of the truth at $d{=}5$, $10$ and $20$. A paired replication on $50$ further datasets at $d{=}50$ ($n{=}5000$, fixed seeds and a protocol fixed in advance, the deployed checkpoint) gives mean absolute errors of $0.187$ nats for \ours{} and $0.201$ nats for KSG ($95\%$ bootstrap intervals $[0.149, 0.227]$ and $[0.171, 0.231]$), with biases of opposite sign ($-0.149$ and $+0.195$ nats); the paired difference in absolute error, $-0.013$ nats ($[-0.070, 0.045]$), does not distinguish the two estimators. The roughly twofold underestimate of the full chain-rule KSG estimator is therefore a property of that estimator in $d$ dimensions and is absent under the Markov decomposition. Timed sequentially on the same node, \ours{} (one H100 GPU and one CPU thread, including structure selection and held-out evaluation) took a median $0.76$\,s per dataset against $3.58$\,s for KSG on one CPU thread of the same AMD EPYC 9454 host, a paired ratio of $4.7\times$; this is a deployment comparison, not equal hardware.

\IfFileExists{tables/tab_deep_tree_recovery.tex}{%
\paragraph{Deep-tree recovery under known deep dependence.}
Near-zero deep-edge NLL on real data is observationally consistent with two explanations: genuinely weak deep dependence, or estimators defaulting toward independence as recursion error accumulates. To separate them we constructed a simplified D-vine ($d{=}30$, $29$ trees) with Clayton($\theta{=}2$) pair-copulas at \emph{every} tree level --- true per-edge dependence $0.432$ nats at all depths (verified by quadrature, Monte Carlo, and grid integration); note this is an adversarial stress test (conditional $\tau{=}0.5$ even at tree $29$), and that the scaling generator of Appendix~\ref{app:highdim_scaling} is first-order Markov (dependence at tree $1$ only), so it cannot probe this question. Both \ours{} and pyvine-TLL fit this D-vine in its generating order, TLL edge by edge. Table~\ref{tab:deep_tree_recovery} reports recovered per-edge dependence by depth at $n_{\mathrm{train}}{=}20{,}000$: \ours{} recovers $98$--$99\%$ of the true per-edge dependence at trees $1$--$2$, $93\%$ at tree $3$, $79\%$ at tree $4$, $55\%$ at tree $5$, and $23\%$ at tree $6$; pyvine-TLL attenuates at a similar rate, retaining slightly more dependence than \ours{} at trees $1$, $3$ and $4$ and less at tree $2$ and from tree $5$ on. Deep-tree attenuation is therefore shared by both simplified-vine estimators at this sample size; the experiment does not isolate its cause, and depth-related claims should be read with this boundary in mind.

\begin{table}[t]
    \centering
    \caption{\textbf{Deep-tree recovery stress test.} Simplified D-vine with Clayton($\theta{=}2$) at all $29$ trees ($d{=}30$, $n_{\mathrm{train}}{=}20{,}000$, mean over seeds $\{7,17,42\}$): true versus recovered per-edge dependence by depth bucket. The true edge MI column is the oracle's held-out mean log-density per edge (exact value $0.432$ nats).}
    \label{tab:deep_tree_recovery}
    \small
    \makebox[\columnwidth][c]{\scalebox{0.92}{\begin{tabular}{lcccc}
\toprule
Tree depth & True edge MI (nats) & VDC & pyvine-TLL & VDC/true \\
\midrule
1 & 0.437 & 0.429 & 0.432 & 0.98 \\
2--5 & 0.431 & 0.352 & 0.347 & 0.82 \\
6--10 & 0.432 & 0.030 & 0.000 & 0.07 \\
11--20 & 0.429 & 0.002 & 0.000 & 0.00 \\
21--29 & 0.425 & 0.000 & 0.000 & 0.00 \\
\bottomrule
\end{tabular}
}}
\end{table}
}{}

\IfFileExists{tables/tab_sampling_quality.tex}{%
\subsection{Unconditional Sampling Quality}
\label{app:sampling_quality}

We quantify how well data simulated from fitted vines (via inverse h-functions, Algorithm~\ref{alg:vine_sampling}) reproduces held-out dependence. For each dataset we draw $n_{\mathrm{sim}}{=}5000$ samples from the fitted model and compare pairwise Kendall-$\tau$ matrices (a fixed subset of $200$ pairs for S\&P100) and empirical tail concentrations at level $q{=}0.05$ against the held-out test set, with a bootstrap of the training data as an empirical sampling-variability reference. For each pair $(U_i,U_j)$, these are $\lambda_L(q)=\PP(U_i\leq q,U_j\leq q)/q$ and $\lambda_U(q)=\PP(U_i>1-q,U_j>1-q)/q$. Table~\ref{tab:sampling_quality} reports the results. On S\&P100, \ours{}-simulated data reproduces the held-out Kendall-$\tau$ matrix close to the bootstrap reference (root-mean-square error $0.052$ versus $0.039$), whereas samples from the per-dataset pyvine-TLL refit are far off ($0.192$); tail concentrations are comparable for the two vines. On the Combined Cycle Power Plant (CCPP) data, \ours{} again has the lower $\tau$ error ($0.020$ vs.\ $0.036$, reference $0.011$) but larger lower- and upper-tail errors. With validated inverse-h interpolation and compiled CPU recursion, simulation takes $0.72$\,s including table packing for $5{,}000$ S\&P100 draws at $d{=}100$, versus $0.92$\,s for TLL, both on eight CPU threads (about $1.3\times$ faster). This compares our D-vine with TLL's regular vine. Kernel compilation costs a separate one-time $0.37$\,s. A production replay on another host of the same CPU model took $0.98$\,s versus $0.95$\,s for TLL, indicating approximate parity there rather than a portable speed advantage. The compiled and reference NumPy samplers produce identical coordinates on all six dataset--seed fits; their fitted densities, likelihoods and sampling-quality scores are unchanged. The release automatically falls back to NumPy if compilation or validation fails or the packed-table memory limit is exceeded. Packing adds $309$\,MiB at $d{=}100$; the default limit is $512$\,MiB. Imputation and VaR use the equivalent NumPy implementation with interpolation through the final grid interval. \begin{table}[t]
    \centering
    \caption{\textbf{Sampling quality} ($n_{\mathrm{sim}}{=}5000$; mean $\pm$ standard deviation over seeds $\{7,17,42\}$). Pairwise Kendall-$\tau$ matrix RMSE and lower-/upper-tail concentration errors ($\lambda_L$/$\lambda_U$ MAE) of simulated versus held-out data at $q{=}0.05$, and simulation wall time; the data-reference row is a bootstrap of real training data and measures empirical sampling variability rather than a strict lower bound. Simulation runs sequentially on an AMD EPYC 9454 host (one H100 allocation), with eight CPU threads for both methods. Times average the per-seed medians of three repetitions, include random-number generation and VDC table packing (rounded upward), and exclude the separately reported one-time kernel compilation. VDC uses a D-vine and TLL a regular vine. All rows use the deployed checkpoint.}
    \label{tab:sampling_quality}
    \small
    \makebox[\columnwidth][c]{\resizebox{\columnwidth}{!}{\begin{tabular}{llcccc}
\toprule
Dataset & Method & $\tau$-RMSE $\downarrow$ & $\lambda_L$ MAE $\downarrow$ & $\lambda_U$ MAE $\downarrow$ & Sim.\ time (s) \\
\midrule
CCPP (power plant) & VDC (Ours) & 0.0197 $\pm$ 0.0053 & 0.0530 $\pm$ 0.0034 & 0.0265 $\pm$ 0.0073 & 0.0021 \\
 & pyvine-TLL & 0.0364 $\pm$ 0.0026 & 0.0274 $\pm$ 0.0028 & 0.0196 $\pm$ 0.0020 & 0.054 \\
 & Data reference (train bootstrap) & 0.0113 $\pm$ 0.0020 & 0.0208 $\pm$ 0.0021 & 0.0075 $\pm$ 0.0021 & -- \\
\midrule
S\&P100 & VDC (Ours) & 0.0520 $\pm$ 0.0060 & 0.1520 $\pm$ 0.0622 & 0.1023 $\pm$ 0.0265 & 0.72 \\
 & pyvine-TLL & 0.1916 $\pm$ 0.0107 & 0.2047 $\pm$ 0.1131 & 0.1519 $\pm$ 0.0656 & 0.92 \\
 & Data reference (train bootstrap) & 0.0385 $\pm$ 0.0023 & 0.1413 $\pm$ 0.0297 & 0.1018 $\pm$ 0.0082 & -- \\
\bottomrule
\end{tabular}
}}
\end{table}
}{}

\subsection{VaR Backtesting}
\label{app:var}

We report rolling-window VaR backtesting on S\&P100 daily returns ($d=100$) using empirical marginals and copula dependence. All methods share the same window ($252$ days), refit cadence (every $5$ days), and Monte Carlo budget ($n_{\text{sim}}=5000$); we include historical simulation and Gaussian baselines for context. Figure~\ref{fig:var_calibration} summarizes the outcome: \ours{} and the Gaussian (normal portfolio-return) baseline overshoot the nominal rate at the $1\%$ level (\ours{} observed $1.9\%$, $27/1442$ violations, Kupiec $p=0.003$; Gaussian $1.9\%$), while at the $5\%$ level \ours{}'s unconditional coverage is acceptable ($5.4\%$ observed, Kupiec $p=0.48$) but violations cluster in time (Christoffersen $p=0.003$). The excess violations at $1\%$ and their clustering at $5\%$ identify the calibration issues discussed in Sections~\ref{sec:experiments} and~\ref{sec:conclusion}.

\begin{figure}[t]
    \centering
    \includegraphics[width=4.4in]{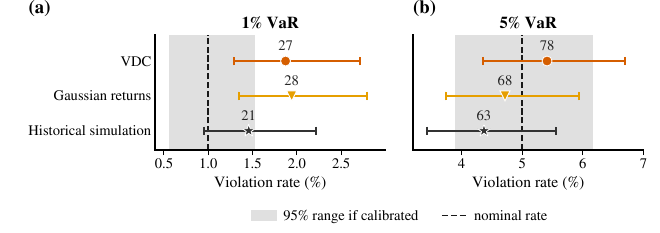}
    \caption{\textbf{VaR calibration on S\&P100 daily returns.} Observed violation rate of each method at the $1\%$ (a) and $5\%$ (b) levels under the rolling-window backtest protocol ($n{=}1442$ backtest days per method and level; numbers above the points are violation counts). Horizontal bars show $95\%$ Wilson score confidence intervals for the observed rates; the dashed line is the nominal rate and the shaded band the central $95\%$ range of the observed rate under correct calibration, $\mathrm{Binomial}(1442,\alpha)/1442$. The points outside the band (\ours{} and Gaussian at $1\%$) are exactly those rejected by the Kupiec test at the $5\%$ level.}
    \label{fig:var_calibration}
\end{figure}

\subsection{Rolling Dependence Dashboard}
\label{app:rolling_dashboard}

The same rolling S\&P100 setup can also be used descriptively rather than predictively. For each refit window, we evaluate the fitted vine on the in-window pseudo-observations and decompose the mean log-copula contribution into tree-level bands and dominant tree-1 edges. Figure~\ref{fig:rolling_dependence_dashboard} shows that window-level dependence varies substantially over time: the mean log-copula density falls from $86$ to $48$ nats between windows ending in late 2020 and early 2025, with an intermediate rise to $77$ nats in early 2023, and then partially rebounds to $55$--$59$ nats (a); the first three trees account for most dependence throughout ($51$--$69\%$; b); and a small set of same-industry pairs of different companies repeatedly ranks among the largest tree-1 contributors (c). The returns matrix stores no per-row dates; window end dates are assigned by mapping its $1{,}695$ rows to the consecutive NYSE trading days between the first and last dates recorded in its metadata, whose count matches exactly. This descriptive dashboard illustrates structured monitoring rather than prediction.

\begin{figure}[!htbp]
    \centering
    \includegraphics[width=\textwidth]{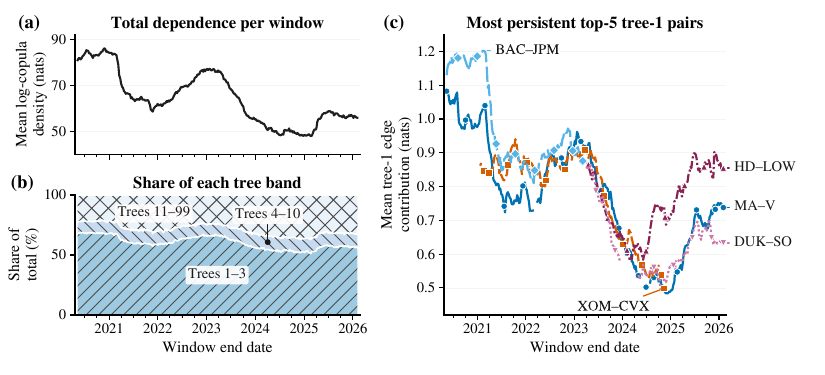}
    \caption{\textbf{Rolling dependence dashboard on S\&P100 returns.} Repeated VDC refits ($289$ rolling windows of $252$ trading days, refit every $5$ days), each evaluated on its own window and plotted at the window's last trading day. (a) Mean log-copula density per window (nats). (b) Share of it captured by each tree band; the first three trees account for the majority of dependence throughout the sample. (c) Mean tree-1 log-copula contribution (nats) for the five cross-name pairs that most persistently rank among the top-5 tree-1 edges (the same-issuer GOOG--GOOGL pair is excluded). Only the top-5 edges are recorded per refit window, so each series is drawn only where the pair ranks in the top five; gaps mark windows where it fell outside.}
    \label{fig:rolling_dependence_dashboard}
\end{figure}

\subsection{Missing Data Imputation}
\label{app:imputation}

\IfFileExists{tables/tab_imputation_exact.tex}{%
\paragraph{Exact conditional imputation.}
We evaluate missing-data imputation under 20\% MCAR (missing completely at random) masking on the Combined Cycle Power Plant (CCPP) data ($d{=}5$). The fitted vine supports conditional sampling through inverse h-functions: per missing pattern, we refit the D-vine with observed variables ordered first ($25$ patterns for the $500$ evaluation rows; fast amortized fitting makes per-pattern refits affordable) and sample the missing coordinates through the inverse Rosenblatt transform (the sequential inverse conditional-CDF map of Algorithm~\ref{alg:vine_sampling}) with the observed prefix held fixed. Missing entries are imputed by the mean of conditional samples in copula space mapped back through the inverse empirical CDFs, which is not exactly the conditional mean in data units. Under this procedure, \ours{} has about $20\%$ higher RMSE than $k$-NN ($6.895$ vs.\ $5.743$). The difference may reflect both conditional-model error and the point estimator: averaging in copula space before applying a nonlinear inverse marginal differs from averaging in data units, the conditional-mean estimate appropriate for squared error. \begin{table}[t]
    \centering
    \caption{\textbf{Inverse-Rosenblatt imputation} ($20\%$ MCAR). RMSE on CCPP for sequential inverse-Rosenblatt conditional sampling under the fitted numerical h-functions (ordered D-vine refitted per missing pattern) versus the approximate importance-weighted procedure of Table~\ref{tab:imputation} and the best simple baseline. Each $\Delta$ column reports the percentage RMSE reduction relative to its named comparator, so positive values favor the exact procedure.}
    \label{tab:imputation_exact}
    \footnotesize
    \begin{tabular}{lccccc}
\toprule
Dataset & VDC exact & VDC approx. & Best baseline & $\Delta$ vs.\ approx. & $\Delta$ vs.\ best \\
\midrule
CCPP & 6.895 & 6.896 & 5.743 ($k$-NN) & 0.0\% & -20.1\% \\
\bottomrule
\end{tabular}

\end{table}
}{}

\paragraph{Approximate conditional procedure.}
For comparison, an approximate conditional procedure draws an unconditional candidate pool from the fitted vine. Writing $U_{\mathrm{obs}}$ and $U_{\mathrm{miss}}$ for the observed and missing copula-coordinate subvectors, respectively, it used kernel importance weights in the observed subspace to approximate $p(U_{\text{miss}} \mid U_{\text{obs}})$, imputed missing entries by a weighted conditional mean in copula space, and mapped back through inverse empirical CDFs. On CCPP the approximate and exact procedures give nearly identical RMSE ($6.896$ vs.\ $6.895$; Table~\ref{tab:imputation_exact}) and both trail $k$-NN. Per-pattern refitting changes the vine, so this comparison does not isolate the sampling approximation alone.

\begin{table}[t]
    \centering
    \caption{\textbf{Imputation RMSE} (20\% MCAR). Lower is better; Improvement is the percentage RMSE reduction of VDC relative to the listed best baseline, so positive values favor VDC.}
    \label{tab:imputation}
    \small
    \IfFileExists{tables/tab_imputation.tex}{
        \makebox[\columnwidth][c]{\scalebox{0.92}{\begin{tabular}{lccccc}
\toprule
Dataset & $d$ & VDC & Best baseline & Baseline & Improvement \\
\midrule
CCPP & 5 & 6.896 & 5.743 & $k$-NN & -20.1\% \\
\bottomrule
\end{tabular}
}}
    }{
        \fbox{\parbox{0.95\columnwidth}{\centering\textbf{[Imputation table missing]}\\Run the table-generation script to produce the imputation table.}}
    }
\end{table}

\section{Algorithms}
\label{app:algos}

This section provides pseudocode for fitting a simplified vine with the frozen edge estimator and for the two operations the fitted vine supports: likelihood evaluation and sampling. Implementations cache, for each edge $e$, the projected density grid $\hat D_e$ and both directional h-functions $h_e$. In Algorithm~\ref{alg:vine_ll}, $\hat c_e(p,q)$ is the edge density at $(p,q)\in[0,1]^2$, evaluated by bilinear interpolation of $\hat D_e$ between cell centers (this interpolant integrates to one), and $\mathrm{LL}$ denotes the log-likelihood accumulator. Algorithm~\ref{alg:vine_sampling} shows the D-vine ordering for notational simplicity; for variables $j>\ell$, write $D_{j,\ell}=\{\ell+1,\ldots,j-1\}$ for the intervening conditioning set (empty when $\ell=j-1$). The released implementation provides inverse-Rosenblatt sampling and conditioning for D-vines and C-vines; for a general regular vine the same recursion applies along the vine's natural order but is not yet implemented.

\begin{algorithm}[H]
    \caption{\textbf{Vine fitting with the frozen edge estimator}}
    \label{alg:vine_fit}
    \begin{algorithmic}[1]
    \State \textbf{Input:} pseudo-observations $u^{(1)},\ldots,u^{(n)} \in [0,1]^d$, frozen $f_\theta$, grid size $m$, IPFP iterations $K$
    \State $U^{(i)}_{j|\emptyset} \gets u^{(i)}_j$ for all $i,j$
    \For{$\ell = 1$ to $d-1$}
        \State Select $T_\ell$: maximum spanning tree on $|\hat\tau|$ of the current pseudo-observations under the proximity condition (Di{\ss}mann), or the prescribed D-vine order
        \For{each edge $e = (j, k \mid D) \in T_\ell$ (batched on the GPU for D- and C-vines)}
            \State $H_e \gets$ $m\times m$ histogram of $\{(U^{(i)}_{j \mid D}, U^{(i)}_{k \mid D})\}_{i=1}^n$
            \State $\hat D_e \gets \Pi^{(K)}_{\mathrm{IPFP}}\bigl(\exp\{\operatorname{clip}(f_\theta(H_e,\mathrm{coords},\log n),-20,20)\}\bigr)$ \Comment{corruption level zero}
            \State cache $\hat D_e$ and its h-functions $h_{j \mid k;D}$, $h_{k \mid j;D}$
            \State $U^{(i)}_{j \mid D \cup \{k\}} \gets h_{j \mid k;D}(U^{(i)}_{j \mid D} \mid U^{(i)}_{k \mid D})$, \ $U^{(i)}_{k \mid D \cup \{j\}} \gets h_{k \mid j;D}(U^{(i)}_{k \mid D} \mid U^{(i)}_{j \mid D})$
        \EndFor
    \EndFor
    \State \textbf{Return:} structure $\{T_\ell\}$ and cached $\{\hat{D}_e, h_e\}$
    \end{algorithmic}
\end{algorithm}

\begin{algorithm}[H]
    \caption{\textbf{Vine copula log-likelihood evaluation}}
    \label{alg:vine_ll}
    \begin{algorithmic}[1]
    \State \textbf{Input:} $u \in [0,1]^d$, vine with cached $\{\hat{D}_e, h_e\}$
    \State $U_{j|\emptyset} \gets u_j$, $\mathrm{LL} \gets 0$
    \For{$\ell = 1$ to $d-1$}
        \For{each edge $e = (j, k \mid D) \in T_\ell$}
            \State $\mathrm{LL} \mathrel{+}= \log \hat{c}_e(U_{j \mid D}, U_{k \mid D})$
            \State $U_{j \mid D \cup \{k\}} \gets h_{j \mid k;D}(U_{j \mid D} \mid U_{k \mid D})$
            \State $U_{k \mid D \cup \{j\}} \gets h_{k \mid j;D}(U_{k \mid D} \mid U_{j \mid D})$
        \EndFor
    \EndFor
    \State \textbf{Return:} copula log-likelihood $\mathrm{LL}$
    \end{algorithmic}
\end{algorithm}

If separate marginal densities are fitted for a raw-space likelihood, their contribution $\sum_j \log f_j(x_j)$ can be added outside the copula model. The experiments in this paper emphasize copula-space likelihoods and empirical marginal transforms.

\begin{algorithm}[H]
    \caption{\textbf{Vine sampling via inverse h-functions}}
    \label{alg:vine_sampling}
    \begin{algorithmic}[1]
    \State \textbf{Input:} D-vine with cached h-functions $\{h_e\}$ and inverse h-functions $\{h_e^{-1}\}$
    \State Draw independent $W_1, \ldots, W_d \sim \mathrm{Unif}(0,1)$; set $U_1 \gets W_1$
    \For{$j = 2$ to $d$}
        \State $V \gets W_j$ \Comment{$V$ is the working conditional probability}
        \For{$\ell = 1$ to $j-1$}
            \State $D \gets \{\ell+1,\ldots,j-1\}$ \Comment{deepest edge first}
            \State $V \gets h^{-1}_{j \mid \ell;D}(V \mid U_{\ell \mid D})$ \Comment{$U_{\ell \mid D}$ from forward h-functions}
        \EndFor
        \State $U_j \gets V$
    \EndFor
    \State \textbf{Return:} $(U_1, \ldots, U_d)$
    \end{algorithmic}
\end{algorithm}

\section{Ablation Studies}
\label{app:ablations}

\subsection{Probit Transformation}
\label{app:probit}

A design choice in copula density estimation is whether the network targets the copula density itself or the joint density of the probit-transformed coordinates, $f_Z(z_u,z_v)=c(u,v)\,\phi(z_u)\,\phi(z_v)$ with $z=\Phi^{-1}(u)$, evaluated at the same uniform cell centers and converted back to a copula density before projection. The grid itself stays uniform in copula coordinates in both variants, so this ablation tests the density parameterization, not probit-spaced cells. Table~\ref{tab:probit} compares both approaches across grid sizes.

This auxiliary experiment uses DDIM-style checkpoints rather than the deployed one-shot uniform-mix checkpoint, so its absolute errors are not comparable to Table~\ref{tab:bivariate}. It compares probit and non-probit parameterizations at each grid size. Comparisons across grid sizes also change the checkpoint step, and the suite includes the defective Joe case identified in the caption.

\begin{table}[t]
    \centering
    \caption{\textbf{Probit vs.\ non-probit transformation.} ISE, MI error and Kendall-$\tau$ error for different grid sizes. At $m=64$, non-probit performs better; at $m=128$, probit performs better. Auxiliary DDIM checkpoints evaluated on an auxiliary nine-case suite that includes a Joe case drawn with a defective sampler; the $m{=}128$ models were evaluated at step $150{,}000$ and the $m{=}64$ models at step $200{,}000$. ISE is the unit-square integral, evaluated consistently across grid sizes.}
    \label{tab:probit}
    \small
    \IfFileExists{tables/tab_probit_comparison.tex}{
        \makebox[\columnwidth][c]{\scalebox{0.92}{\begin{tabular}{lcccc}
\toprule
Configuration & Grid $m$ & ISE $\downarrow$ & MI Error $\downarrow$ & $|\Delta\tau|$ $\downarrow$ \\
\midrule
DDIM (bilinear, no probit) & 64 & \textbf{4.096e-02} & \textbf{0.094} & \textbf{0.074} \\
DDIM (probit) & 64 & 1.193e-01 & 0.141 & 0.099 \\
\midrule
DDIM (bilinear, no probit) & 128 & 2.342e+00 & 0.778 & 0.268 \\
DDIM (probit) & 128 & \textbf{8.020e-01} & \textbf{0.329} & \textbf{0.146} \\
\bottomrule
\end{tabular}
}}
    }{
        \fbox{\parbox{0.95\columnwidth}{\centering\textbf{[Probit table missing]}\\Run the table-generation script to produce the probit comparison table.}}
    }
\end{table}

At $m=64$ the non-probit grid is better on all three metrics; at $m=128$ the probit grid is. The ablation does not isolate the cause of this reversal.

\IfFileExists{tables/tab_noipfp_ablation.tex}{%
\subsection{No-Projection Ablation}
\label{app:noipfp_ablation}

To measure the contribution of the copula projection, Table~\ref{tab:noipfp_ablation} runs the identical frozen checkpoint with the projection disabled (mass renormalization only, so grid marginals are not uniform and the inserted edge object is not a valid pair-copula) against the deployed $K{=}30$ setting, on the synthetic Clayton vine ($n_{\mathrm{train}}{=}20{,}000$, three seeds) and the MAF POWER and GAS copula-space benchmarks ($n_{\mathrm{train}}{=}5{,}000$). This isolates the contribution of enforcing copula constraints from that of amortized density regression. With the copula projection disabled during fitting, the fitted object is no longer a normalized copula density---its nominal score is not a likelihood, and on the $d{=}50$ Clayton benchmark it reports $-52.1$ bits/dim against $-0.60$ with projection---and on POWER and GAS the negative log score rises by $0.51$ and $1.17$ bits/dim, respectively. These unnormalized score differences are diagnostic, not measures of predictive accuracy; the experiment demonstrates the importance of enforcing copula constraints.

\begin{table}[t]
    \centering
    \caption{\textbf{No-projection ablation.} Negative log scores in bits/dim with IPFP disabled ($K{=}0$; mass renormalization only) versus deployed ($K{=}30$); mean over seeds where applicable. The $K{=}0$ joint score is unnormalized and is not a valid NLL. Its values and differences ($K{=}0$ minus $K{=}30$) are diagnostic only.}
    \label{tab:noipfp_ablation}
    \small
    \makebox[\columnwidth][c]{\scalebox{0.92}{\begin{tabular}{lccc}
\toprule
Benchmark & \shortstack{Unnormalized score\\$K{=}0$} & \shortstack{Reported NLL\\$K{=}30$} & Score difference \\
\midrule
Clayton vine ($d{=}50$) & -52.0894 & -0.6033 & -51.4862 \\
GAS (MAF, $d{=}8$) & -0.3076 & -1.4773 & +1.1697 \\
POWER (MAF, $d{=}6$) & 0.3671 & -0.1413 & +0.5084 \\
\bottomrule
\end{tabular}
}}
\end{table}
}{}

\subsection{Number of IPFP Iterations}
\label{app:ipfp_iterations}

We report marginal-density residuals: $\epsilon_{\mathrm{row}}=\max_a|\Delta\sum_bD_{ab}-1|$ and the analogous column residual. These equal $m$ times the row/column mass residuals; the pre-projection diagnostic in Appendix~\ref{app:implementation} uses the mass convention. At $K{=}0$ the raw output is normalized to total mass one before computing these residuals. A marginal density can still exceed one, so the residual $8.86$ is compatible with unit total mass.

To justify the deployed choice of $K{=}30$ iterations, Table~\ref{tab:ipfp_iter_ablation} reports marginal error and runtime for a sweep over $K$ on raw network outputs produced by the deployed checkpoint on four representative copula families (Gaussian, Clayton, Frank, Gumbel). Without any projection ($K{=}0$) the maximum marginal deviation from uniform is about $9$; a single IPFP pass already brings the column marginal error to $\sim\!10^{-6}$ and the row marginal error to $\sim\!0.13$; at $K{=}15$ the maximum row/column marginal error is $\sim\!10^{-3}$ at a median cost of $0.66$\,ms; at the deployed $K{=}30$ setting the error is roughly $2\times10^{-4}$ at $1.23$\,ms; and, for these four synthetic edges, $K{=}100$ pushes the error to the single-precision floor (about $8\times10^{-7}$) while still taking a median $3.88$\,ms. We therefore use $K{=}30$ as the main accuracy/latency trade-off, while Table~\ref{tab:ipfp_iter_ablation} records both the faster $K{=}15$ option and the stricter $K{=}100$ option for applications with different marginal-tolerance requirements.

We also tested whether finite IPFP tolerance materially changes full-vine UCI likelihood or PIT calibration. Table~\ref{tab:ipfp_uci_sensitivity} reruns the VDC UCI benchmark while varying only the number of IPFP iterations. Moving from $K{=}15$ through $K{=}30$ to $K{=}100$ tightens the worst marginal error where needed, but NLL and the probability-integral-transform Kolmogorov--Smirnov statistic (PIT-KS; Appendix~\ref{app:uci_density}) change by at most $0.001$ bits/dim and $0.0015$, respectively. This suggests that the remaining calibration behavior is not primarily caused by the finite projection tolerance, and supports $K{=}30$ as a practical choice for accuracy and fitting time.

\begin{table}[!htbp]
    \centering
    \caption{\textbf{UCI sensitivity to IPFP iterations.} VDC full-vine held-out NLL, PIT-KS, and worst fitted-edge marginal-density residual (defined above) as the number of IPFP iterations changes, on the MAF splits with $n_{\mathrm{train}}{=}5{,}000$ (the setting of Table~\ref{tab:vine_nll}). The table shows $K{=}15$, the deployed setting $K{=}30$, and a stricter setting $K{=}100$; $K{=}50$ is in the saved run record.}
    \label{tab:ipfp_uci_sensitivity}
    \small
    \IfFileExists{tables/tab_ipfp_uci_sensitivity.tex}{
        \makebox[\columnwidth][c]{\scalebox{0.92}{\begin{tabular}{lrrrr}
\toprule
Dataset & IPFP $K$ & NLL (bits/dim) $\downarrow$ & PIT-KS $\downarrow$ & Max marg.\ err. $\downarrow$ \\
\midrule
POWER ($d{=}6$) & 15 & -0.1413 & 0.017 & 4.5e-07 \\
POWER ($d{=}6$) & 30 & -0.1413 & 0.017 & 1.6e-07 \\
POWER ($d{=}6$) & 100 & -0.1413 & 0.017 & 1.6e-07 \\
\midrule
GAS ($d{=}8$) & 15 & -1.4765 & 0.049 & 1.2e-02 \\
GAS ($d{=}8$) & 30 & -1.4773 & 0.049 & 8.4e-03 \\
GAS ($d{=}8$) & 100 & -1.4775 & 0.048 & 2.7e-03 \\
\midrule
HEPMASS ($d{=}21$) & 15 & -0.2372 & 0.021 & 6.3e-04 \\
HEPMASS ($d{=}21$) & 30 & -0.2372 & 0.021 & 3.9e-05 \\
HEPMASS ($d{=}21$) & 100 & -0.2372 & 0.021 & 2.0e-07 \\
\midrule
MINIBOONE ($d{=}43$) & 15 & -0.9684 & 0.028 & 1.1e-02 \\
MINIBOONE ($d{=}43$) & 30 & -0.9689 & 0.027 & 2.5e-03 \\
MINIBOONE ($d{=}43$) & 100 & -0.9691 & 0.027 & 2.1e-05 \\
\bottomrule
\end{tabular}
}}
    }{
        \fbox{\parbox{0.95\columnwidth}{\centering\textbf{[UCI IPFP sensitivity table missing]}}}
    }
\end{table}

\begin{table}[!htbp]
    \centering
    \caption{\textbf{IPFP iteration ablation.} Maximum row/column marginal-density residual ($\max_a|\Delta\sum_bD_{ab}-1|$ and its column analogue) and wall-clock time as a function of IPFP iteration count $K$ on the deployed checkpoint. The $\star$ marks the canonical $K{=}30$ setting. Errors are the maximum over a batch of raw outputs for four representative copula families (Gaussian, Clayton, Frank, Gumbel) and over all grid rows or columns. Times are medians of $100$ repetitions of the whole-batch projection on one H100 GPU, after $10$ warmups per $K$, with GPU synchronization immediately before and after each measurement and randomized $K$ order. Network inference and initialization are excluded.}
    \label{tab:ipfp_iter_ablation}
    \small
    \IfFileExists{tables/tab_ipfp_iter_ablation.tex}{
        \makebox[\columnwidth][c]{\scalebox{0.92}{\begin{tabular}{rccc}
\toprule
Iterations $K$ & Max row err. $\downarrow$ & Max col err. $\downarrow$ & Time (ms) \\
\midrule
0    & $8.86$        & $7.76$        & $0.03$ \\
1    & $1.26\!\times\!10^{-1}$ & $3.7\!\times\!10^{-6}$ & $0.12$ \\
2    & $5.65\!\times\!10^{-2}$ & $8.3\!\times\!10^{-7}$ & $0.16$ \\
5    & $1.56\!\times\!10^{-2}$ & $8.3\!\times\!10^{-7}$ & $0.28$ \\
10   & $4.26\!\times\!10^{-3}$ & $8.3\!\times\!10^{-7}$ & $0.47$ \\
15   & $1.53\!\times\!10^{-3}$ & $7.7\!\times\!10^{-7}$ & $0.66$ \\
20   & $6.78\!\times\!10^{-4}$ & $7.7\!\times\!10^{-7}$ & $0.85$ \\
\textbf{30}$^\star$ & $\mathbf{1.79\!\times\!10^{-4}}$ & $\mathbf{8.3\!\times\!10^{-7}}$ & $\mathbf{1.23}$ \\
50   & $1.49\!\times\!10^{-5}$ & $7.2\!\times\!10^{-7}$ & $1.98$ \\
100  & $7.15\!\times\!10^{-7}$ & $7.7\!\times\!10^{-7}$ & $3.88$ \\
\bottomrule
\end{tabular}
}}
    }{
        \fbox{\parbox{0.95\columnwidth}{\centering\textbf{[IPFP iter-ablation table missing]}\\Run the supplemental evaluation script to produce the IPFP iteration-ablation table.}}
    }
\end{table}

\clearpage
\section{Exploratory Retraining on Corrected Data}
\label{app:retrained}

This section evaluates two subsequently trained estimators; results outside this section use the original frozen checkpoint, except the comparisons of other trained checkpoints in Appendices~\ref{app:corruption_ablation} and~\ref{app:probit}. These experiments test whether corrected and more diverse training data improve amortized estimation, and do not replace the main paper's accuracy or speed claims. Both estimators use the same U-Net architecture, initialized from a separate $60{,}000$-step pilot trained from scratch on corrected data, then fine-tuned for $8{,}000$ steps with fresh continuous-mixture samples and independent samples for the likelihood loss. The pilot is not descended from the original checkpoint. One pilot initialization and one fine-tuning seed are represented in the tables; two replicates were also evaluated. A second fine-tuning seed gives similar Table~\ref{tab:bivariate} results (KL reductions of $11.8\%$ and $49.0\%$, improvements on $38/45$ and $41/45$ fits), and its chronological S\&P100 differences from the original model are $-0.0008$ (VDC-128) and $-0.0020$ (VDC-64) bits/dim, with intervals including zero. An independently initialized pilot fine-tuned with the same recipes gives $11.7\%$ and $48.9\%$ ($36/45$ and $41/45$ fits) and again improves VDC-128 in seven of eight UCI settings, but its VDC-128 is worse than the original model on chronological S\&P100 by $0.0053$ bits/dim ($40$-day moving-block $95\%$ interval $[0.0021, 0.0087]$; VDC-64: $+0.0025$, $[-0.0006, 0.0051]$).

The fine-tuning distribution uses eight families (Gaussian, Student-$t$, Clayton, Gumbel, Frank, Joe, BB1, BB7), positive and negative dependence or rotations, and five bins of component $|\tau|$ spanning $0.005$--$0.85$. Mixtures have probability $0.70$, with $2,3,4,6$, or $8$ components and Dirichlet concentration $0.1,0.3,1$, or $3$; independence has probability $0.05$. Sample sizes are drawn from $\{50,100,200,500,1000,2000,5000,10000,20000,50000\}$. VDC-64 uses raw latent uniforms for half the inputs and rank pseudo-observations for the remainder; VDC-128 uses latent uniforms with probability $0.75$. Both use uniform cells; their probit coordinate features do not imply probit-spaced cells. Each continuous bilinear density is projected with its integrated basis weights, and its conditional CDFs and inverses are computed consistently with that density. This representation differs from the approximate density/h-function interpolation of the main model. The checkpoints were fixed before the evaluations below; no further selection used these results.

\paragraph{Bivariate likelihood on the paper's cases.}
Table~\ref{tab:retrained_bivariate} uses exactly the nine cases of Table~\ref{tab:benchmark_spec} and its five fresh evaluation seeds, $n_{\mathrm{train}}=2000$, and $20{,}000$ independent test samples per fit. Inputs are raw copula samples, as in Table~\ref{tab:bivariate}. The original estimator and baseline scores come from the Table~\ref{tab:bivariate} record; the oracle test likelihood is reproduced for every case and seed. Relative to the original model, mean held-out KL decreases by $11.5\%$ for VDC-64 and $49.1\%$ for VDC-128, with improvements on $37/45$ and $41/45$ fits. Parametric selection remains most accurate on these in-library families. This is an in-distribution assessment, not evidence of generalization to previously unexamined families: model development also used families from the main paper's OOD suite. We compare native likelihoods rather than treating coefficients on different grid sizes as directly comparable cell-average ISEs.

\begin{table}[!htbp]
\centering
\caption{\textbf{Retrained estimators on the Table~\ref{tab:bivariate} evaluation samples.} Native held-out KL in nats per edge, mean $\pm$ standard deviation over $45$ fits (nine fixed cases, five fresh sample seeds per case). The spread combines case heterogeneity and sampling variation; it is not a confidence interval or variation over network-training seeds.}
\label{tab:retrained_bivariate}
\small
\begin{tabular}{lcc}
\toprule
Estimator & Held-out KL (nats) & Lower KL than original \\
\midrule
Original VDC & $0.00611 \pm 0.00437$ & -- \\
Retrained VDC-64 & $0.00541 \pm 0.00406$ & 37/45 \\
Retrained VDC-128 & $0.00311 \pm 0.00213$ & 41/45 \\
pyvine-TLL & $0.01348 \pm 0.00760$ & 4/45 \\
Parametric selection & $0.00036 \pm 0.00082$ & 42/45 \\
\bottomrule
\end{tabular}

\end{table}

\paragraph{Full-vine evaluation and cost.}
Table~\ref{tab:retrained_uci} uses the same MAF inputs and exact test rows as Tables~\ref{tab:vine_nll} and~\ref{tab:vine_nll_50k}. VDC-128 improves mean NLL over the original in seven of eight settings, including a reversal of the TLL comparison on HEPMASS at $50{,}000$ training rows; MINIBOONE at $5{,}000$ is slightly worse. The improvements vary across datasets; no estimator leads in every setting. PIT-KS uses every retained test row, rather than a $5{,}000$-row diagnostic subset.

On chronological S\&P100, NLL is $-0.41833$ for VDC-64 and $-0.41876$ for VDC-128, versus $-0.41849$ for the original model and $-0.29276$ for TLL (bits/dim). The paired differences from the original are small: $40$-day moving-block bootstrap $95\%$ intervals are $[-0.00277,0.00317]$ and $[-0.00345,0.00323]$, respectively, conditional on the fitted models. All $339$ chronological test days are held out. We exclude the random S\&P100 split from this assessment because $272$ of its $339$ test days overlap model-development data, including $64$ used to score candidates; this overlap does not affect the original checkpoint's results.

These gains carry a computational cost. On an H100 allocation with four CPU threads, single-run chronological S\&P100 fit times are $16.0$\,s for VDC-64 and $81.0$\,s for VDC-128, versus $15.6$\,s for four-thread TLL. A separate warmed bivariate benchmark averages $4.92$ and $18.87$\,ms per edge over five repetitions of a $45$-edge batch; those batched measurements are not replacements for Table~\ref{tab:bivariate}'s single-edge timings. Evaluation uses float32, TF32 disabled, physical neural batch size $16$, and outer edge batches of $64$. The main paper's fitting-speed ratios are therefore not transferred to these estimators. Replication beyond two pilot initializations and a fresh OOD study remain future work.

\begin{table}[t]
\centering
\caption{\textbf{Retrained estimators on UCI.} NLL in bits/dim (Original, TLL, VDC-64, VDC-128; lower is better). KS-64/128 are mean marginal PIT-KS over all $50{,}000$ retained test rows for POWER, GAS, and HEPMASS, and all $3{,}648$ for MINIBOONE. Each row uses one training subsample and one model-training seed; these point estimates do not quantify retraining uncertainty. Original and TLL scores reproduce the main paper's tables.}
\label{tab:retrained_uci}
\scriptsize
\resizebox{\columnwidth}{!}{\begin{tabular}{llrrrrrr}
\toprule
Dataset & Train & Original & TLL & VDC-64 & VDC-128 & KS-64 & KS-128 \\
\midrule
POWER & 5k & -0.1413 & -0.1822 & -0.1544 & -0.1639 & 0.0170 & 0.0149 \\
GAS & 5k & -1.4773 & -1.7050 & -1.5461 & -1.5957 & 0.0448 & 0.0368 \\
HEPMASS & 5k & -0.2372 & -0.2009 & -0.2579 & -0.2643 & 0.0172 & 0.0161 \\
MINIBOONE & 5k & -0.9689 & -0.7444 & -0.9550 & -0.9661 & 0.0242 & 0.0232 \\
POWER & 50k & -0.1495 & -0.2169 & -0.1802 & -0.1924 & 0.0117 & 0.0103 \\
GAS & 50k & -1.4926 & -1.8113 & -1.6149 & -1.6800 & 0.0314 & 0.0245 \\
HEPMASS & 50k & -0.2468 & -0.2653 & -0.2845 & -0.2936 & 0.0150 & 0.0138 \\
MINIBOONE & 29,556 & -0.9784 & -0.8716 & -0.9791 & -0.9856 & 0.0229 & 0.0202 \\
\bottomrule
\end{tabular}
}
\end{table}

\fi

\end{document}